\documentstyle[jair,twoside,11pt,theapa,amsmath,graphicx]{article}

\jairheading{31}{2008}{217-257}{09/07}{02/08} \ShortHeadings{Loosely
Coupled Formulations for Automated Planning} {Van den Briel, Vossen
\& Kambhampati} \firstpageno{217}

\begin{document}

\title{Loosely Coupled Formulations for Automated Planning: An Integer Programming Perspective}

\author{\name Menkes H.L. van den Briel \email menkes@asu.edu \\
       \addr Department of Industrial Engineering \\
       Arizona State University, Tempe, AZ 85281 USA \\
       \AND
       \name Thomas Vossen \email vossen@colorado.edu \\
       \addr Leeds School of Business \\
       University of Colorado at Boulder, Boulder CO, 80309 USA \\
       \AND
       \name Subbarao Kambhampati \email rao@asu.edu \\
       \addr Department of Computer Science and Engineering \\
       Arizona State University, Tempe, AZ 85281 USA \\
       }


\maketitle

\begin{abstract}
We represent planning as a set of loosely coupled network flow problems, where each network corresponds to one of the state variables in the planning domain. The network nodes correspond to the state variable values and the network arcs correspond to the value transitions. The planning problem is to find a path (a sequence of actions) in each network such that, when merged, they constitute a feasible plan. In this paper we present a number of integer programming formulations that model these loosely coupled networks with varying degrees of flexibility. Since merging may introduce exponentially many ordering constraints we implement a so-called branch-and-cut algorithm, in which these constraints are dynamically generated and added to the formulation when needed. Our results are very promising, they improve upon previous planning as integer programming approaches and lay the foundation for integer programming approaches for cost optimal planning.
\end{abstract}

\section{Introduction}\label{sec:introduction}
While integer programming\footnote{We use the term integer programming to refer to integer linear programming unless stated otherwise.} approaches for automated planning have not been able to scale well against other compilation approaches (i.e.\ satisfiability and constraint satisfaction), they have been extremely successful in the solution of many real-world large scale optimization problems. Given that the integer programming framework has the potential to incorporate several important aspects of real-world automated planning problems (for example, numeric quantities and objective functions involving costs and utilities), there is significant motivation to investigate more effective integer programming formulations for classical planning as they could lay the groundwork for large scale optimization (in terms of cost and resources) in automated planning. In this paper, we study a novel decomposition based approach for automated planning that yields very effective integer programming formulations.

Decomposition is a general approach to solving problems more
efficiently. It involves breaking a problem up into several smaller
subproblems and solving each of the subproblems separately. In this
paper we use decomposition to break up a planning problem into
several interacting (i.e.\ loosely coupled) components. In such a
decomposition, the planning problem involves both finding solutions to
the individual components and trying to merge them
into a feasible plan. This general approach, however,
prompts the following questions: (1) what are the components, (2)
what are the component solutions, and (3) how hard is it to merge
the individual component solutions into a feasible plan?

\subsection{The Components}
\begin{figure}[t]
    \centerline{\includegraphics[width=5.2in]{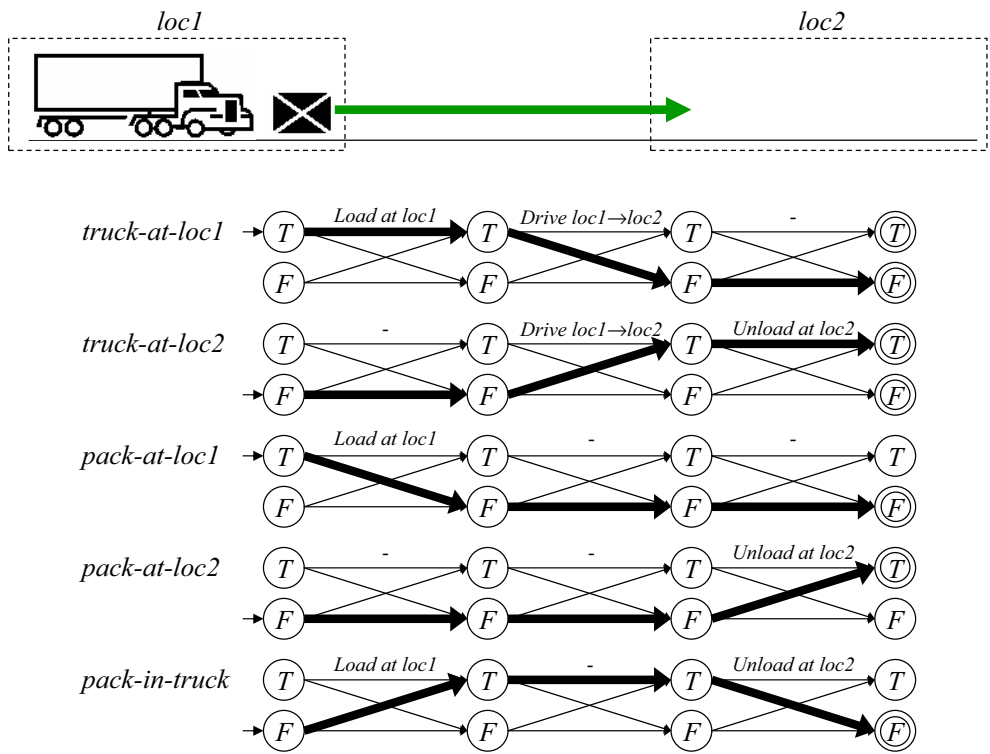}}
    \caption{Logistics example broken up into five components (binary-valued
    state variables) that are represented by network flow problems.}
\label{fig:fluentnetwork-bin}
\end{figure}

We let the components represent the state variables of the planning
problem. Figure \ref{fig:fluentnetwork-bin} illustrates this idea
using a small logistics example, with one truck and a package that
needs to be moved from location 1 to location 2. There are a total
of five components in this example, one for each state variable. We
represent the components by an appropriately defined network, where
the network nodes correspond to the values of the state variable
(for atoms this is $T$ = true and $F$ = false), and the
network arcs correspond to the value transitions. The source node in
each network, represented by a small in-arc, corresponds to the
initial value of the state variable. The sink node(s), represented
by double circles, correspond to the goal value(s) of the state
variable. Note that the effects of an action can trigger value
transitions in the state variables. For example, loading the package
at location 1 makes the atom $pack\text{-}in\text{-}truck$ true and
$pack\text{-}at\text{-}loc1$ false. In addition, loading the package at location 1
requires that the atom $truck\text{-}at\text{-}loc1$ is true.

\begin{figure}[t]
    \centerline{\includegraphics[width=5.2in]{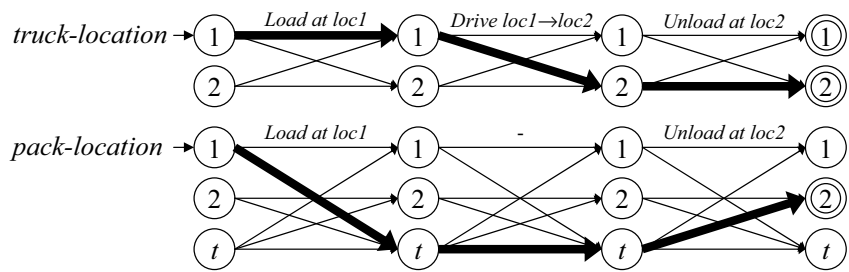}}
    \caption{Logistics example broken up into two components (multi-valued
    state variables) that are represented by network flow problems.}
\label{fig:fluentnetwork-mul}
\end{figure}

While the idea of components representing the state variables of the
planning problem can be used with any state variable representation,
it is particularly synergistic with multi-valued state variables.
Multi-valued state variables provide a more compact representation
of the planning problem than their binary-valued counterparts.
Therefore, by making the conversion to multi-valued state variables
we can reduce the number of components and create a better
partitioning of the constraints. Figure \ref{fig:fluentnetwork-mul}
illustrates the use of multi-valued state variables on our small
logistics example. There are two multi-valued state variables in
this problem, one to characterize the location of the truck and one
to characterize the location of the package. In our network
representation, the nodes correspond to the state variable values
($1 = at\text{-}loc1$, $2 = at\text{-}loc2$, and $t =
in\text{-}truck$), and the arcs correspond to the value transitions.

\subsection{The Component Solutions}
We let the component solutions represent a path of value transitions in the state variables. In the networks, nodes and arcs appear in layers. Each layer represents a plan period in which, depending on the structure of the network, one or more value transitions can occur. The networks in Figures \ref{fig:fluentnetwork-bin} and \ref{fig:fluentnetwork-mul} each have three layers (i.e.\ plan periods) and their structure allows values to persist or change exactly once per period. The layers are used to solve the planning problem incrementally. That is, we start with one layer in each network and try to solve the planning problem. If no plan is found, all networks are extended by one extra layer and a new attempt is made to solve the planning problem. This process is repeated until a plan is found or a time limit is reached. In Figures \ref{fig:fluentnetwork-bin} and \ref{fig:fluentnetwork-mul}, a path (i.e.\ a solution) from the source node to one of the sink nodes is highlighted in each network. Since the execution of an action triggers value transitions in the state variables, each path in a network corresponds to a sequence of actions. Consequently, \emph{the planning problem can be thought of as a collection of network flow problems where the problem is to find a path (i.e.\ a sequence of actions) in each of the networks}. However, interactions between the networks impose side constraints on the network flow problems, which complicate the solution process.

\subsection{The Merging Process}
We solve these loosely coupled networks using integer programming formulations. One design choice we make is that we expand all networks (i.e.\ components) together, so the cost of finding solutions for the individual networks as well as merging them depends on the difficulty of solving the integer programming formulation. This, in turn, typically depends on the size of the integer programming formulation, which is partly determined by the number of layers in each of the networks. The simplest idea is to have the number of layers of the networks equal the length of the plan, just as in sequential planning where the plan length equals the number of actions in the plan. In this case, there will be as many transitions in the networks as there are actions in the plan, with the only difference that a sequence of actions corresponding to a path in a network could contain no-op actions.

An idea to reduce the required number of layers is by allowing multiple actions to be executed in the same plan period. This is exactly what is done in Graphplan \cite{BLUFUR1995} and in other planners that have adopted the Graphplan-style definition of parallelism. That is, two actions can be executed in parallel (i.e.\ in the same plan period) as long as they are non-interfering. In our formulations we adopt more general notions of parallelism. In particular, we relax the strict relation between the number of layers in the networks and the length of the plan by changing the network representation of the state variables. For example, by allowing multiple transitions in each network per plan period we permit interfering actions to be executed in the same plan period. This, however, raises issues about how solutions to the individual networks are searched and how they are combined. When the network representations for the state variables allow multiple transitions in each network per plan period, and thus become more flexible, it becomes harder to merge the solutions into a feasible plan. Therefore, to evaluate the tradeoffs in allowing such flexible representations, we look at a variety of integer programming formulations.

We refer to the integer programming formulation that uses the network representation shown in Figures \ref{fig:fluentnetwork-bin} and \ref{fig:fluentnetwork-mul} as the \emph{one state change} model, because it allows at most one transition (i.e.\ state change) per plan period in each state variable. Note that in this network representation a plan period mimics the Graphplan-style parallelism. That is, two actions can be executed in the same plan period if one action does not delete the precondition or add-effect of the other action. A more flexible representation in which values can change at most once and persist before and after each change we refer to as the \emph{generalized one state change} model. Clearly, we can increase the number of changes that we allow in each plan period. The representations in which values can change at most twice or $k$ times, we refer to as the \emph{generalized two state change} and the \emph{generalized k state change} model respectively. One disadvantage with the generalized $k$ state change model is that it creates one variable for each way to do $k$ value changes, and thus introduces exponentially many variables per plan period. Therefore, another network representation that we consider allows a path of value transitions in which each value can be visited at most once per plan period. This way, we can limit the number of variables, but may introduce cycles in our networks. The integer programming formulation that uses this representation is referred to as the \emph{state change path} model.

In general, by allowing multiple transitions in each network per plan period (i.e.\ layer), the more complex the merging process becomes. In particular, the merging process checks whether the actions in the solutions of the individual networks can be linearized into a feasible plan. In our integer programming formulations, ordering constraints ensure feasible linearizations. There may, however, be exponentially many ordering constraints when we generalize the Graphplan-style parallelism. Rather than inserting all these constraints in the integer programming formulation up front, we add them as needed using a branch-and-cut algorithm. A branch-and-cut algorithm is a branch-and-bound algorithm in which certain constraints are generated dynamically throughout the branch-and-bound tree.

We show that the performance of our integer programming (IP) formulations show new potential and are competitive with SATPLAN04 \cite{KAU2004}. This is a significant result because it forms a basis for other more sophisticated IP-based planning systems capable of handling numeric constraints and non-uniform action costs. In particular, the new potential of our IP formulations has led to their successful use in solving partial satisfaction planning problems \cite{DOetal2007}. Moreover, it has initiated a new line of work in which integer and linear programming are used in heuristic state-space search for automated planning \cite{BENetal2007,BRIetal2007A}.

The remainder of this paper is organized as follows. In Section
\ref{sec:background} we provide a brief background on integer
programming and discuss some approaches that have used integer
programming to solve planning problems. In Section
\ref{sec:formulations} we present a series of integer programming
formulations that each adopt a different network representation. We
describe how we set up these loosely coupled networks, provide the
corresponding integer programming formulation, and discuss the
different variables and constraints. In Section
\ref{sec:branch-and-cut} we describe the branch-and-cut algorithm
that is used for solving these formulations. We provide a general
background on the branch-and-cut concept and show how we apply it to
our formulations by means of an example. Section \ref{sec:results}
provides experimental results to determine
which characteristics in our approach have the greatest impact on
performance. Related work is discussed in Section \ref{sec:related
work} and some conclusions are given in Section \ref{sec:conclusions}.

\section{Background} \label{sec:background}
Since our formulations are based on integer programming, we briefly
review this technique and discuss its use in planning. A \emph{mixed
integer program} is represented by a linear objective function and a
set of linear inequalities:
\begin{eqnarray*}
\min \{cx:Ax \geq b, x_1, ..., x_p \geq 0 \text{ and integer}, x_{p+1}, ..., x_n \geq 0 \},
\end{eqnarray*}
where $A$ is an $(m\times n)$ matrix, $c$ is an $n$-dimensional row
vector, $b$ is an $m$-dimensional column vector, and $x$ an
$n$-dimensional column vector of variables. If all variables are
continuous ($p=0$) we have a \emph{linear program}, if all variables
are integer ($p=n$) we have an \emph{integer program}, and if $x_1,
..., x_p \in \{0,1\}$ we have a \emph{mixed 0-1 program}. The set $S = \{x_1, ..., x_p \geq 0 \text{ and integer}, x_{p+1}, ..., x_n \geq 0:Ax \geq b\}$ is called the \emph{feasible region}, and an
$n$-dimensional column vector $x$ is called a \emph{feasible
solution} if $x\in S$. Moreover, the function $cx$ is called the \emph{objective function}, and the feasible solution $x^*$ is called an \emph{optimal solution} if the objective function is as small as possible, that is, $cx^* = \min \{cx:x\in S \}$

Mixed integer programming provides a rich modeling formalism that is more general than propositional logic. Any propositional clause can be represented by one linear inequality in 0-1 variables, but a single linear inequality in 0-1 variables may require exponentially many clauses \cite{HOO1988}.

The most widely used method for solving (mixed) integer programs is by applying a branch-and-bound algorithm to the \emph{linear programming relaxation}, which is much easier to solve\footnote{While the integer programming problem is $NP$-complete \cite{GARJOH1979} the linear programming problem is polynomially solvable \cite{KAR1984}.}. The linear programming (LP) relaxation is a linear program obtained from the original (mixed) integer program by relaxing the integrality constraints:
\begin{eqnarray*}
\min \{cx:Ax \geq b,x_1, ..., x_n \geq 0 \}
\end{eqnarray*}
Generally, the LP relaxation is solved at every node in the
branch-and-bound tree, until (1) the LP relaxation gives an integer
solution, (2) the LP relaxation value is inferior to the current
best feasible solution, or (3) the LP relaxation is infeasible,
which implies that the corresponding (mixed) integer program is
infeasible.

An \emph{ideal} formulation of an integer program is one for which the solution of the linear programming relaxation is integral. Even though every integer program has an ideal formulation \cite{WOL1998}, in practice it is very hard to characterize the ideal formulation as it may require an exponential number of inequalities. In problems where the ideal formulation cannot be determined, it is often desirable to find a \emph{strong} formulation of the integer program. Suppose that the feasible regions $P_1 = \{x \in R^n : A_1x \geq b_1 \}$ and $P_2 = \{x \in R^n : A_2x \geq b_2\}$ describe the linear programming relaxations of two IP formulations of a problem. Then we say that formulation for $P_1$ is stronger than formulation for $P_2$ if $P_1 \subset P_2$. That is, the feasible region $P_1$ is subsumed by the feasible region $P_2$. In other words $P_1$ improves the quality of the linear relaxation of $P_2$ by removing fractional extreme points.

There exist numerous powerful software packages that solve mixed
integer programs. In our experiments we make use of the commercial
solver CPLEX 10.0 \cite{CPLEX}, which is currently one of the
best LP/IP solvers.

The use of integer programming techniques to solve artificial
intelligence planning problems has an intuitive appeal,
especially given the success IP has had in solving similar types of
problems. For example, IP has been used extensively for solving problems
in transportation, logistics, and manufacturing. Examples include crew
scheduling, vehicle routing, and production planning problems
\cite{JOHetal2000}. One potential advantage is that IP techniques can
provide a natural way to incorporate several important aspects of
real-world planning problems, such as numeric constraints and
objective functions involving costs and utilities.

Planning as integer programming has, nevertheless, received only limited attention. One of the first approaches is described by Bylander \citeyear{BYL1997}, who proposes an LP heuristic for partial order planning algorithms. While the LP heuristic helps to reduce the number of expanded nodes, the evaluation is rather time-consuming. In general, the performance of IP often depends on the structure of the problem and on how the problem is formulated. The importance of developing strong IP formulations is discussed by Vossen et al.\ \citeyear{VOSetal1999}, who compare two formulations for classical planning: (1) a straightforward formulation based on the conversion of the propositional representation by SATPLAN which yields only mediocre results, and (2) a less intuitive formulation based on the representation of state transitions which leads to considerable performance improvements. Several ideas that further improve formulation based on the representation of state transitions are described by Dimopoulos \citeyear{DIM2001}. Some of these ideas are implemented in the IP-based planner Optiplan \cite{BRIKAM2005}. Approaches that rely on domain-specific knowledge are proposed by Bockmayr and Dimopoulos \citeyear{BOCDIM1998,BOCDIM1999}. By exploiting the structure of the planning problem these IP formulations often provide encouraging results. The use of LP and IP has also been explored for non-classical planning. Dimopoulos and Gerevini \citeyear{DIMGER2002} describe an IP formulation for temporal planning and Wolfman and Weld \citeyear{WOLWEL1999} use LP formulations in combination with a satisfiability-based planner to solve resource planning problems. Kautz and Walser \citeyear{KAUWAL1999} also solve resource planning problems, but use domain-specific IP formulations.

\section{Formulations} \label{sec:formulations}
This section describes four IP formulations that model the planning problem as a collection of loosely coupled network flow problems. Each network represents a state variable, in which the nodes correspond to the state variable values, and the arcs correspond to the value transitions. The state variables are based on the SAS+ planning formalism \cite{BACNEB1995}, which is a planning formalism that uses multi-valued state variables instead of binary-valued atoms. An action in SAS+ is modeled by its pre-, post- and prevail-conditions. The pre- and post-conditions express which state variables are changed and what values they must have before and after the execution of the action, and the prevail-conditions specify which of the unchanged variables must have some specific value before and during the execution of an action. A SAS+ planning problem is described by a tuple $\Pi = \langle C, A, s_0, s_* \rangle$ where:
\begin{itemize}
    \item $C = \{c_1, ..., c_n\}$ is a finite set of state variables, where each state variable $c\in C$ has an associated domain $V_c$ and an implicitly defined extended domain $V^+_c = V_c \cup \{u\}$, where $u$ denotes the \emph{undefined value}. For each state variable $c \in C$, $s[c]$ denotes the value of $c$ in state $s$. The value of $c$ is said to be \emph{defined} in state $s$ if and only if $s[c] \neq u$. The total state space $S = V_{c_1}\times ... \times V_{c_n}$ and the partial state space $S^+ = V^+_{c_1}\times ... \times V^+_{c_n}$ are implicitly defined.
    \item $A$ is a finite set of actions of the form $\langle pre, post, prev \rangle$, where $pre$ denotes the pre-conditions, $post$ denotes the post-conditions, and $prev$ denotes the prevail-conditions. For each action $a \in A$, $pre[c], post[c]$ and $prev[c]$ denotes the respective conditions on state variable $c$. The following two restrictions are imposed on all actions: (1) Once the value of a state variable is defined, it can never become undefined. Hence, for all $c \in C$, if $pre[c]\neq u$ then $pre[c]\neq post[c]\neq u$; (2) A prevail- and post-condition of an action can never define a value on the same state variable. Hence, for all $c \in C$, either $post[c] = u$ or $prev[c] = u$ or both.
    \item $s_0 \in S$ denotes the initial state and $s_* \in S^+$ denotes the goal state. While SAS+ planning allows the initial state and goal state to be both partial states, we assume that $s_0$ is a total state and $s_*$ is a partial state. We say that state $s$ is \emph{satisfied} by state $t$ if and only if for all $c\in C$ we have $s[c] = u$ or $s[c] = t[c]$. This implies that if $s_*[c] = u$ for state variable $c$, then any defined value $f \in V_c$ satisfies the goal for $c$.
\end{itemize}

To obtain a SAS+ description of the planning problem we use the
translator component of the Fast Downward planner \cite{HEL2006}.
The translator is a stand-alone component that contains a general
purpose algorithm which transforms a propositional description of
the planning problem into a SAS+ description. The algorithm provides
an efficient grounding that minimizes the state description length
and is based on the preprocessing algorithm of the MIPS planner
\cite{EDEHEL1999}.

In the remainder of this section we introduce some notation and
describe our IP formulations. The formulations are presented in such
a way that they progressively generalize the Graphplan-style
parallelism through the incorporation of more flexible network
representations. For each formulation we will describe the
underlying network, and define the variables and constraints. We
will not concentrate on the objective function as much because the
constraints will tolerate only feasible plans.

\subsection{Notation}
\begin{figure}[h]
    \centerline{\includegraphics[height=1.5in]{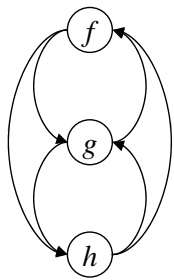}}
    \caption{An example of a domain transition graph, where $V_c = \{f,g,h\}$
    are the possible values (states) of $c$ and
    $E_c = \{(f,g),(f,h),(g,f),(g,h),(h,f),(h,g)\}$ are the possible
    value transitions in $c$.}\label{fig:scgraph}
\end{figure}

For the formulations that are described in this paper we assume that
the following information is given:
\begin{itemize}
  \item $C$: a set of state variables;
  \item $V_c$: a set of possible values (i.e.\ domain) for each state variable $c \in C$;
  \item $E_c$: a set of possible value transitions for each state variable $c \in C$;
  \item $G_c=(V_c,E_c)$ : a directed domain transition graph for every $c\in C$;
\end{itemize}
State variables can be represented by a domain transition graph, where the nodes correspond to the possible values, and the arcs correspond to the possible value transitions. An example of the domain transition graph of a variable is given in Figure \ref{fig:scgraph}. While the example depicts a complete graph, a domain transition graph does not need to be a complete graph.

\begin{figure}[h]
    \centerline{\includegraphics[height=1.5in]{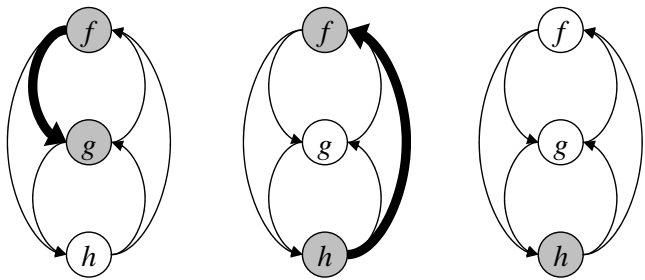}}
    \caption{An example of how action effects and prevail conditions are represented in
    a domain transition graph. Action $a$ has implications on three state
    variables $C^a = \{c_1, c_2, c_3\}$.
    The effects of $a$ are represented by $E^a_{c_1} = \{(f,g)\}$ and $E^a_{c_2} =
    \{(h,f)\}$, and the prevail condition of $a$ is represented by $V^a_{c_3} = \{h\}$.}
    \label{fig:scgraph action}
\end{figure}

Furthermore, we assume as given:
\begin{itemize}
  \item $E^a_c\subseteq E_c$ represents the effect of action $a$ in $c$;
  \item $V^a_c\subseteq V_c$ represents the prevail condition of action $a$ in $c$;
  \item $A^E_c := \{a\in A: |E^a_c|>0$\} represents the actions that have an effect in $c$, and $A^E_c(e)$ represents the actions that have the effect $e$ in $c$;
  \item $A^V_c := \{a\in A: |V^a_c|>0$\} represents the actions that have a prevail condition in $c$, and $A^V_c(f)$ represents the actions that have the prevail condition $f$ in $c$;
  \item $C^a := \{c\in C:a\in A^E_c\cup A^V_c$\} represents the state variables on which action $a$ has
  an effect or a prevail condition.
\end{itemize}
Hence, each action is defined by its effects (i.e.\ pre- and
post-conditions) and its prevail conditions. In SAS+ planning, actions can
have at most one effect or prevail condition in each state variable.
In other words, for each $a \in A$ and $c \in C$, we have that $E^a_c$ and $V^a_c$ are empty or
$|E^a_c|+|V^a_c| \leq 1$. An example of how the effects and prevail
conditions affect one or more domain transition graphs is given in Figure
\ref{fig:scgraph action}.

In addition, we use the following notation:
\begin{itemize}
  \item $V^+_c(f)$: to denote the in-arcs of node $f$ in the domain transition graph
  $G_c$;
  \item $V^-_c(f)$: to denote the out-arcs of node $f$ in the domain transition graph
  $G_c$;
  \item $P^+_{c,k}(f)$: to denote paths of length $k$ in the domain transition graph $G_c$ that end at node $f$. Note that $P^+_{c,1}(f)$ = $V^+_c(f)$.
  \item $P^-_{c,k}(f)$: to denote paths of length $k$ in the domain transition graph $G_c$ that start at node $f$. Note that $P^-_{c,1}(f)$ = $V^-_c(f)$.
  \item $P^{\sim}_{c,k}(f)$: to denote paths of length $k$ in the domain transition graph $G_c$ that visit node $f$, but that do not start or end at $f$.
\end{itemize}

\subsection{One State Change (1SC) Formulation}
Our first IP formulation incorporates the network representation that we have seen in Figures \ref{fig:fluentnetwork-bin} and \ref{fig:fluentnetwork-mul}. The name \emph{one state change} relates to the number of transitions that we allow in each state variable per plan period. The restriction of allowing only one value transition in each network also restricts which actions we can execute in the same plan period. It happens to be the case that the network representation of the 1SC formulation incorporates the standard notion of action parallelism which is used in Graphplan \cite{BLUFUR1995}. The idea is that actions can be executed in the same plan period as long as they do not delete the precondition or add-effect of another action. In terms of value transitions in state variables, this is saying that actions can be executed in the same plan period as long as they do not change the same state variable (i.e.\ there is only one value change or value persistence in each state variable).

\subsubsection{State Change Network}
\begin{figure}[h]
    \centerline{\includegraphics[width=5.2in]{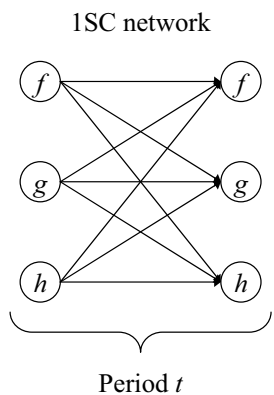}}
    \caption{One state change (1SC) network.}\label{fig:1scnetwork}
\end{figure}

Figure \ref{fig:1scnetwork} shows a single layer (i.e.\ period) of
the network which underlies the 1SC formulation. If we set up the IP
formulation with $T$ plan periods, then there will be $T+1$ layers
of nodes and $T$ layers of arcs in the network (the zeroth layer of
nodes is for the initial state and the remaining $T$ layers of nodes and
arcs are for the successive plan periods). For each possible state
transition there is an arc in the state change network. The
horizontal arcs correspond to the persistence of a value, and the
diagonal arcs correspond to the value changes. A solution path to an
individual network follows the arcs whose transitions are supported
by the action effect and prevail conditions that appear in the
solution plan.

\subsubsection{Variables}\label{sec:vars-1sc}
We have two types of variables in this formulation: action variables
to represent the execution of an action, and arc flow variables to
represent the state transitions in each network. We use separate
variables for changes in a state variable (the diagonal arcs in the
1SC network) and for the persistence of a value in a state variable
(the horizontal arcs in the 1SC network). The variables are defined
as follows:

\begin{itemize}
  \item $x^a_t \in \{0,1\}$, for $a\in A, 1\leq t\leq T$; $x^a_t$ is
  equal to $1$ if action $a$ is executed at plan period $t$, and $0$
  otherwise.
  \item $\bar{y}_{c,f,t} \in \{0,1\}$, for $c\in C$, $f\in V_c$,
  $1\leq t\leq T$; $\bar{y}_{c,f,t}$ is equal to $1$ if the value $f$ of
  state variable $c$ persists at period $t$, and $0$ otherwise.
  \item $y_{c,e,t} \in \{0,1\}$, for $c\in C, e\in E_c, 1\leq t\leq T$;
  $y_{c,e,t}$ is equal to $1$ if the transition $e\in E_c$ in state variable $c$
  is executed at period $t$, and $0$ otherwise.\\
\end{itemize}

\subsubsection{Constraints}
There are two classes of constraints. We have constraints for the network flows in each state variable network and constraints for the action effects that determine the interactions between these networks. The 1SC integer programming formulation is:
\begin{itemize}
    \item State change flows for all $c\in C$, $f\in V_c$
\begin{eqnarray}
  \sum_{e\in V^-_c(f)} y_{c,e,1} + \bar{y}_{c,f,1} & = & \left\{
\begin{array}{ll}
1 \quad\textrm{if } f = s_0[c]\\
0 \quad\textrm{otherwise}.
\end{array}\right. \label{eq:1sc flow1}\\
  \sum_{e\in V^-_c(f)} y_{c,e,t+1} + \bar{y}_{c,f,t+1} & = & \sum_{e\in V^+_c(f)} y_{c,e,t} + \bar{y}_{c,f,t} \quad\text{for } 1\leq t\leq T-1 \label{eq:1sc flow2}\\
  \sum_{e\in V^+_c(f)} y_{c,e,T} + \bar{y}_{c,f,T} & = & 1 \quad\textrm{if } f = s_*[c] \label{eq:1sc flow3}
\end{eqnarray}
    \item Action implications for all $c\in C$, $1\leq t\leq T$
\begin{eqnarray}
\sum_{a\in A:e\in E^a_c} x^a_t & = & y_{c,e,t} \quad \text{for }
e\in E_c \label{eq:1sc effect} \\
 x^a_t & \leq & \bar{y}_{c,f,t} \quad \text{for }
a\in A, f\in V^a_c \label{eq:1sc prevail}
\end{eqnarray}
\end{itemize}
Constraints \eqref{eq:1sc flow1}, \eqref{eq:1sc flow2}, and \eqref{eq:1sc flow3} are the network flow constraints for state variable $c\in C$. Constraint \eqref{eq:1sc flow1} ensures that the path of state transitions begins in the initial state of the state variable and constraint \eqref{eq:1sc flow3} ensures that, if a goal exists, the path ends in the goal state of the state variable. Note that, if the goal value for state variable $c$ is undefined (i.e.\ $s_*[c] = u$) then the path of state transitions may end in any of the values $f\in V_c$. Hence, we do not need a goal constraint for the state variables whose goal states $s_*[c]$ are undefined. Constraint \eqref{eq:1sc flow2} is the flow conservation equation and enforces the continuity of the constructed path.

Actions may introduce interactions between the state variables. For
instance, the effects of the $load$ action in our logistics example
affect two different state variables. Actions link state variables
to each other and these interactions are represented by the action
implication constraints. For each transition $e \in E_c$, constraints \eqref{eq:1sc effect} link the
action execution variables that have $e$ as an effect (i.e.\ $e \in E^a_c$) to the arc flow variables. For
example, if an action $x^a_t$ with effect $e \in E^a_c$ is
executed, then the path in state variable $c$ must follow the arc
represented by $y_{c,e,t}$. Likewise, if we choose to follow the arc
represented by $y_{c,e,t}$, then exactly one action $x^a_t$ with
$e \in E^a_c$ must be executed. The summation on the left hand side
prevents two or more actions from interfering with each other, hence
only one action may cause the state change $e$ in state variable $c$
at period $t$.

Prevail conditions of an action link state variables in a similar
way as the action effects do. Specifically, constraint \eqref{eq:1sc
prevail} states that \emph{if} action $a$ is executed at period $t$
($x^a_t = 1$), \emph{then} the prevail condition $f\in V^a_c$ is required in state
variable $c$ at period $t$ ($\bar{y}_{c,f,t} = 1$).

\subsection{Generalized One State Change (G1SC) Formulation}
In our second formulation we incorporate the same network representation as in the 1SC formulation, but adopt a more general interpretation of the value transitions, which leads to an unconventional notion of action parallelism. For the G1SC formulation we relax the condition that parallel actions can be arranged in \emph{any order} by requiring a weaker condition. We allow actions to be executed in the same plan period as long as \emph{there exists} some ordering that is feasible. More specifically, within a plan period a set of actions is feasible if (1) there exists an ordering of the actions such that all preconditions are satisfied, and (2) there is at most one state change in each of the state variables. This generalization of conditions is similar to what Rintanen, Heljanko and Niemel\"{a} \citeyear{RINetal2006} refer to as the $\exists$-step semantics semantics.

To illustrate the basic concept, let us again examine our small
logistics example introduced in Figure \ref{fig:fluentnetwork-bin}.
The solution to this problem is to load the package at location 1,
drive the truck from location 1 to location 2, and unload the
package at location 2. Clearly, this plan would require three plan
periods under Graphplan-style parallelism as these three actions interfere
with each other. If, however, we allow the $load\ at\ loc1$ and the
$drive\ loc1 \rightarrow loc2$ action to be executed in the same plan period,
then there exists some ordering between these two actions that is
feasible, namely load the package at the location 1 before driving
the truck to location 2. The key idea behind this example should be
clear: while it may not be possible to find a set of actions that
can be linearized in any order, there may nevertheless exist
\emph{some} ordering of the actions that is viable. The question is,
of course, how to incorporate this idea into an IP formulation.

\begin{figure}[h]
    \centerline{\includegraphics[width=5.2in]{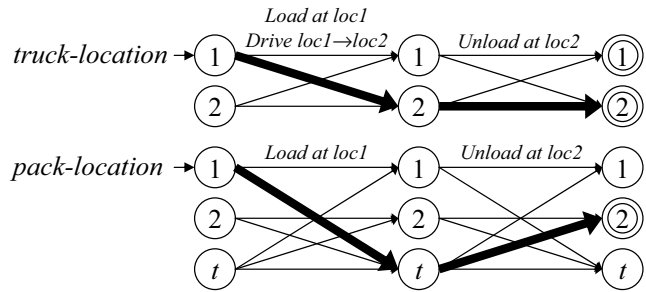}}
    \caption{Logistics example represented by network flow problems with
    generalized arcs.}\label{fig:network-g1sc}
\end{figure}

This example illustrates that we are looking for a set of constraints
that allow sets of actions for which: (1) all action preconditions
are met, (2) there exists an ordering of the actions at each plan
period that is feasible, and (3) within each state variable, the
value is changed at most once. The incorporation of these ideas only
requires minor modifications to the 1SC formulation. Specifically,
we need to change the action implication constraints for the prevail
conditions and add a new set of constraints which we call the
ordering implication constraints.
\subsubsection{State Change Network}
\begin{figure}[h]
    \centerline{\includegraphics[width=5.2in]{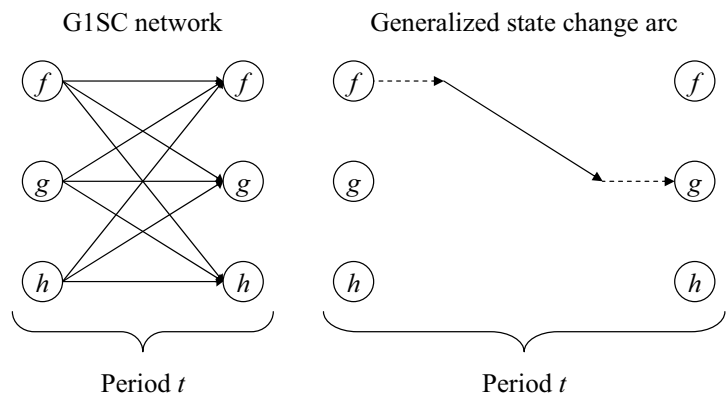}}
    \caption{Generalized one state change (G1SC) network.}\label{fig:g1scnetwork}
\end{figure}

The minor modifications are revealed in the G1SC network. While the
network itself is identical to the 1SC network, the interpretation
of the transition arcs is somewhat different. To incorporate the new
set of conditions, we implicitly allow values to persist
(the dashed horizontal arcs in the G1SC network) at the tail
and head of each transition arc. The interpretation of these
implicit arcs is that in each plan period a value may be required as a prevail condition,
then the value may change, and the new value may also be required as a prevail condition as shown in Figure \ref{fig:g1scnetwork}.
\subsubsection{Variables}
Since the G1SC network is similar to the 1SC network the same variables are used, thus, action variables to represent the execution of an action, and arc flow variables to represent the flow through each network. The difference in the interpretation of the state change arcs is dealt with in the constraints of the G1SC formulation, and therefore does not introduce any new variables. For the variable definitions, we refer to Section \ref{sec:vars-1sc}.
\subsubsection{Constraints}
We now have three classes of constraints, that is, constraints for
the network flows in each state variable network, constraints for
linking the flows with the action effects and prevail conditions,
and ordering constraints to ensure that the actions in the plan can
be linearized into a feasible ordering.

The network flow constraints for the G1SC formulation are identical to those in the 1SC formulation given by \eqref{eq:1sc flow1}-\eqref{eq:1sc flow3}. Moreover, the constraints that link the flows with the action effects are equal to the action effect constraints in the 1SC formulation given by \eqref{eq:1sc effect}. The G1SC formulation differs from the 1SC formulation in that it relaxes the condition that parallel actions can be arranged in any order by requiring a weaker condition. This weaker condition affects the constraints that link the flows with the action prevail conditions, and introduces a new set of ordering constraints. These constraints of the G1SC formulation are given as follows:
\begin{itemize}
    \item Action implications for all $c\in C$, $1\leq t\leq T$
\begin{eqnarray}
 x^a_t & \leq & \bar{y}_{c,f,t} + \sum_{e\in V^+_c(f)} y_{c,e,t}
 + \sum_{e\in V^-_c(f)} y_{c,e,t} \quad \text{for }
a\in A, f\in V^a_c \label{eq:g1sc prevail}
\end{eqnarray}
    \item Ordering implications
\begin{eqnarray}
  \sum_{a \in V(\Delta )} x^a_t & \leq & |V(\Delta )| - 1
  \text{ for all cycles $\Delta \in G^{prec}$ }\label{eq:g1sc cycle}
\end{eqnarray}
\end{itemize}

Constraint \eqref{eq:g1sc prevail} incorporates this new set of
conditions for which actions can be executed in the same plan period. In particular, we need to
ensure that for each state variable $c$, the value $f \in V_c$ holds
if it is required by the prevail condition of action $a$ at plan
period $t$. There are three possibilities: (1) The value $f$ holds for
$c$ throughout the period. (2) The value $f$ holds initially for
$c$, but the value is changed to a value other than $f$ by another
action. (3) The value $f$ does not hold initially for $c$, but the
value is changed to $f$ by another action. In either of the three
cases the value $f$ holds at some point in period $t$ so that the
prevail condition for action $a$ can be satisfied. In words, the value $f$ may
prevail implicitly as long as there is a state change
that includes $f$. As before, the prevail implication constraints
link the action prevail conditions to the corresponding network
arcs.

The action implication constraints ensure that the preconditions of
the actions in the plan are satisfied. This, however, does not
guarantee that the actions can be linearized into a feasible order.
Figure \ref{fig:g1scnetwork} indicates that there are implied
orderings between actions. Actions that require the value $f$ as a prevail condition must
be executed before the action that changes $f$ into $g$. Likewise,
an action that changes $f$ into $g$ must be executed before
actions that require the value $g$ as a prevail condition. The state change flow and action
implication constraints outlined above indicate that there is an
ordering between the actions, but this ordering could be cyclic and
therefore infeasible. To make sure that an ordering is acyclic we
start by creating a directed \textit{implied precedence graph}
$G^{prec} =(V^{prec},E^{prec})$. In this graph the nodes $a\in
V^{prec}$ correspond to the actions, that is, $V^{prec}=A$, and we
create a directed arc (i.e.\ an ordering) between two nodes $(a,b)\in
E^{prec}$ if action $a$ has to be executed before action $b$ in time
period $t$, or if $b$ has to be executed after $a$. In particular, we
have
\begin{eqnarray*}
  E^{prec} & = & \bigcup_{\substack{(a,b)\in A\times A , c\in C, f\in V^a_c, e\in E^b_c: \\
 e\in V^-_{c,f}}} (a, b) \cup \bigcup_{\substack{(a,b)\in A\times A, c\in C, g\in V^b_c, e\in E^a_c: \\ e\in V^+_{c,g}}} (a, b)
\end{eqnarray*}

\begin{figure}[h]
    \centerline{\includegraphics[width=4.2in]{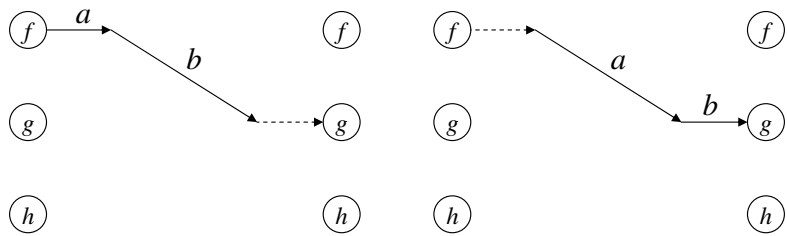}}
    \caption{Implied orderings for the G1SC formulation.}\label{fig:impliedorder}
\end{figure}

The implied orderings become immediately clear from Figure \ref{fig:impliedorder}. The figure on the left depicts the first set of orderings in the expression of $E^{prec}$. It says that the ordering between two actions $a$ and $b$ that are executed in the same plan period is implied if action $a$ requires a value to prevail that action $b$ deletes. Similarly, the figure on the right depicts second set of orderings in the expression of $E^{prec}$. That is, an ordering is implied if action $a$ adds the prevail condition of $b$.

The ordering implication constraints ensure that the actions in the
final solution can be linearized. They basically involve putting an
$n$-ary mutex relation between the actions that are involved in each
cycle. Unfortunately, the number of ordering implication constraints
grows exponentially in the number of actions. As a result, it will
be impossible to solve the resulting formulation using standard
approaches. We address this complication by implementing a
branch-and-cut approach in which the ordering implication
constraints are added dynamically to the formulation. This approach is
discussed in Section \ref{sec:branch-and-cut}.

\subsection{Generalized $k$ State Change (G$k$SC) Formulation}
In the G1SC formulation actions can be executed in the same plan period if (1)
there exists an ordering of the actions such that all preconditions
are satisfied, and (2) there occurs at most one value change in each
of the state variables. One obvious generalization of this would be
to relax the second condition and allow at most $k_c$ value changes
in each state variable $c$, where $k_c \leq |V_c| - 1$. By
allowing multiple value changes in a state variable per plan period
we, in fact, permit a series of value changes. Specifically, the
G$k$SC model allows series of value changes.

Obviously, there is a tradeoff between loosening the networks versus the amount of work it takes to merge the individual plans. While we have not implemented the G$k$SC formulation, we provide some insight in this tradeoff by describing and evaluating the G$k$SC formulation with $k_c=2$ for all $c\in C$ We will refer to this special case as the generalized two state change (G2SC) formulation. One reason we restrict ourselves to this special case is that the general case of $k$ state changes would introduce exponentially many variables in the formulation. There are IP techniques, however, that deal with exponentially many variables \cite{DESetal2005}, but we will not discuss them here.
\subsubsection{State Change Network}
\begin{figure}[h]
    \centerline{\includegraphics[width=5.2in]{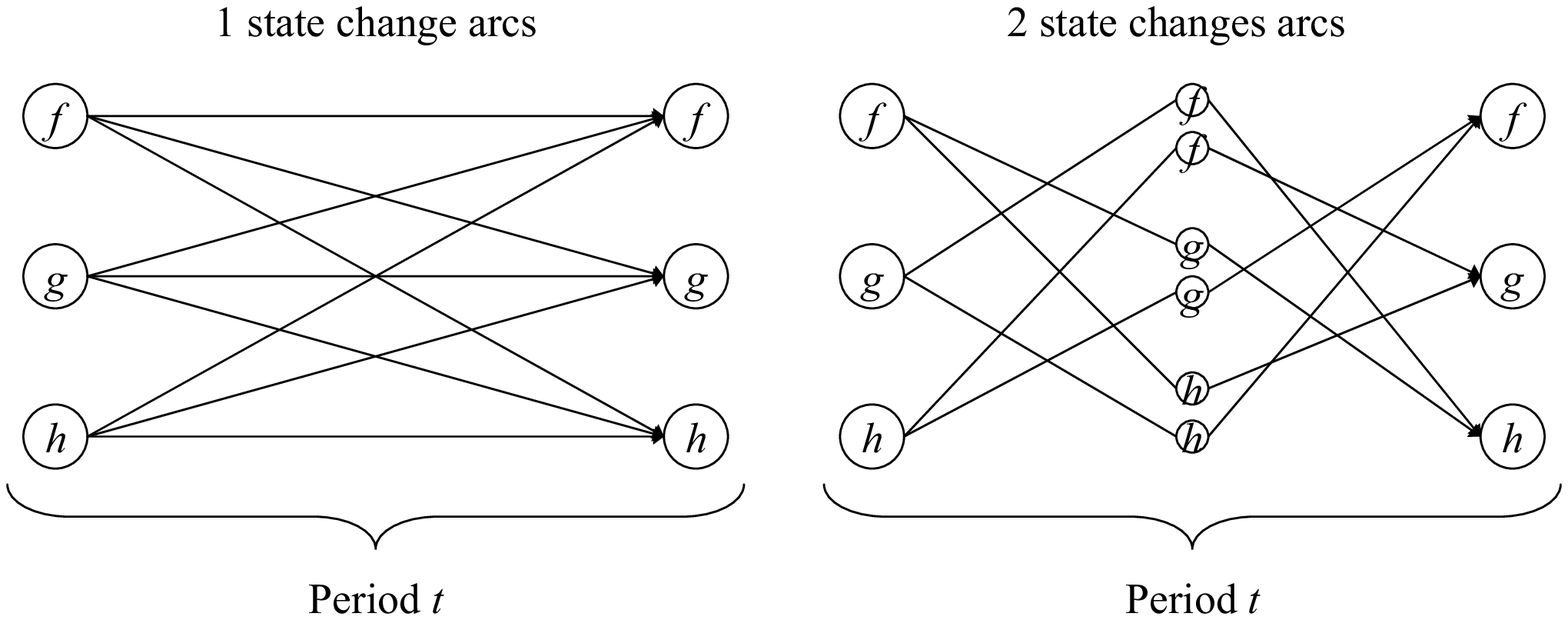}}
    \caption{Generalized two state change (G2SC) network. On the left the subnetwork that consists of generalized one state change arcs and no-op arcs, on the right the subnetwork that consists of the generalized two state change arcs. The subnetwork for the two state change arcs may include cyclic transitions, such as, $\{(f,g), (g,f)\}$ as long as $f$ is not the prevail condition of some action.} \label{fig:g2scnetwork}
\end{figure}

The network that underlies the G2SC formulation is equivalent to
G1SC, but spans an extra layer of nodes and arcs. This extra layer
allows us to have a series of two transitions per plan period. All
transitions are generalized and implicitly allow values to persist
just as in the G1SC network. Figure \ref{fig:g2scnetwork} displays
the network corresponding to the G2SC formulation. In the G2SC
network there are generalized one and two state change arcs. For
example, there is a generalized one state change arc for the
transition $(f,g)$, and there is a generalized two state changes
arc for the transitions $\{(f,g), (g,h)\}$. Since all arcs are
generalized, each value that is visited can also be persisted. We also allow cyclic transitions, such as, $\{(f,g), (g,f)\}$ if $f$ is not the prevail condition of some action. If we were to allow cyclic transitions in which $f$ is a prevail condition of an action, then the action ordering in a plan period can not be implied anymore (i.e.\ the prevail condition on $f$ would either have to occur before the value transitions to $g$, or after it transitions back to $f$). Thus if there is no prevail condition on $f$ then we can safely allow the cyclic transition $\{(f,g), (g,f)\}$.

\subsubsection{Variables}
As before we have variables representing the execution of an action,
and variables representing the flows over one state change (diagonal
arcs) or persistence (horizontal arcs). In addition, we have
variables representing paths over two consecutive state changes.
Hence, we have variables for each pair of state changes $(f,g,h)$
such that $(f,g)\in E_c$ and $(g,h)\in E_c$. We will restrict these
paths to visit unique values only, that is, $f\neq g$, $g\neq h$,
and $h\neq f$, or if $f$ is not a prevail condition of any action then we also allow paths where $f=h$. The variables from the G1SC formulation are also used in G2SC formulation. There is, however, an additional variable to represent the arcs that allow for two state changes:

\begin{itemize}
  \item $y_{c,e_1,e_2,t} \in \{0,1\}$, for $c\in C$,
  $(e_1,e_2)\in P_{c,2}$, $1\leq t\leq T$; $y_{c,e_1,e_2,t}$ is
  equal to $1$ if there exists a value $f\in V_c$ and transitions $e_1,e_2\in E_c$, such that $e_1\in V_c^+(f)$ and $e_2\in V_c^-(f)$, in state variable $c$ are executed at period $t$, and $0$ otherwise.
\end{itemize}

\subsubsection{Constraints}
We again have our three classes of constraints, which are given as
follows:

\begin{itemize}
    \item State change flows for all $c\in C$, $f\in V_c$
\begin{eqnarray}
  \sum_{(e_1, e_2)\in P^-_{c,2}(f)} y_{c,e_1,e_2,1} + \sum_{e\in V^-_c(f)} y_{c,e,1} + \bar{y}_{c,f,1} & = & \left\{
  \begin{array}{ll}
  1 \quad\textrm{if } f = s_0[c]\\
  0 \quad\textrm{otherwise}.
  \end{array}\right.  \label{eq:g2sc flow1}\\
  \sum_{(e_1, e_2)\in P^-_{c,2}(f)} y_{c,e_1,e_2,t+1} + \sum_{e\in V^-_c(f)} y_{c,e,t+1} + \bar{y}_{c,f,t+1} & = & \nonumber \\
  \qquad \sum_{(e_1, e_2)\in P^+_{c,2}(f)} y_{c,e_1,e_2,t} + \sum_{e\in V^+_c(f)} y_{c,e,t} + \bar{y}_{c,f,t} & & \text{for } 1\leq t\leq T-1 \label{eq:g2sc flow2}\\
  \sum_{(e_1, e_2)\in P^+_{c,2}(f)} y_{c,e_1,e_2,T} + \sum_{e\in V^+_c(f)} y_{c,e,T} + \bar{y}_{c,f,T} & = & 1 \quad\text{if }\{f \in s_*[c]\} \label{eq:g2sc flow3}
\end{eqnarray}
\end{itemize}
\begin{itemize}
    \item Action implications for all $c\in C$, $1\leq t\leq T$
\begin{eqnarray}
\sum_{a\in A:e\in E^a_c} x^a_t & = & y_{c,e,t} + \sum_{(e_1,e_2)\in
P_{c,2}:e_1=e \vee e_2=e} y_{c,e_1,e_2,t} \quad \text{for } e\in E_c
\label{eq:g2sc effect}
\end{eqnarray}
\begin{eqnarray}
 x^a_t & \leq & \bar{y}_{c,f,t} + \sum_{e\in V^+_c(f)} y_{c,e,t}
 + \sum_{e\in V^-_c(f)} y_{c,e,t} + \sum_{(e_1,e_2)\in P^{\sim}_{c,2}(f)} y_{c,e_1,e_2,t} + \nonumber \\
 & & \sum_{(e_1,e_2)\in P^+_{c,2}(f)} y_{c,e_1,e_2,t} + \sum_{(e_1,e_2)\in P^-_{c,2}(f)} y_{c,e_1,e_2,t}\quad \text{for } a\in A,
f\in V^a_c\label{eq:g2sc prevail}
\end{eqnarray}
\end{itemize}
\begin{itemize}
    \item Ordering implications
\begin{eqnarray}
  \sum_{a \in V(\Delta )} x^a_t & \leq & |V(\Delta )| - 1
  \text{ for all cycles $\Delta \in G^{prec}$ }\label{eq:g2sc cycle}
\end{eqnarray}
\end{itemize}

Constraints \eqref{eq:g2sc flow1}, \eqref{eq:g2sc flow2}, and
\eqref{eq:g2sc flow3} represent the flow constraints for the G2SC
network. Constraints \eqref{eq:g2sc effect} and \eqref{eq:g2sc
prevail} link the action effects and prevail conditions with the
corresponding flows, and constraint \ref{eq:g2sc cycle} ensures that
the actions can be linearized into some feasible ordering.

\subsection{State Change Path (PathSC) Formulation}
\begin{figure}
    \centerline{\includegraphics[width=5.2in]{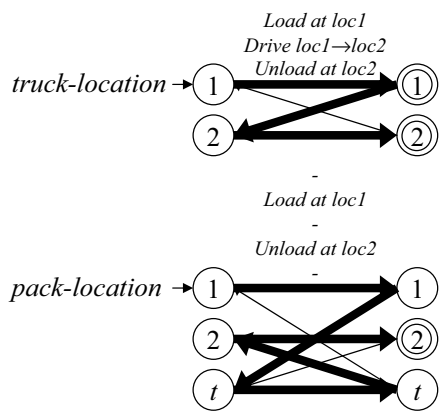}}
    \caption{Logistics example represented by network flow problems
    that allow a path of value transitions per plan period such that each value
    can be true at most once.}\label{fig:network-pathsc}
\end{figure}

There are several ways to generalize the network representation of
the G1SC formulation and loosen the interaction between the
networks. The G$k$SC formulation presented one generalization that
allows up to $k$ transitions in each state variable per plan period.
Since it uses exponentially many variables another way to generalize the network representation of
the G1SC formulation is by requiring that each value can
be true at most once per plan period. To illustrate this idea we
consider our logistics example again, but we now use a network
representation that allows a path of transitions per plan period as
depicted in Figure \ref{fig:network-pathsc}.

Recall that the solution to the logistics example consists of three
actions: first load the package at location 1, then drive the truck
from location 1 to location 2, and last unload the package at
location 2. Clearly, this solution would not be allowed
within a single plan period under Graphplan-style parallelism.
Moreover, it would also not be allowed within a single period in the
G1SC formulation. The reason for this is that the number of value
changes in the $package\text{-}location$ state variable is two.
First, it changes from $pack\text{-}at\text{-}loc1$ to
$pack\text{-}in\text{-}truck$, and then it changes from
$pack\text{-}in\text{-}truck$ to $pack\text{-}at\text{-}loc2$. As
before, however, there does exists an ordering of the three actions
that is feasible. The key idea behind this example is to show that
we can allow multiple value changes in a single period. If we limit
the value changes in a state variable to simple paths, that is, in one
period each value is visited at most once, then we can still use
implied precedences to determine the ordering restrictions.

\subsubsection{State Change Network}
In this formulation each value can be true at most once in each plan
period, hence the number of value transitions for each plan period is
limited to $k_c$ where $k_c = |V_c|- 1$ for each $c \in C$. In the
PathSC network, nodes appear in layers and correspond to the values
of the state variable. However, each layer now consists of twice as
many nodes. If we set up an IP encoding with a maximum number of plan periods
$T$ then there will be $T$ layers. Arcs within a layer correspond to
transitions or to value persistence, and arcs between layers
ensure that all plan periods are connected to each other.

\begin{figure}[h]
    \centerline{\includegraphics[width=3.4in]{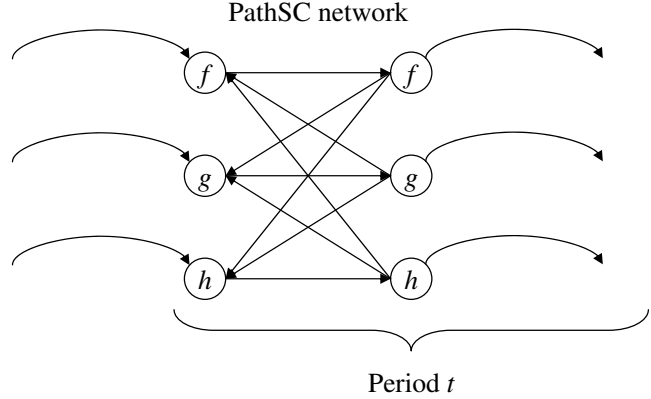}}
    \caption{Path state change (PathSC) network.}\label{fig:pathscnetwork}
\end{figure}

Figure \ref{fig:pathscnetwork} displays a network corresponding to
the state variable $c$ with domain $V_c = \{f,g,h\}$ that allows
multiple transitions per plan period. The arcs pointing rightwards
correspond to the persistence of a value, while the arcs pointing
leftwards correspond to the value changes. If more than one plan
period is needed the curved arcs pointing rightwards link the layers
between two consecutive plan periods. Note that with unit
capacity on the arcs, any path in the network can visit each node at
most once.

\subsubsection{Variables}
We now have action execution variables and arc flow variables (as defined in the previous formulations), and linking variables that connect the networks between two consecutive time periods. These variables are defined as
follows:

\begin{itemize}
  \item $z_{c,f,t} \in \{0,1\}$, for $c\in C, f\in V_c, 0\leq t\leq T$; $z_{c,f,t}$ is equal to $1$ if the value $f$ of state variable $c$ is the end value at period $t$, and $0$ otherwise.
\end{itemize}

\subsubsection{Constraints}
As in the previous formulations, we have state change flow constraints, action implication constraints, and ordering implication constraints. The main difference is the underlying network. The PathSC integer programming formulation is given as follows:
\begin{itemize}
    \item State change flows for all $c\in C$, $f\in V_c$
\begin{eqnarray}
  z_{c,f,0} & = & \left\{
\begin{array}{ll}
1 \quad\textrm{if } f = s_0[c]\\
0 \quad\textrm{otherwise}.
\end{array}\right. \label{eq:pathsc-init-flow} \\
  \sum_{e\in V^+_c(f)} y_{c,e,t} + z_{c,f,t-1} & = & \bar{y}_{c,f,t} \label{eq:pathsc-flow1}\\
  \bar{y}_{c,f,t} & = & \sum_{e\in V^-_c(f)} y_{c,e,t} + z_{c,f,t} \quad \text{for } 1\leq t\leq T-1 \label{eq:pathsc-flow2} \\
  z_{c,f,T} & = & 1 \quad \text{if } f \in s_*[c] \label{eq:pathsc-goal-flow}
\end{eqnarray}
    \item Action implications for all $c\in C$, $1\leq t \leq T$
\begin{eqnarray}
\sum_{a\in A:e\in E^a_c} x^a_t & = & y_{c,e,t} \quad \text{for }
e\in E_c \label{eq:pathsc-statechange} \\
 x^a_t & \leq & \bar{y}_{c,f,t} \quad\text{for } f \in
 V^a_c \label{eq:pathsc-prevail}
\end{eqnarray}
    \item Ordering implications
\begin{eqnarray}
  \sum_{a \in V(\Delta )} x^a_t & \leq & |V(\Delta )| - 1
  \text{ for all cycles } \Delta \in G^{prec'}\label{eq:pathsc-cycle}
\end{eqnarray}
\end{itemize}
Constraints \eqref{eq:pathsc-init-flow}-\eqref{eq:pathsc-goal-flow} are the network flow constraints. For each node, except for the initial and goal state nodes, they ensure a balance of flow (i.e.\ flow-in must equal flow-out). The initial state node has a supply of one unit of flow and the goal state node has a demand of one unit of flow, which are given by constraints \eqref{eq:pathsc-init-flow} and \eqref{eq:pathsc-goal-flow} respectively. The interactions that actions impose upon different state variables are represented by the action implication constraints \eqref{eq:pathsc-statechange} and \eqref{eq:pathsc-prevail}, which have been discussed earlier.

The implied precedence graph for this formulation is given by $G^{prec'} = (V^{prec'},E^{prec'})$. It has an extra set of arcs to incorporate the implied precedences that are introduced when two actions imply a state change in the same class $c \in C$. The nodes $a\in V^{prec'}$ again correspond to actions, and there is an arc $(a,b)\in E^{prec'}$ if action $a$ has to be executed before action $b$ in the same time period, or if $b$ has to be executed after $a$. More specifically, we have

\begin{eqnarray*}
 E^{prec'} & = & E^{prec} \cup \bigcup_{\substack{(a,b)\in A\times A, c\in C, f \in V_c, e\in E^a_c, e'\in E^b_c:\\
 e\in V^+_c(f) \wedge e'\in V^-_c(f)}} (a, b)\\
\end{eqnarray*}

As before, the ordering implication constraints \eqref{eq:pathsc-cycle} ensure that the actions in the solution plan can be linearized into a feasible ordering.

\section{Branch-and-Cut Algorithm}\label{sec:branch-and-cut}
IP problems are usually solved with an LP-based branch-and-bound
algorithm. The basic structure of this technique involves a binary
enumeration tree in which branches are pruned according to bounds
provided by the LP relaxation. The root node in the enumeration tree
represents the LP relaxation of the original IP problem and each
other node represents a subproblem that has the same objective
function and constraints as the root node except for some additional
bound constraints. Most IP solvers use an LP-based branch-and-bound algorithm
in combination with various preprocessing and probing techniques. In
the last few years there has been significant improvement in the
performance of these solvers \cite{BIX2002}.

In an LP-based branch-and-bound algorithm, the LP relaxation of the
original IP problem (the solution to the root node) will rarely be
integer. When some integer variable $x$ has a fractional solution
$v$ we branch to create two new subproblems, such that the bound
constraint $x\leq\lfloor v \rfloor$ is added to the left-child node,
and $x\geq\lceil v \rceil$ is added to the right-child node. This
branching process is carried out recursively to expand those
subproblems whose solution remains fractional. Eventually, after
enough bounds are placed on the variables, an integer solution is
found. The value of the best integer solution found so far, $Z^*$, is
referred to as the incumbent and is used for pruning.

In a minimization problem, branches emanating from nodes whose
solution value $Z_{\text{LP}}$ is greater than the current incumbent,
$Z^*$, can never give rise to a better integer solution as each child
node has a smaller feasible region than its parent. Hence, we can
safely eliminate such nodes from further consideration and prune them.
Nodes whose feasible region have been reduced to the empty set,
because too many bounds are placed on the variables, can be pruned
as well.

When solving an IP problem with an LP-based branch-and-bound algorithm we must consider the following two decisions. If several integer variables have a fractional solution, which variable should we branch on next, and if the branch we are currently working on is pruned, which subproblem should we solve next? Basic rules include use the ``most fractional variable'' rule for branching variable selection and the ``best objective value'' rule for node selection.

\begin{figure}
    \centerline{\includegraphics[height=4.2in]{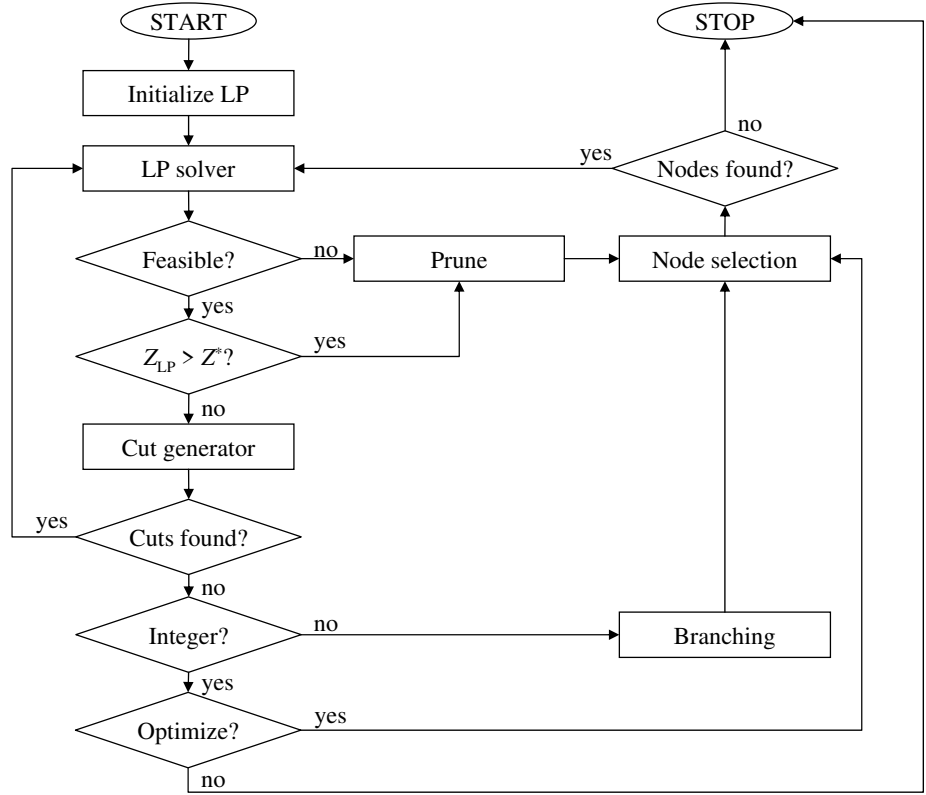}}
    \caption{Flowchart of our branch-and-cut algorithm. For finding any feasible solution (i.e.\ optimize = no) the algorithm stops as soon as the first feasible integer solution is found. When searching for the optimal solution (i.e.\ optimize = yes) for the given formulation we continue until no open nodes are left.}\label{fig:bcoverview}
\end{figure}

For our formulations a standard LP-based branch-and-bound algorithm approach is very ineffective
due to the large number (potentially exponentially many) ordering
implication constraints in the G1SC, G2SC, and PathSC formulations.
While it is possible to reduce the number of constraints by
introducing additional variables \cite{MAR1991}, the resulting
formulations would still be intractable for all but the smallest
problem instances. Therefore, we solve the IP formulations with a
so-called \emph{branch-and-cut} algorithm, which considers the
ordering implication constraints implicitly. A branch-and-cut
algorithm is a branch-and-bound algorithm in which certain
constraints are generated dynamically throughout the
branch-and-bound tree. A flowchart of our branch-and-cut algorithm
is given in Figure \ref{fig:bcoverview}.

If, after solving the LP relaxation, we are unable to prune the node
on the basis of the LP solution, the branch-and-cut algorithm tries
to find a violated cut, that is, a constraint that is valid but not
satisfied by the current solution. This is also known as the
\emph{separation problem}. If one or more violated cuts are found,
the constraints are added to the formulation and the LP is solved
again. If none are found, the algorithm creates a branch in the
enumeration tree (if the solution to the current subproblem is
fractional) or generates a feasible solution (if the solution to the
current subproblem is integral).

The basic idea of branch-and-cut is to leave out constraints from
the LP relaxation of which there are too many to handle efficiently,
and add them to the formulation only when they become binding at the
solution to the current LP. Branch-and-cut algorithms have
successfully been applied in solving hard large-scale optimization
problems in a wide variety of applications including scheduling,
routing, graph partitioning, network design, and facility location
problems \cite{capfis1997}.

In our branch-and-cut algorithm we can stop as soon as we find the first feasible solution, or we can implicitly enumerate all nodes (through pruning) and find the optimal solution for a given objective function. Note that our formulations can only be used to find bounded length optimal plans. That is, find the optimal plan given a plan period (i.e.\ a bounded length). In our experimental results, however, we focus on finding feasible solutions.

\subsection{Constraint Generation}
At any point during runtime that the cut generator is called we have
a solution to the current LP problem, which consists of the LP
relaxation of the original IP problem plus any added bound constraints and added cuts. In our
implementation of the branch-and-cut algorithm, we start with an LP
relaxation in which the ordering implication constraints are
omitted. So given a solution to the current LP relaxation, which
could be fractional, the separation problem is to determine whether
the solution violates one of the omitted ordering implication
constraints. If so, we identify the violated ordering implication
constraints, add them to the formulation, and resolve the new
problem.

\subsubsection{Cycle Identification}
In the G1SC, G2SC, and PathSC formulations an ordering implication
constraint is violated if there is a cycle in the implied precedence
graph. Separation problems involving cycles occur in numerous
applications. Probably the best known of its kind is the
traveling salesman problem in which subtours (i.e.\ cycles) are
identified and subtour elimination constraints are added to the
current LP. Our algorithm for separating cycles is based on the one described by Padberg and Rinaldi
\citeyear{PADRIN1991}. We are interested in finding the shortest
cycle in the implied precedence graph, as the shortest cycle cuts off
more fractional extreme points. The general idea behind this approach
is as follows:

\begin{enumerate}
  \item Given a solution to the LP relaxation, determine the
  subgraph $G_t$ for plan period $t$ consisting of all the nodes $a$ for which
$x^a_t > 0$.
  \item For all the arcs $(a,b)\in G_t$, define the weights $w_{a,b} :=
x^a_t + x^b_t -1$.
  \item Determine the shortest path distance $d_{a,b}$ for all pairs
($(a,b)\in G_t$) based on arc weights $\bar{w}_{a,b} := 1 -
  w_{a,b}$ (for example, using the Floyd-Warshall all-pairs shortest path
  algorithm).
  \item If $d_{a,b} - w_{b,a} < 0$ for some arc $(a,b)\in G_t$, there exists
a violated cycle constraint.
\end{enumerate}

While the general principles behind branch-and-cut algorithms are
rather straightforward, there are a number of algorithmic and
implementation issues that may have a significant impact on overall
performance. At the heart of these issues is the trade-off between
computation time spent at each node in the enumeration tree and the
number of nodes that are explored. One issue, for example, is to
decide when to generate violated cuts. Another issue is which of the
generated cuts (if any) should be added to the LP relaxation, and
whether and when to delete constraints that were added to the LP
before. In our implementation, we have only addressed these issues
in a straightforward manner: cuts are generated at every node in the
enumeration tree, the first cut found by the algorithm is added, and
constraints are never deleted from the LP relaxation. However, given the
potential of more advanced strategies that has been observed
in other applications, we believe there still may be
considerable room for improvement.

\subsubsection{Example}
In this section we will show the workings of our branch-and-cut
algorithm on the G1SC formulation using a small hypothetical example
involving two state variables $c_1$ and $c_2$, five actions $A1$,
$A2$, $A3$, $A4$, and $A5$, and one plan period. In particular we
will show how the cycle detection procedure works and how an
ordering implication constraint is generated.

\begin{figure}[h]
    \centerline{\includegraphics[width=5.2in]{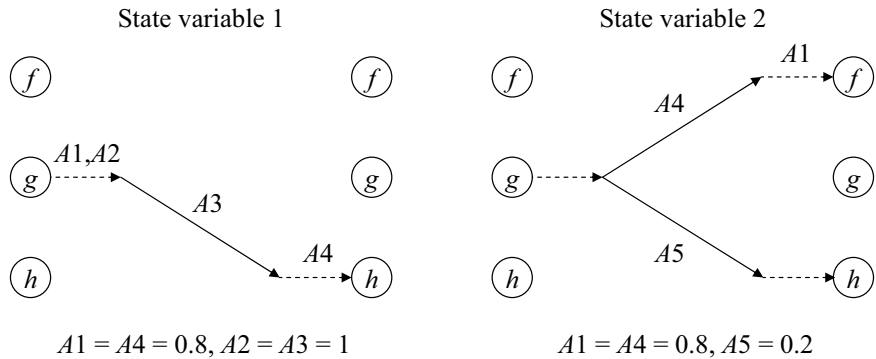}}
    \caption{Solution to a small hypothetical planning example.
    The solution to the current LP has flows over the indicated paths and
    executes actions $A1$, $A2$, $A3$, $A4$, and $A5$.}\label{fig:order1}
\end{figure}

Figure \ref{fig:order1} depicts a solution to the current LP of the
planning problem. For state variable $c_1$ we have that actions $A1$ and
$A2$ have a prevail condition on $g$, $A4$ has a prevail condition on $h$, and action
$A3$ has an effect that changes $g$ into $h$. Likewise, for state variable $c_2$ we
have that action $A4$ has an effect that changes $g$ into $f$, action $A5$ changes $g$
into $h$, and action $A1$ has a prevail condition on $f$. Note that the given
solution is fractional. Therefore some of the action variables have fractional values. In particular, we have $x^{A1} = x^{A4} = 0.8$, $x^{A5} = 0.2$, and $x^{A2} = x^{A3} = 1$. In other words, actions $A2$ and $A3$ are fully executed while actions $A1$, $A4$ and $A5$ are only fractionally executed. Clearly, in automated planning the fractional execution of an action has no meaning whatsoever, but it
is very common that the LP relaxation of an IP formulation gives a
fractional solution. We simply try to show that we can find a
violated cut even when we have a fractional solution. Also, note that
the actions $A4$ and $A5$ have interfering effects in $c_2$. While
this would generally be infeasible, the actions are executed only fractionally, so this is actually a feasible solution to the LP relaxation of the IP formulation.

\begin{figure}[h]
    \centerline{\includegraphics[width=5.2in]{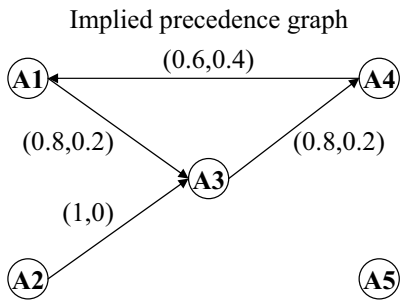}}
    \caption{Implied precedence graph for this example, where the labels show $(w_{a,b}, \bar{w}_{a,b})$.}\label{fig:order2}
\end{figure}

In order to determine whether the actions can be linearized into a
feasible ordering we first create the implied precedence graph
$G^{prec} = (V^{prec}, E^{prec})$, where we have $V^{prec} = \{A1,
A2, A3, A4, A5\}$ and $E^{prec} = $
$\{(A1,A3)$,$(A2,A3)$,$(A3,A4)$,$(A4,A1)\}$. The ordering $(A1,A3)$,
for example, is established by the effects of these actions in state
variable $c_1$. $A1$ has a prevail condition $g$ in $c_1$ while $A3$ changes $g$ to
$h$ in $c_1$, which implies that $A1$ must be executed before $A3$.
The other orderings are established in a similar way. The complete
implied precedence graph for this example is given in Figure
\ref{fig:order2}.

The cycle detection algorithm gets the implied precedence graph and
the solution to the current LP as input. Weights for each arc
$(a,b)\in E^{prec}$ are determined by the values of the action
variables in the current solution. We have the LP solution that is
given in Figure \ref{fig:order1}, so in this example we have
$w_{A1,A3} = w_{A3,A4} = 0.8$, $w_{A2,A3} = 1$, and $w_{A4,A1} =
0.6$. The length of the shortest path from $A1$ to $A4$ using
weights $\bar{w}_{a,b}$ is equal to $0.4$ $(0.2 + 0.2)$. Hence, we
have $d_{A1,A4} = 0.4$ and $w_{A4,A1} = 0.6$. Since $d_{A1,A4} -
w_{A4,A1} < 0$, we have a violated cycle (i.e.\ violated ordering
implication) that includes all actions that are on the shortest path
from $A1$ to $A4$ (i.e.\ $A1$, $A3$, and $A4$, which can be retrieved by the shortest path algorithm). This generates the
following ordering implication constraint $x^{A1}_1 + x^{A3}_1 +
x^{A4}_1 \leq 2$, which will be added to the current LP. Note that
this ordering constraint is violated by the current LP solution, as
$x^{A1}_1 + x^{A3}_1 + x^{A4}_1 = 0.8 + 1 + 0.8 = 2.6$. Once the
constraint is added to the LP, the next solution will select a set
of actions that does not violate the newly added cut. This
procedure continues until no cuts are violated and the solution is
integer.

\section{Experimental Results}\label{sec:results}
The described formulations are based on two key ideas. The first idea is to decompose the planning problem into several loosely coupled components and represent these components by an appropriately defined network. The second idea is to reduce the number of plan periods by adopting different notions of parallelism and use a branch-and-cut algorithm to dynamically add constraints to the formulation in order to deal with the exponentially many action ordering constraints in an efficient manner.

\begin{figure}[h]
\centering
\begin{tabular}{|l|l|}
  \hline
  \textbf{1SC} & \textbf{G1SC} \\
  Each state variable can change or     & Each state variable can change (and   \\
  prevail a value at most once per      & prevail a value before and after each \\
  plan period.                          & change) at most once per plan period. \\
                                        & \\
  \text{\ \ \ \ \ \ \ \ \ \ \ \ \ }\includegraphics[height=1.2in]{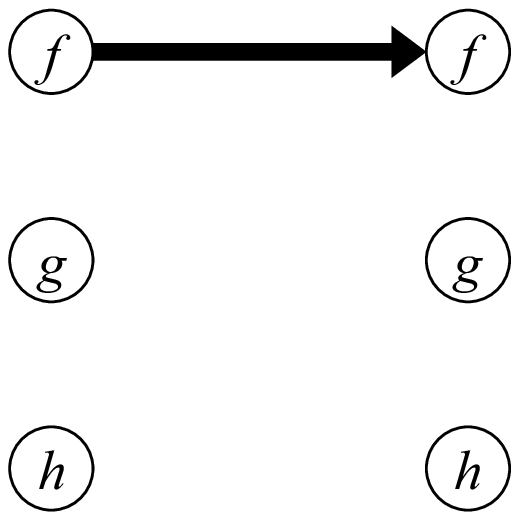} &
  \text{\ \ \ \ \ \ }\includegraphics[height=1.2in]{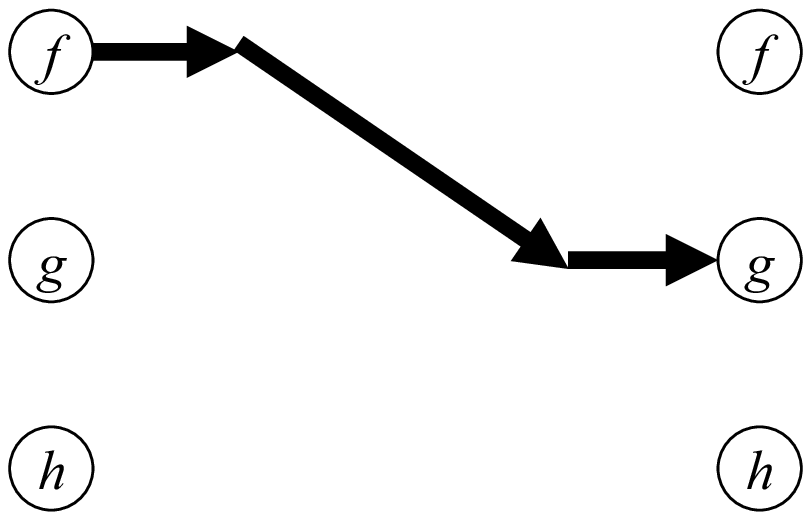} \\
  \hline
  \textbf{G2SC} & \textbf{PathSC} \\
  Each state variable can change (and   & The state variable can change any  \\
  prevail a value before and after each & number of times, but each value    \\
  change) at most \emph{twice} per plan & can be true at most once per plan \\
  period. Cyclic changes $(f,g,f)$ are  & period. \\
  allowed only if $f$ is \emph{not} the & \\
  prevail condition of some action      & \\
                                        & \\
  \text{\ \ \ }\includegraphics[height=1.2in]{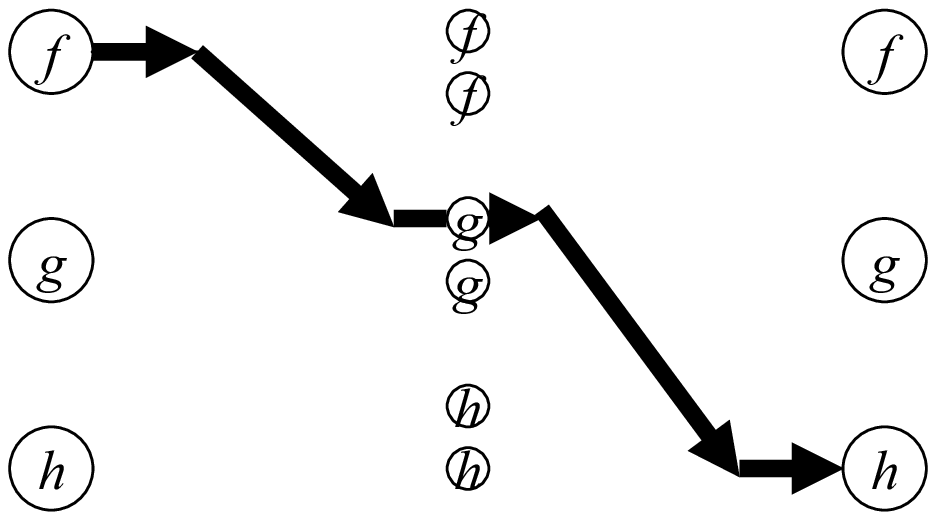} &
  \text{\ \ \ \ \ \ }\includegraphics[height=1.2in]{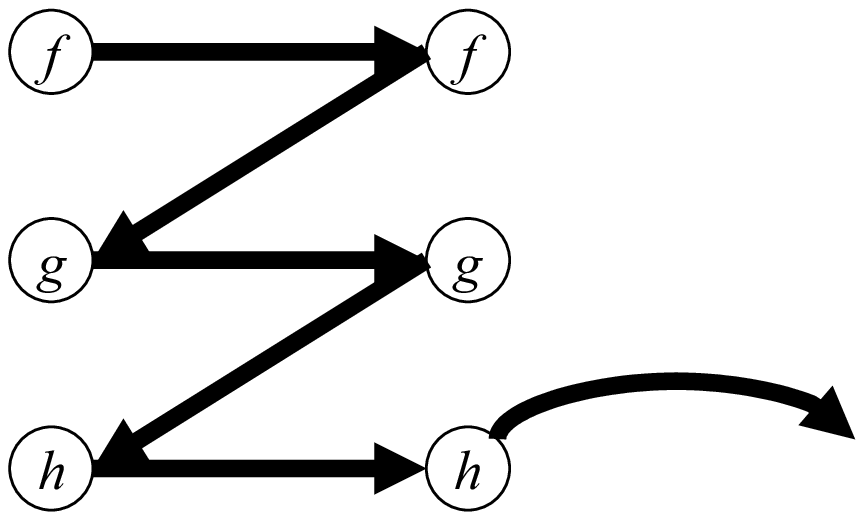} \\
  \hline
\end{tabular}
\caption{Overview of the 1SC, G1SC, G2SC, and PathSC
formulations.}\label{fig:overview}
\end{figure}

To evaluate the tradeoffs of allowing more flexible network representations we compare the performance of the one state change (1SC) formulation, the generalized one state change formulation (G1SC), the generalized two state change (G2SC) formulation, and the state change path (PathSC) formulation. For easy reference, an overview of these formulations is given in Figure \ref{fig:overview}.

In our experiments we focus on finding feasible solutions. Note, however, that our formulations can be used to do bounded length optimal planning. That is, given a plan period (i.e.\ a bounded length), find the optimal solution.

\subsection{Experimental Setup}
To compare and analyze our formulations we use the STRIPS domains from the second and third international planning competitions (IPC2 and IPC3 respectively). That is, Blocksworld, Logistics, Miconic, Freecell from IPC2 and Depots, Driverlog, Zenotravel, Rovers, Satellite, and Freecell from IPC3. We do not compare our formulations on the STRIPS domains from IPC4 and IPC5 mainly because of a peripheral limitation of the current implementation of the G2SC and PathSC formulations. In particular, the G2SC formulation cannot handle operators that change a state variable from an undefined value to a defined value, and the PathSC formulation cannot handle such operators if the domain size of the state variable is larger than two. Because of these limitations we could not test the G2SC formulation on the Miconic, Satellite and Rovers domains, and we could not test the PathSC formulation on the Satellite domain.

In order to setup our formulations we translate a STRIPS planning problem into a multi-valued state description using the translator of the Fast Downward planner \cite{HEL2006}. Each formulation uses its own network representation and starts by setting the number of plan periods $T$ equal to one. We try to solve this initial formulation and if no plan is found, $T$ is increased by one, and then try to solve this new formulation. Hence, the IP formulation is solved repeatedly until the first feasible plan is found or a 30 minute time limit (the same time limit that is used in the international planning competitions) is reached. We use CPLEX 10.0 \cite{CPLEX}, a commercial LP/IP solver, for solving the IP formulations on a 2.67GHz Linux machine with 1GB of memory.

We set up our experiments as follows. First, in Section \ref{sec:results-summary} we provide a brief overview of our main results by looking at aggregated results from IPC2 and IPC3. Second, in Section  \ref{sec:results1}, we give a more detailed analysis on our loosely coupled encodings for planning and focus on the tradeoffs of reducing the number of plan periods to solve a planning problem versus the increased difficulty in merging the solutions to the different components. Third, in Section \ref{sec:results3} we briefly touch upon how different state variable representations of the same planning problem can influence performance.

\subsection{Results Overview}\label{sec:results-summary}
In this general overview we compare our formulations to the following planning systems: Optiplan \cite{BRIKAM2005}, SATPLAN04 \cite{KAU2004}, SATPLAN06 \cite{KAUSEL2006}, and Satplanner \cite{RINetal2006}\footnote{We note that that SATPLAN04, SATPLAN06, Optiplan, and the 1SC formulation are ``step-optimal'' while the G1SC, G2SC, and PathSC formulations are not. There is, however, considerable controversy in the planning community as to whether the step-optimality guaranteed by Graphplan-style planners has any connection to plan quality metrics that users would be interested in. We refer the reader to Kambhampati \citeyear{KAM2006} for a longer discussion of this issue both by us and several prominent researchers in the planning community. Given this background, we believe it is quite reasonable to compare our formulations to step-optimal approaches, especially since our main aim here is to show that IP formulations have come a long way and that they can be made competitive with respect to SAT-based encodings. This in turn makes it worthwhile to consider exploiting other features of IP formulations, such as their amenability to a variety of optimization objectives as we have done in our recent work \cite{BRIetal2007A}.}.

Optiplan is an integer programming based planner that participated in the optimal track of the fourth international planning competition\footnote{A list of participating planners and their results is available at http://ipc04.icaps-conference.org/}. Like our formulations, Optiplan models state transitions but it does not use a factored representation of the planning domain. In particular, Optiplan represents state transitions in the atoms of the planning domain, whereas our formulations use multi-valued state variables. Apart from this, Optiplan is very similar to the 1SC formulation as they both adopt the Graphplan-style parallelism.

SATPLAN04, SATPLAN06, and Satplanner are satisfiability based planners. SATPLAN04 and SATPLAN06 are versions of the well known system SATPLAN \cite{KAUSEL1992}, which has a long track record in the international planning competitions. Satplanner has not received that much attention, but is among the state-of-the-art in planning as satisfiability. Like our formulations Satplanner generalizes the Graphplan-style parallelism to improve planning efficiency.

\begin{figure*}[h]
\centering
\begin{tabular}{cc}
  \includegraphics[width=2.9in]{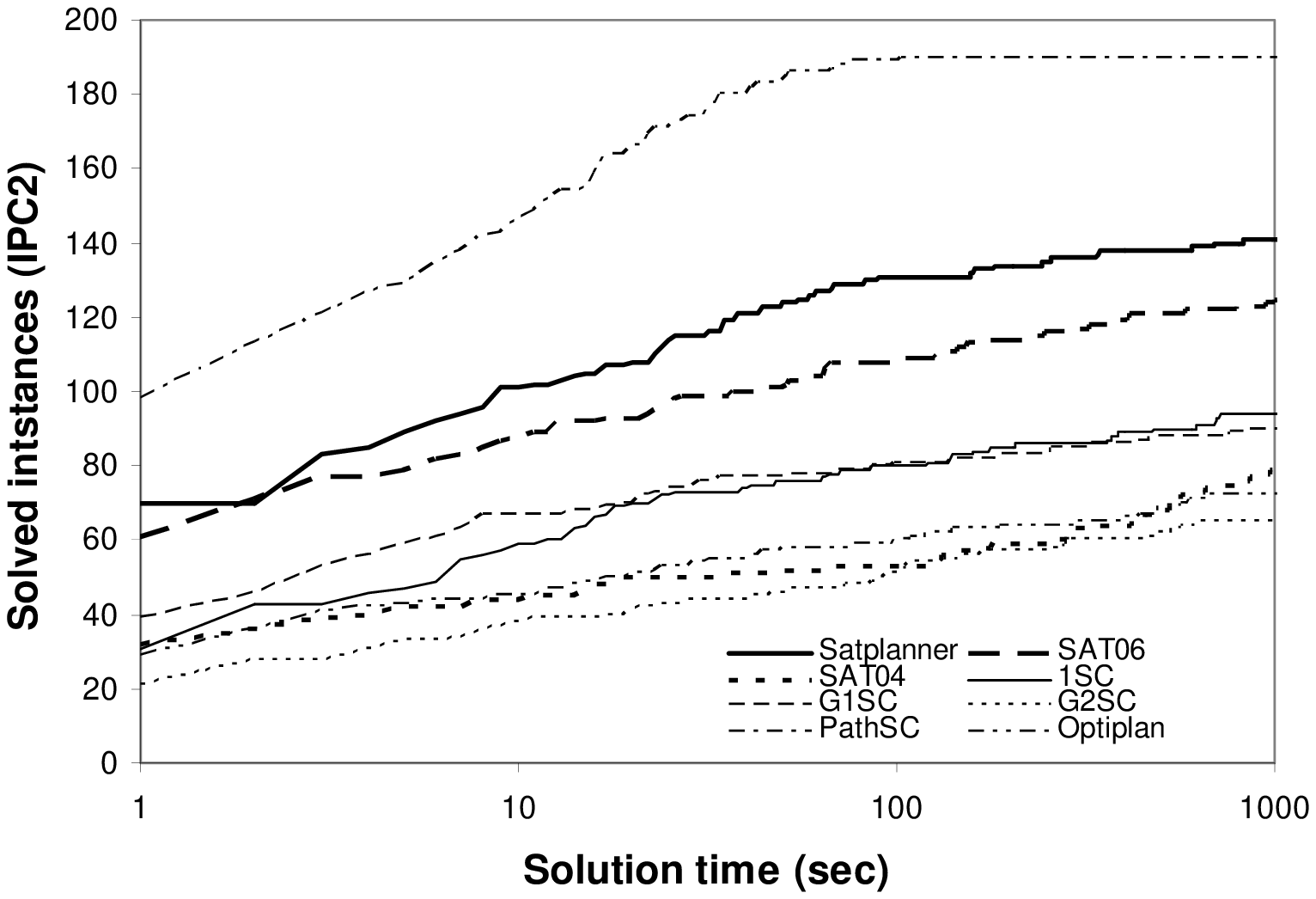} &
  \includegraphics[width=2.9in]{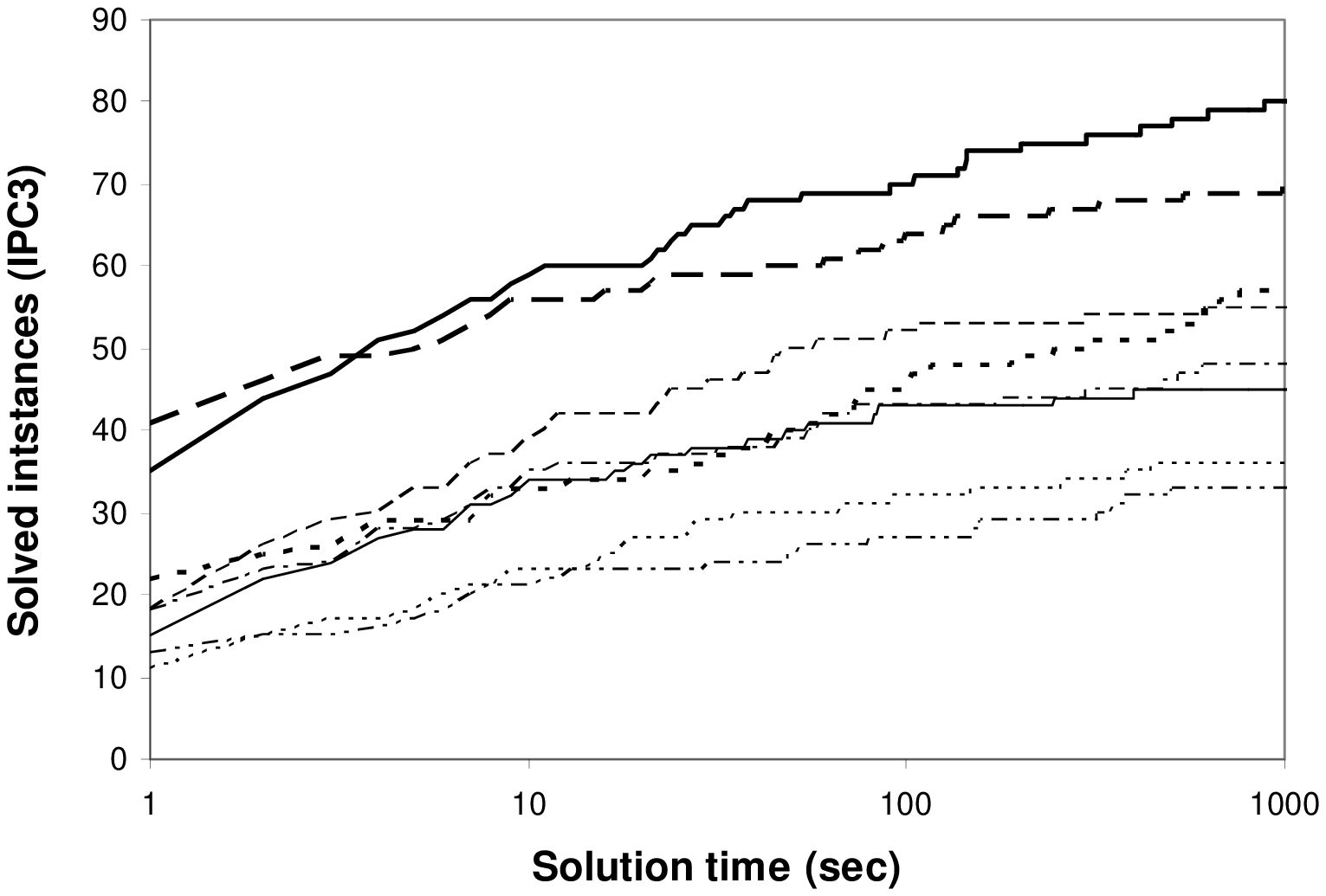}
\end{tabular}
\caption{Aggregate results of the second and third international planning competitions.} \label{fig:results-summary}
\end{figure*}

The main results are summarized by Figure \ref{fig:results-summary}. It displays aggregate results from IPC2 and IPC3, where the number of instances solved (y-axis) is drawn as a function of log time (x-axis). We must note that the graph with the IPC2 results favors the PathSC formulation over all other planners. However, as we will see in Section \ref{sec:results1}, this is mainly a reflection of its exceptional performance in the Miconic domain rather than its overall performance in IPC2. Morever, the graph with the IPC3 results does not include the Satellite domain. We decided to remove this domain, because we could not run it on the public versions of SATPLAN04 and SATPLAN06 nor the G2SC and PathSC formulations. While the results in Figure \ref{fig:results-summary} provide a rather coarse overview, they sum up the following main findings.

\begin{itemize}
  \item \emph{Factored planning using loosely coupled formulations helps improve performance}. Note that all integer programming formulations that use factored representations, that is 1SC, G1SC, G2SC, and PathSC (except the G2SC formulation which could not be tested on all domains), are able to solve more problem instances in a given amount of time than Optiplan, which does not use a factored representation. Especially, the difference between 1SC and Optiplan is remarkable as they both adopt the Graphplan-style parallelism. In Section \ref{sec:results1}, however, we will see that Optiplan does perform well in domains that are either serial by design or have a significant serial component.
  \item \emph{Decreasing the encoding size by relaxing the Graphplan-style parallelism helps improve performance}. This is not too surprising, Dimopoulos et al.\ \citeyear{DIMetal1997} already note that a reduction in the number of plan periods helps improve planning performance. However, this does not always hold because of the tradeoff between reducing the number plan periods versus the increased difficulty in merging the solutions to the different components. In Section \ref{sec:results1} we will see that different relaxations of Graphplan-style parallelism lead to different results. For example, the PathSC formulation shows superior performance in Miconic and Driverlog, but does poorly in Blocksworld, Freecell, and Zenotravel. Likewise, the G2SC formulation does well in Freecell, but it does not seem to excel in any other domain.
  \item \emph{Planning as integer programming shows new potential}. The conventional wisdom in the planning community has been that planning as integer programming cannot compete with planning as satisfiability or constraint satisfaction. In Figure \ref{fig:results-summary}, however, we see that the 1SC, G1SC and PathSC formulation can compete quite well with SATPLAN04. While SATPLAN04 is not state-of-the-art in planning as satisfiability anymore, it does show that planning as integer programming has come a long way. The fact that IP is competitive allows us to exploit its other virtues such as optimization \cite{DOetal2007,BENetal2007,BRIetal2007A}.
\end{itemize}

\subsection{Comparing Loosely Coupled Formulations for Planning}\label{sec:results1}
In this section we compare our IP formulations and try to evaluate the benefits of allowing more flexible network representations. Specifically, we are interested in the effects of reducing the number of plan periods required to solve the planning problem versus dealing with merging solutions to the different components. Reducing the number of plan periods can lead to smaller encodings, which can lead to improved performance. However, it also makes the merging of the loosely coupled components harder, which could worsen performance.

In order to compare our formulations we will analyze the following two things. First, we examine the performance of our formulations by comparing their solution times on problem instances from IPC2 and IPC3. In this comparison we will include results from Optiplan as it gives us an idea of the differences between a formulation based on Graphplan and formulations based on loosely coupled components. Moreover, it will also show us the improvements in IP based approaches for planning. Second, we examine the number of plan periods that each formulation needs to solve each problem instance. Also, we will look at the tradeoffs between reducing the number of plan periods and the increased difficulty in merging the solutions of the loosely coupled components. In this comparison we will include results from Satplanner because, just like our formulations, it adopts a generalized notion of the Graphplan-style parallelism.

We use the following figures and table. Figure \ref{fig:ipctime} shows the total solution time (y-axis) needed to solve the problem instances (x-axis), Figure \ref{fig:ipcsteps} shows the number of plan periods (y-axis) to solve the problem instances (x-axis), and Table \ref{tab:ipccuts} shows the number of ordering constraints that were added during the solution process, which can be seen as an indicator of the merging effort. The selected problem instances in Table \ref{tab:ipccuts} represent the five largest instances that could be solved by all of our formulations (in some domains, however, not all formulations could solve at least five problem instances).

The label $GPsteps$ in Figure \ref{fig:ipcsteps} represents the number of plan steps that SATPLAN06, a state-of-the-art Graphplan-based planner, would use. In the Satellite domain, however, we use the results from the 1SC formulation as we were unable to run the public version of SATPLAN06 in this domain. We like to point out that Figure \ref{fig:ipcsteps} is not intended to favor one formulation over the other, it simply shows that it is possible to generate encodings for automated planning that use drastically fewer plan periods than Graphplan-based encodings.

\subsubsection{Results: Planning Performance}
Blocksworld is the only domain in which Optiplan solves more problems than our formulations. In Zenotravel and Satellite, Optiplan is generally outperformed with respect to solution time, and in Rovers and Freecell, Optiplan is generally outperformed with respect to the number of problems solved. As for the other IP formulations, the G1SC provides the overall best performance and the performance of the PathSC formulation is somewhat irregular. For example, in Miconic, Driverlog and Rovers the PathSC formulation does very well, but in Depots and Freecell it does rather poorly.

\begin{figure*}
\centering
\begin{tabular}{cc}
  \includegraphics[width=2.9in]{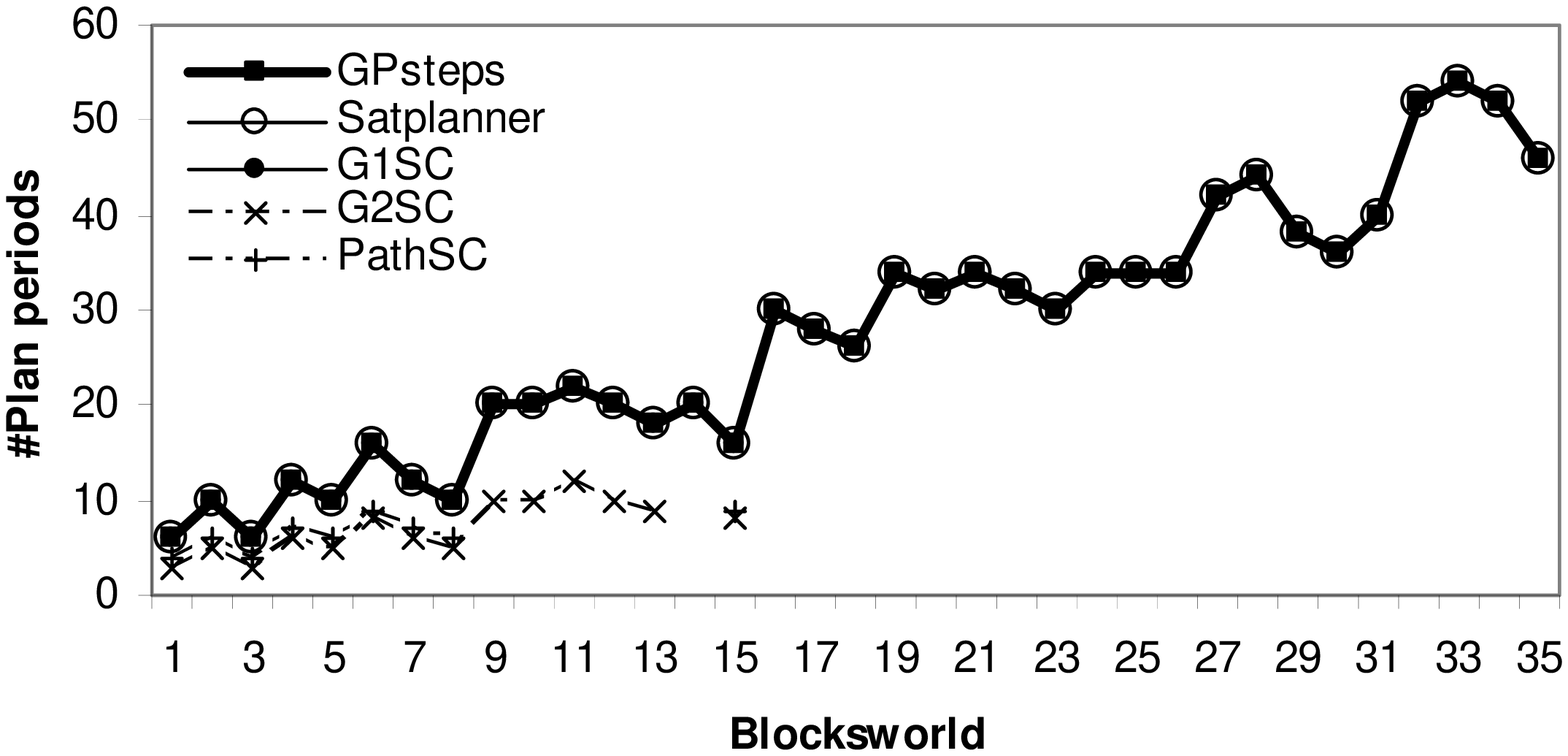} &
  \includegraphics[width=2.9in]{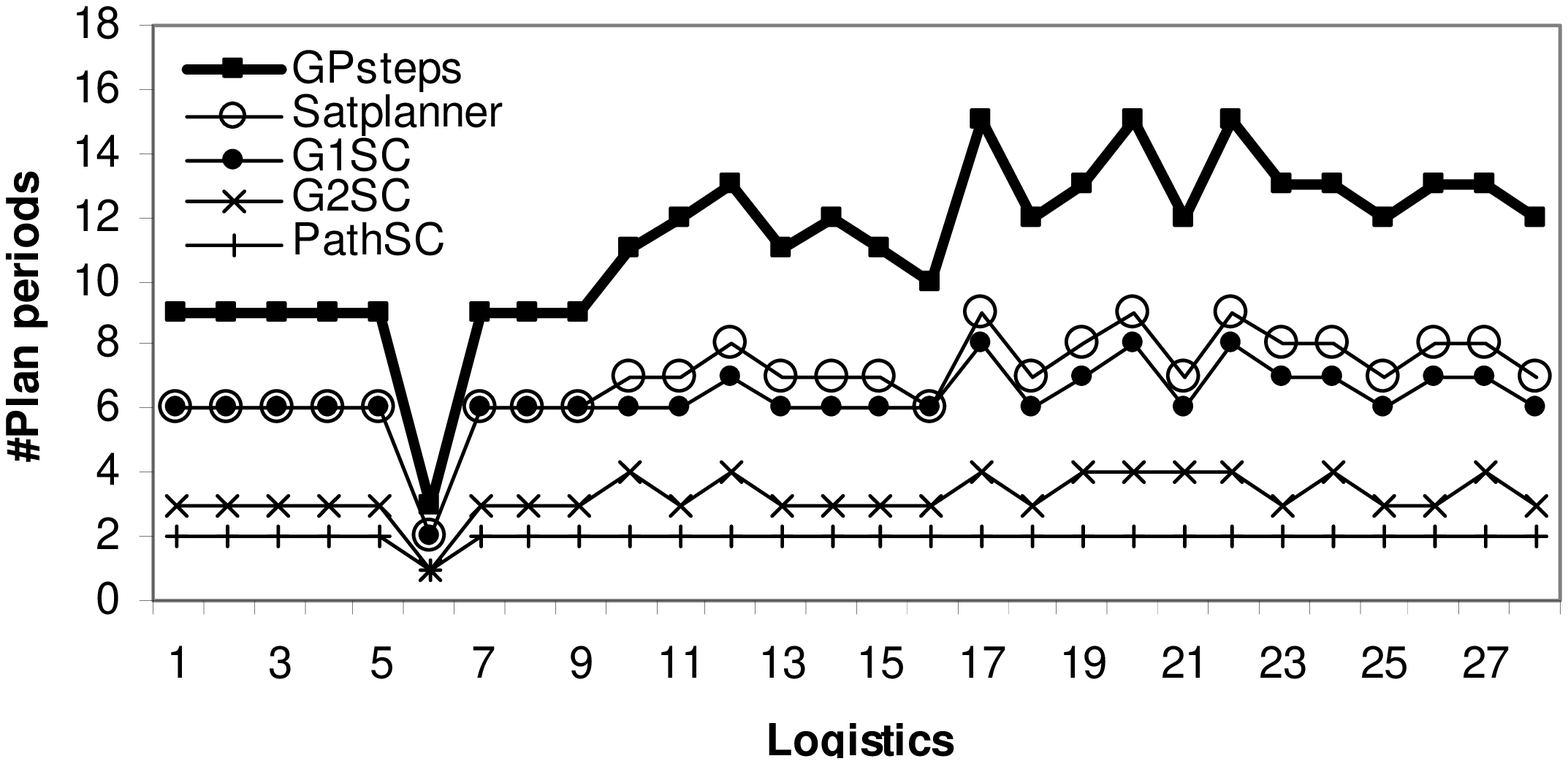} \\
  \includegraphics[width=2.9in]{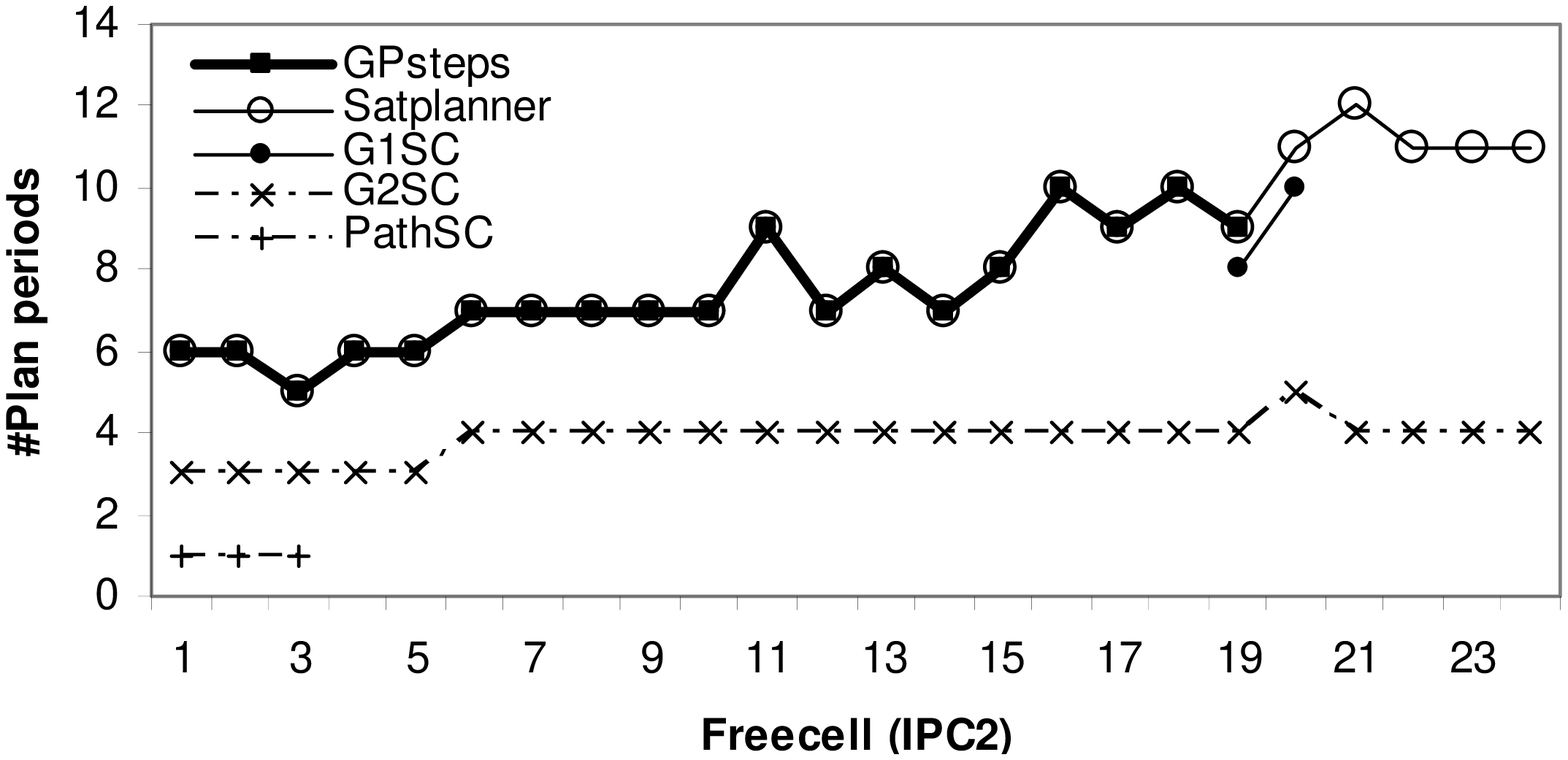} &
  \includegraphics[width=2.9in]{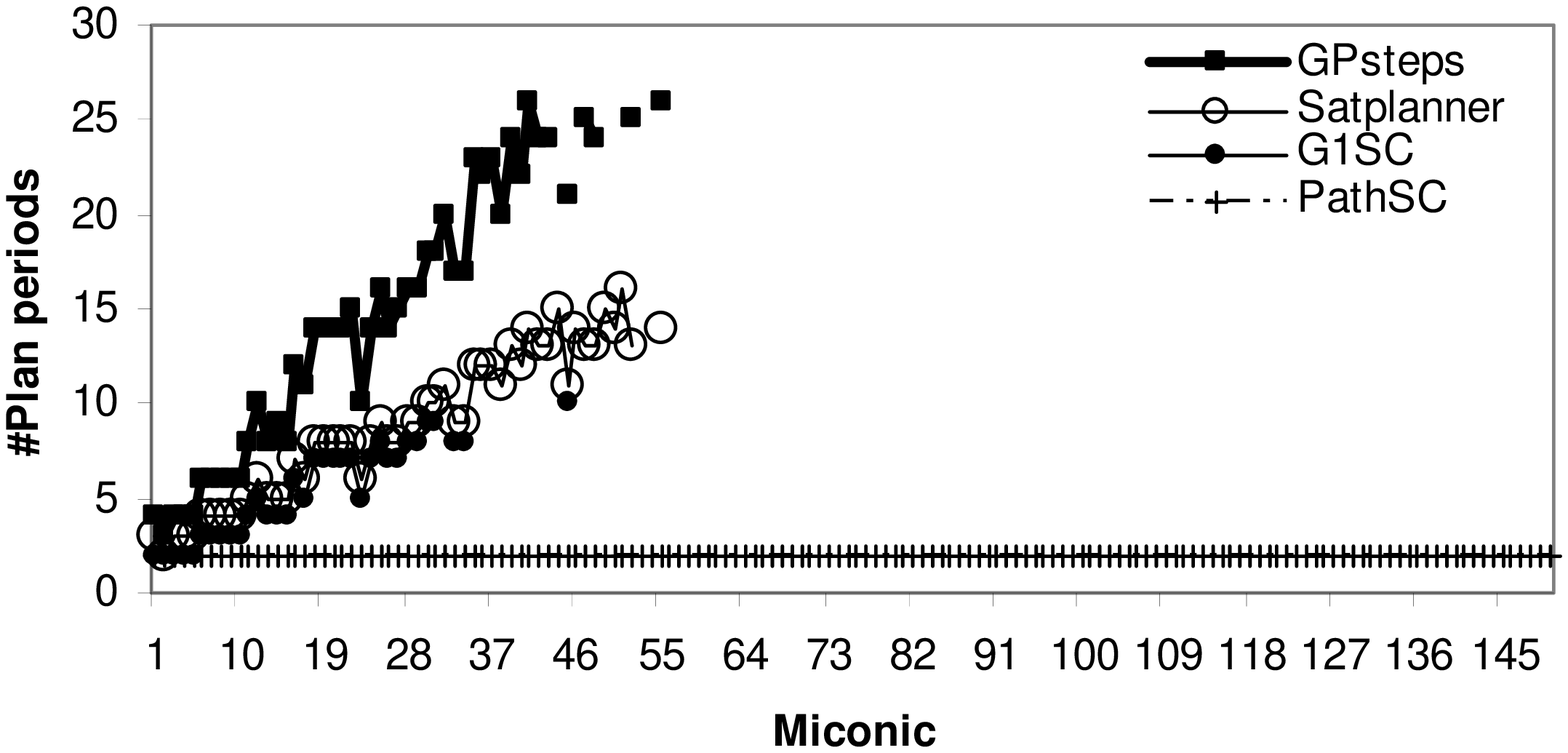} \\
  \includegraphics[width=2.9in]{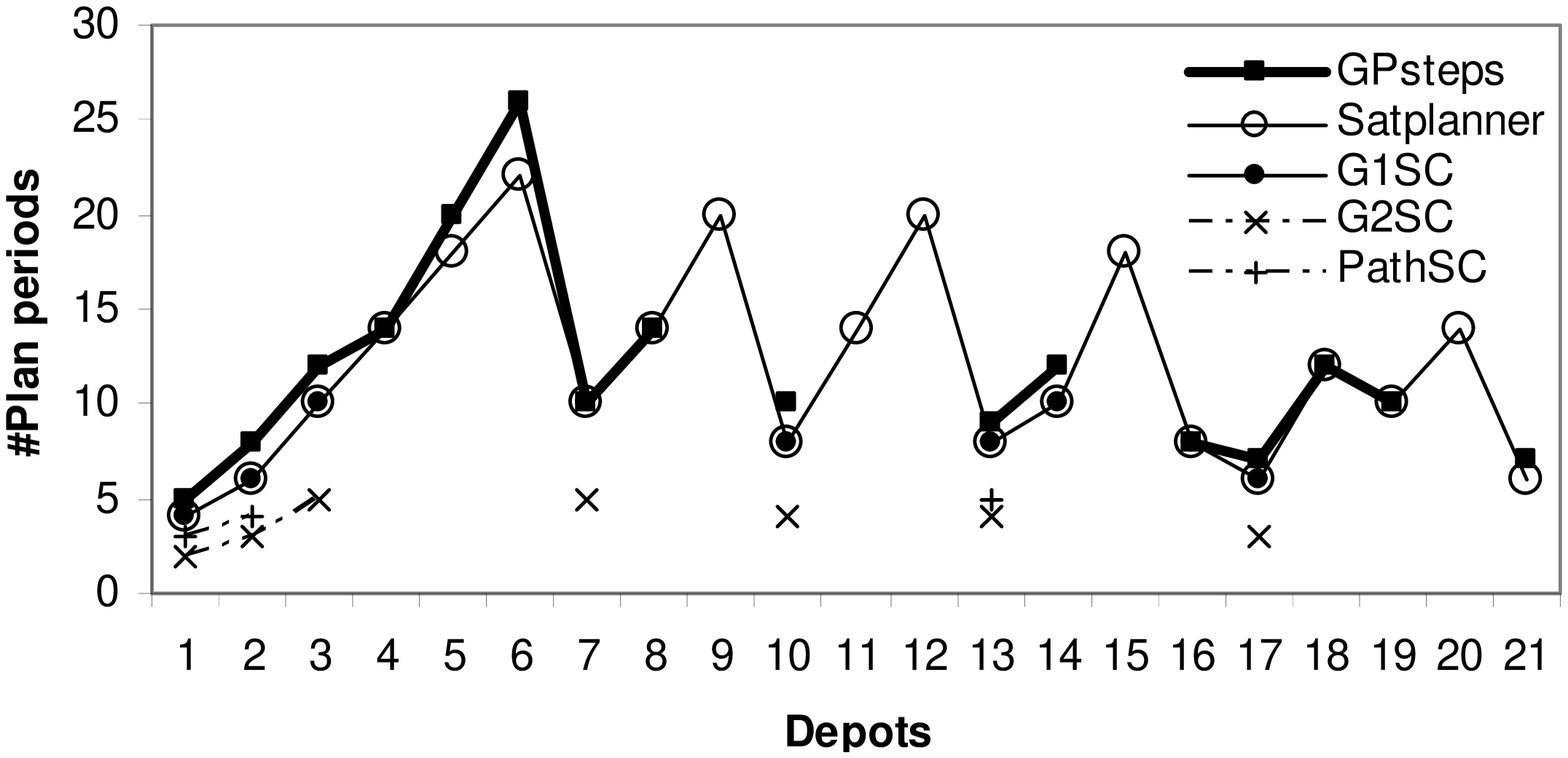} &
  \includegraphics[width=2.9in]{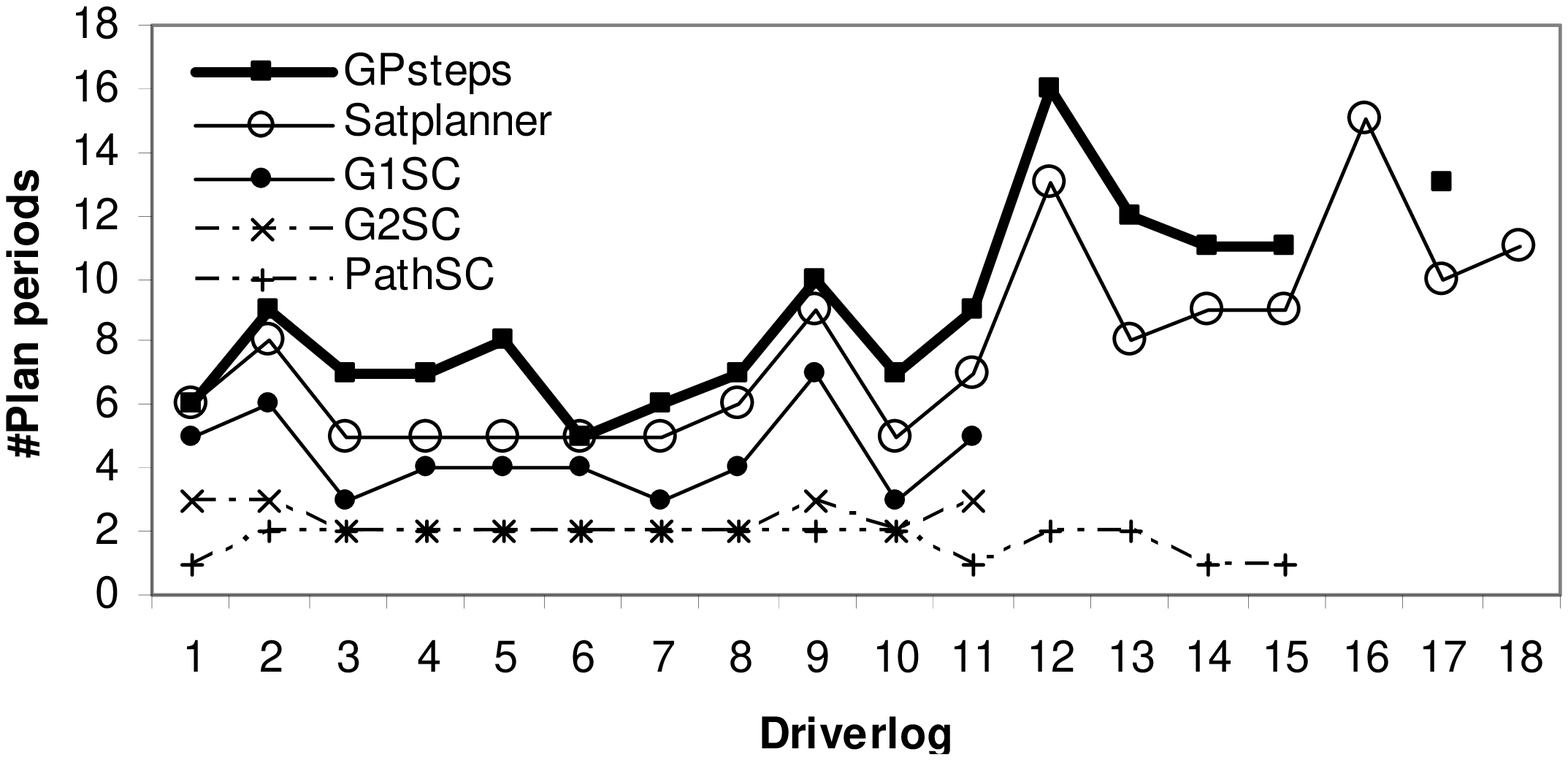} \\
  \includegraphics[width=2.9in]{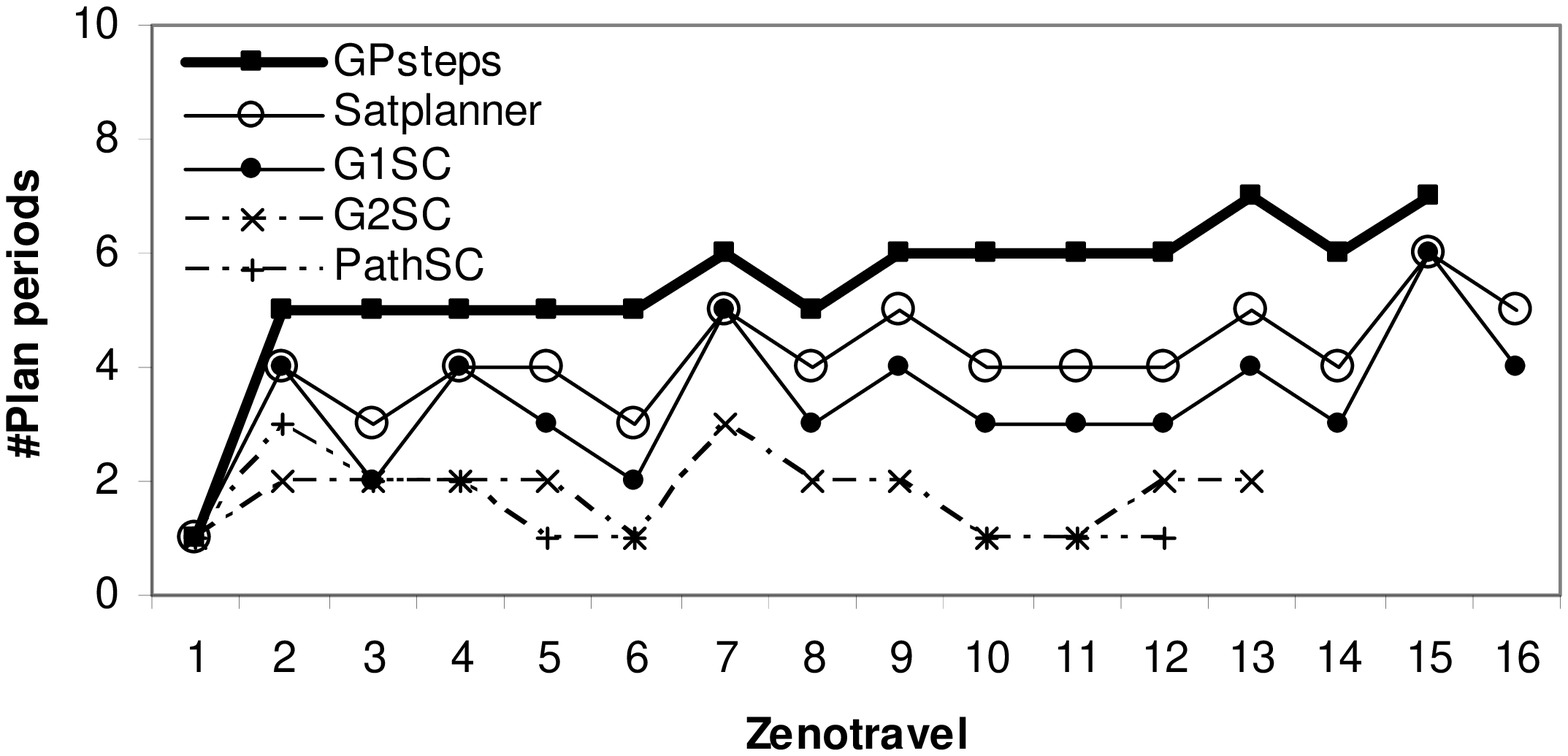} &
  \includegraphics[width=2.9in]{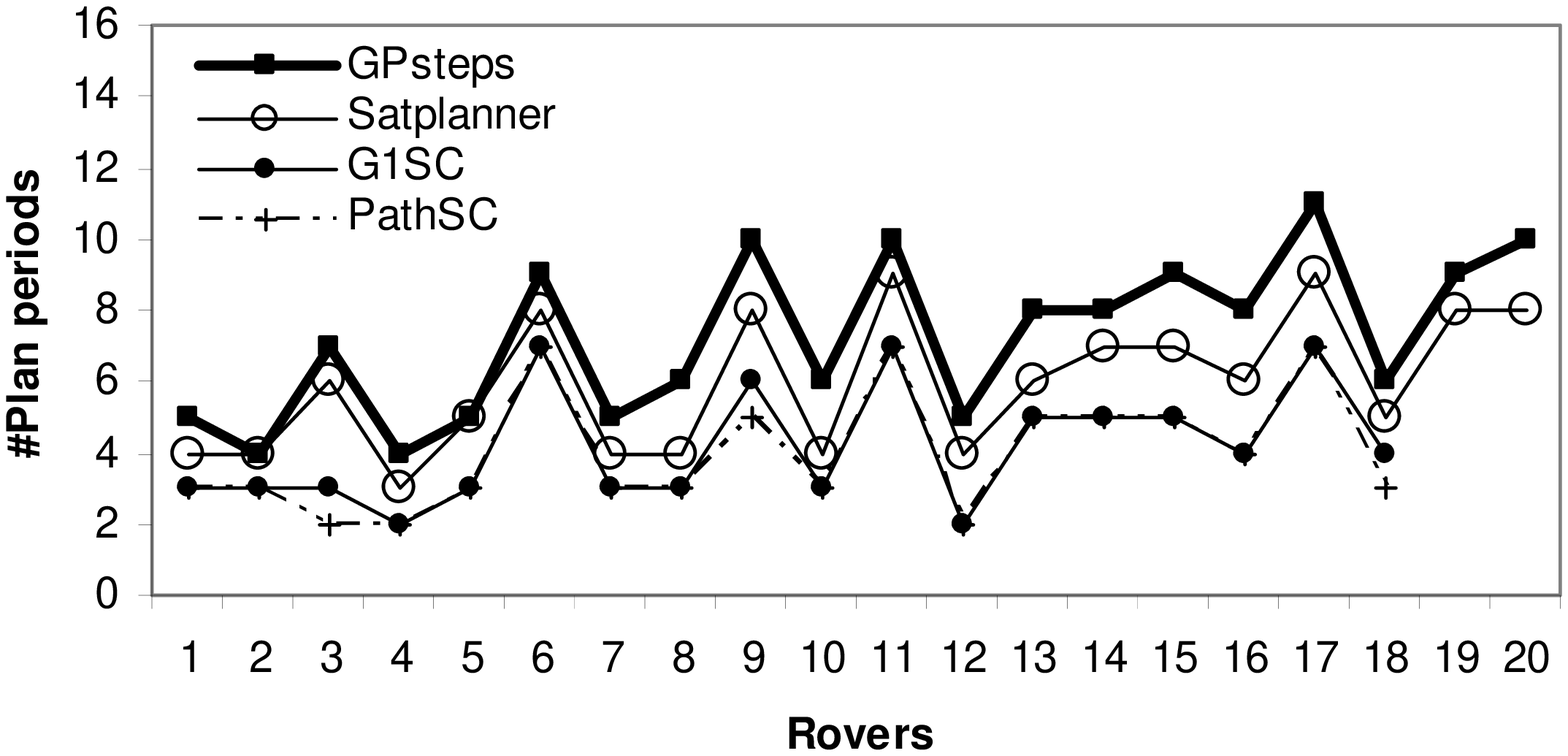} \\
  \includegraphics[width=2.9in]{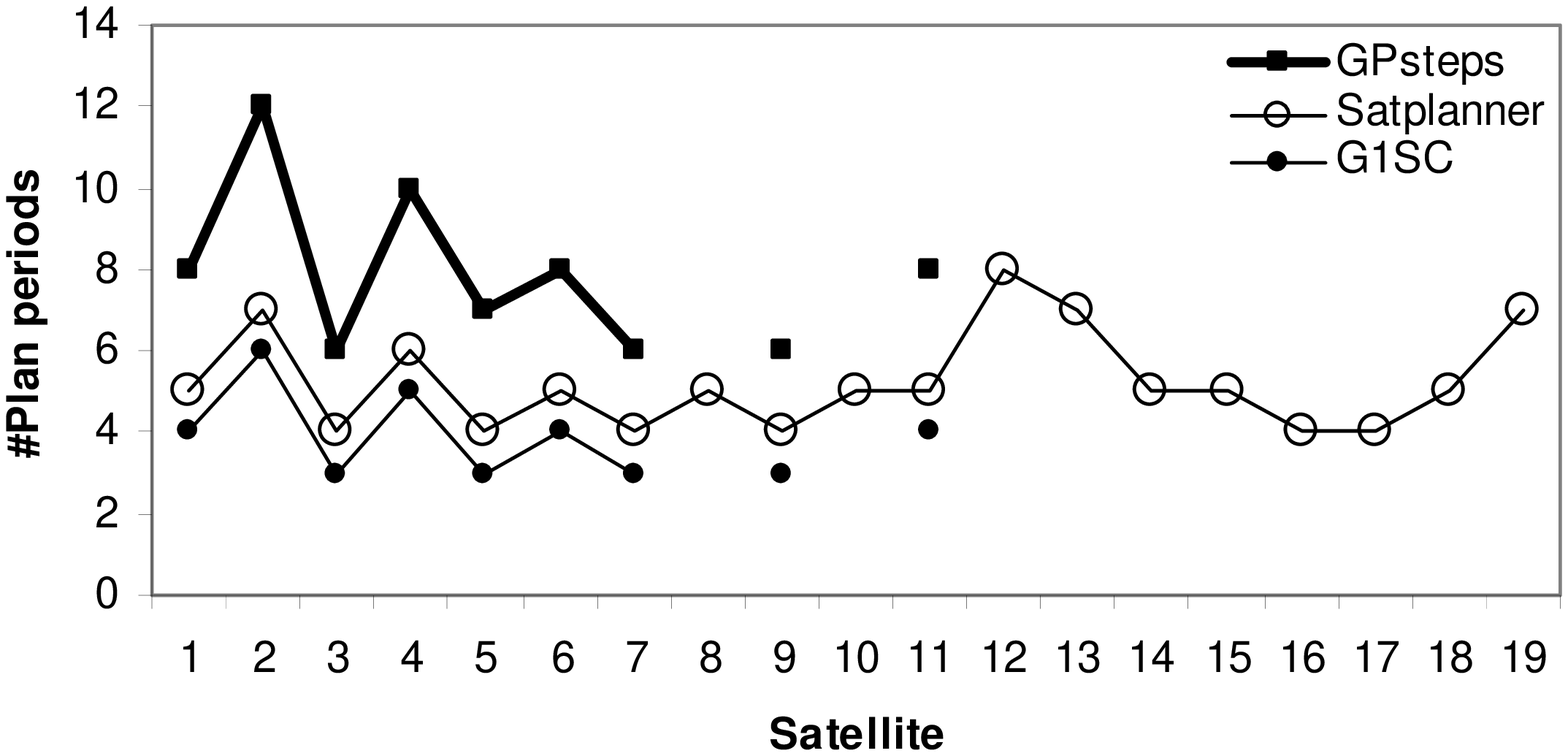} &
  \includegraphics[width=2.9in]{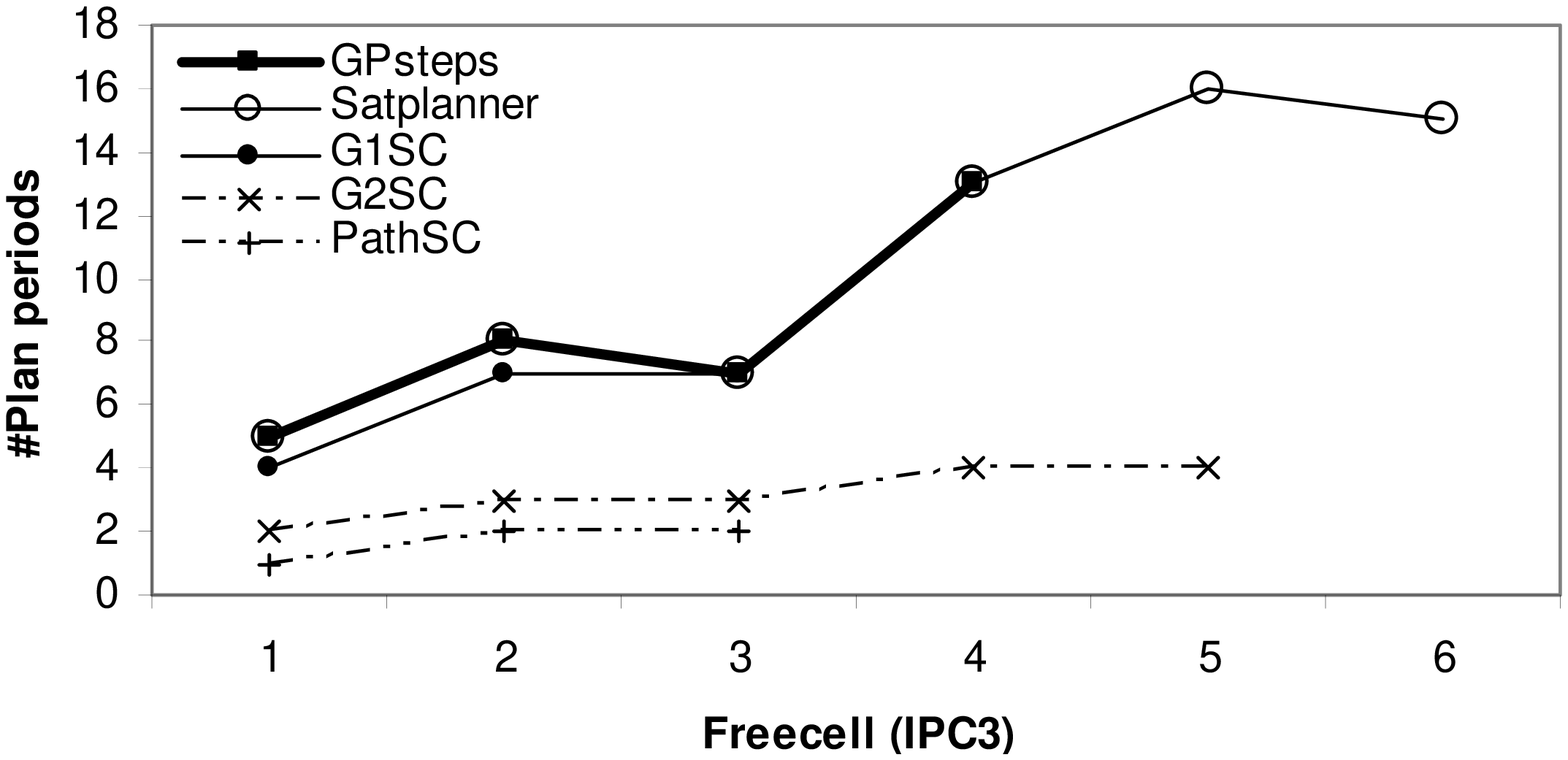}
\end{tabular}
\caption{Number of plan periods required for each formulation to solve the planning problems in the second and third international planning competition.} \label{fig:ipcsteps}
\end{figure*}

In the Logistics domain all formulations that generalize the Graphplan-style parallelism (i.e.\ G1SC, G2SC, and PathSC) scale better than the 1SC formulation and Optiplan, which adopt the Graphplan-style parallelism. Among G1SC, G2SC, and PathSC formulations there is no clear best performer, but in the larger Logistics problems the G1SC formulation seems to do slightly better. The Logistics domain provides a great example of the tradeoff between flexibility and merging. By allowing more actions to be executed in each plan period, generally shorter plans (in terms of number of plan periods) are needed to solve the planning problem (see Figure \ref{fig:ipcsteps}), but at the same time merging the solutions to the individual components will be harder as one has to respect more ordering constraints (see Table \ref{tab:ipccuts}).

\textbf{Optiplan versus 1SC. } If we compare the 1SC formulation with Optiplan, we note that Optiplan fares well in domains that are either serial by design (Blocksworld) or in domains that have a significant serial aspect (Depots). We think that Optiplan's advantage over the 1SC formulation in these domains is due to the following two possibilities. First, our intuition is that in serial domains the reachability and relevance analysis in Graphplan is stronger in detecting infeasible action choices (due to mutex propagation) than the network flow restrictions in the 1SC formulation. Second, it appears that the state variables in these domains are more tightly coupled (i.e.\ the actions have more effects, thus transitions in one state variable are coupled with several transitions in other state variables) than in most other domains, which may negatively affect the performance of the 1SC formulation.

\begin{figure*}
\centering
\begin{tabular}{cc}
  \includegraphics[width=2.9in]{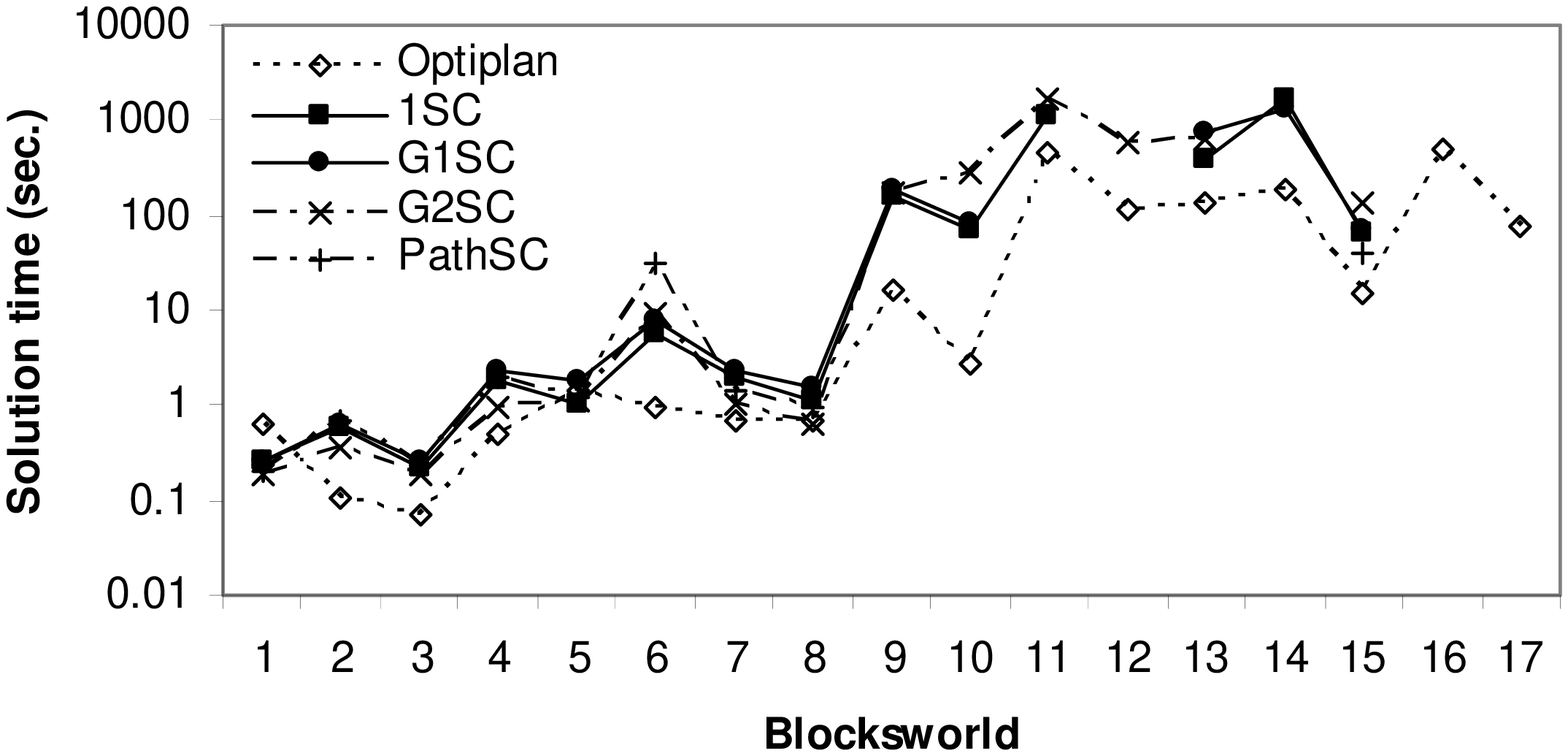} &
  \includegraphics[width=2.9in]{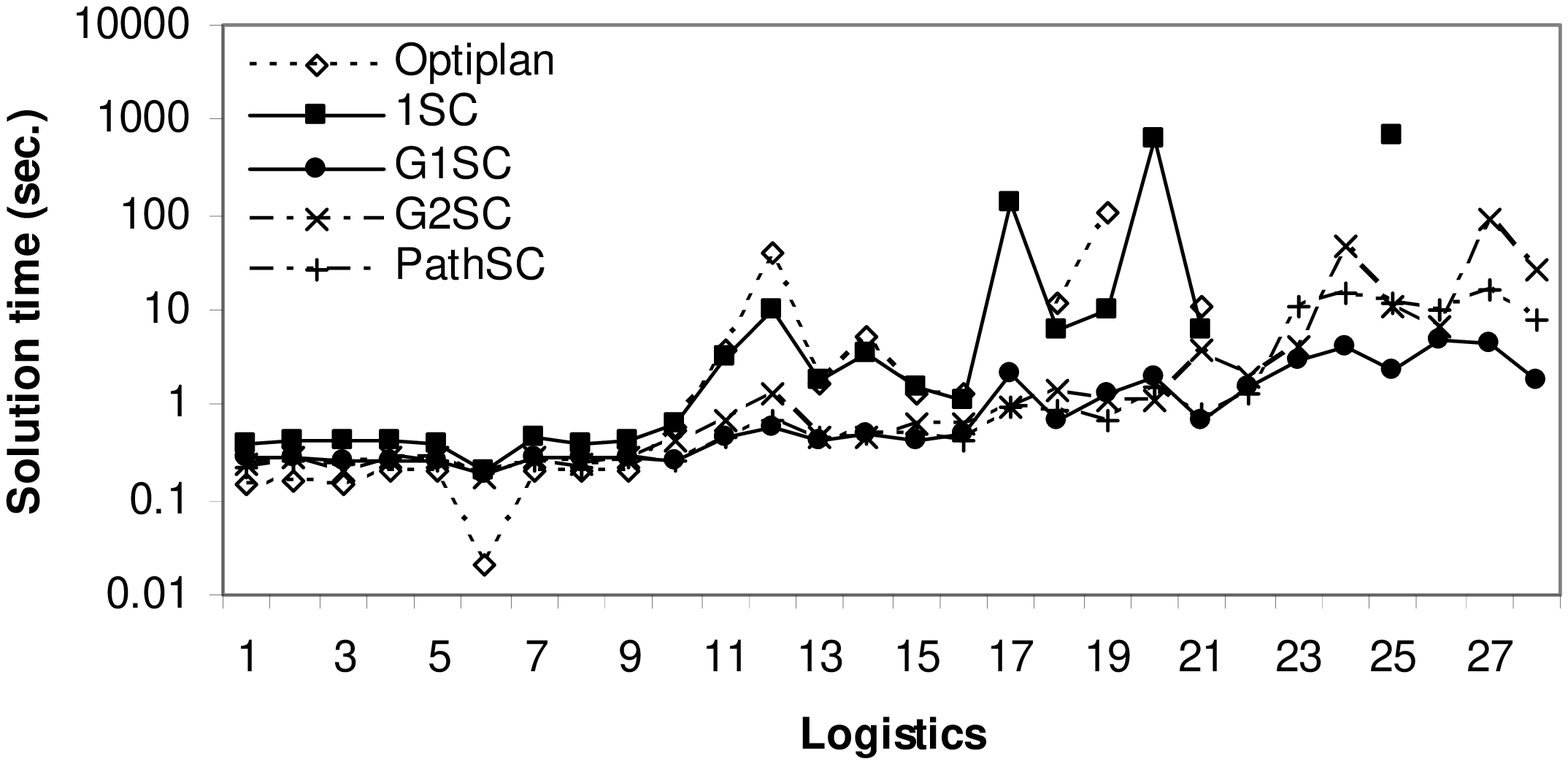} \\
  \includegraphics[width=2.9in]{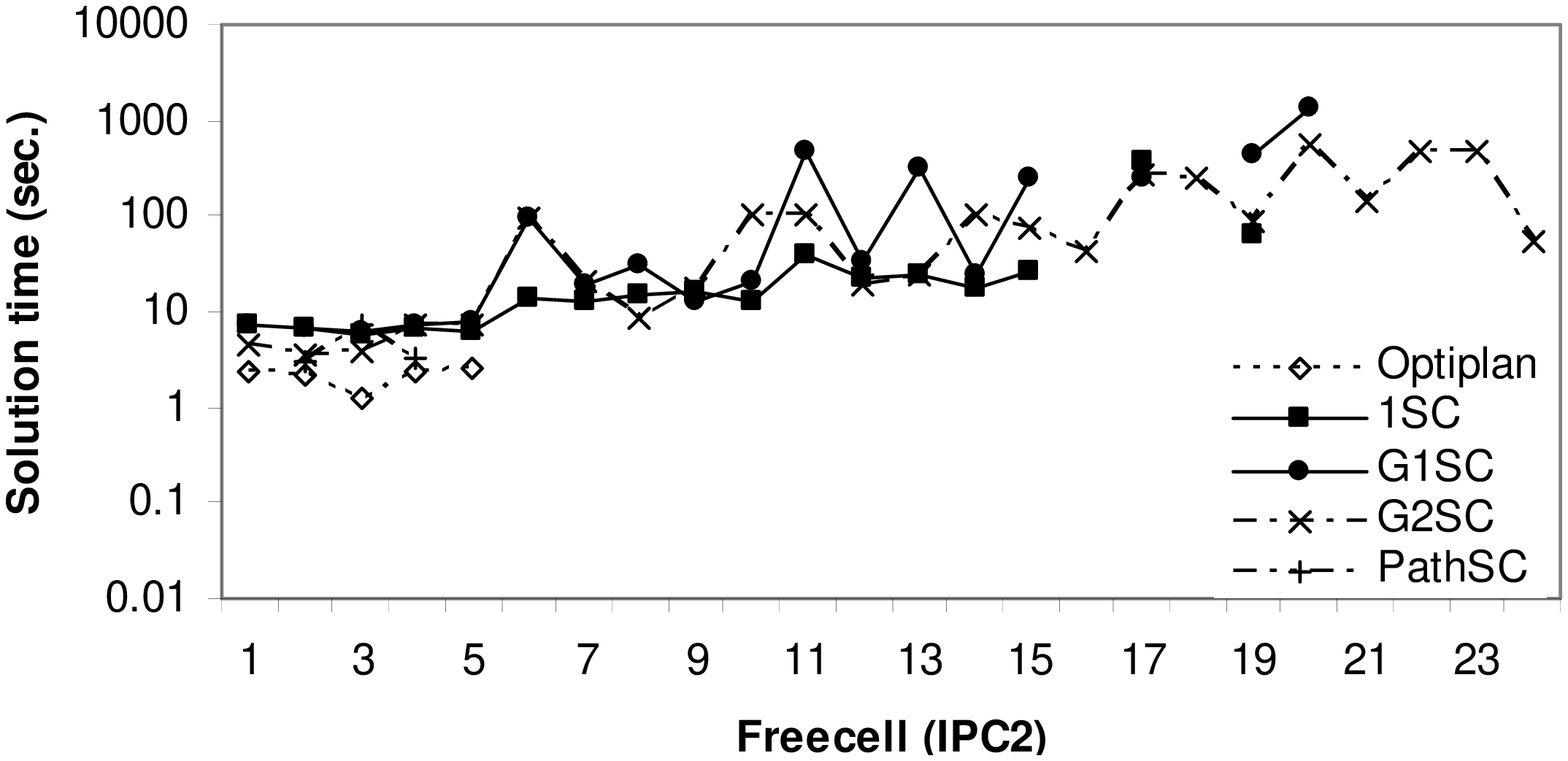} &
  \includegraphics[width=2.9in]{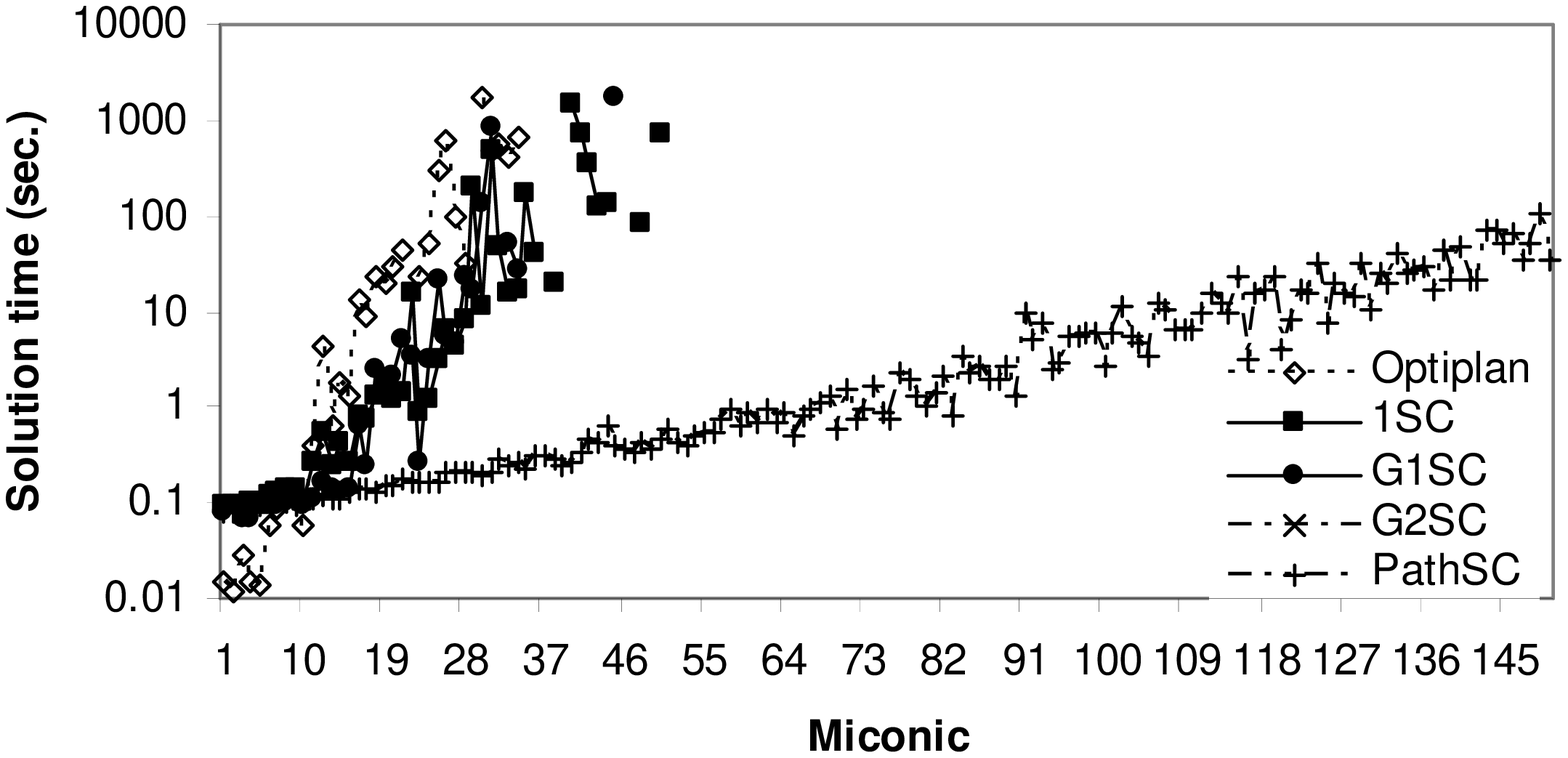} \\
  \includegraphics[width=2.9in]{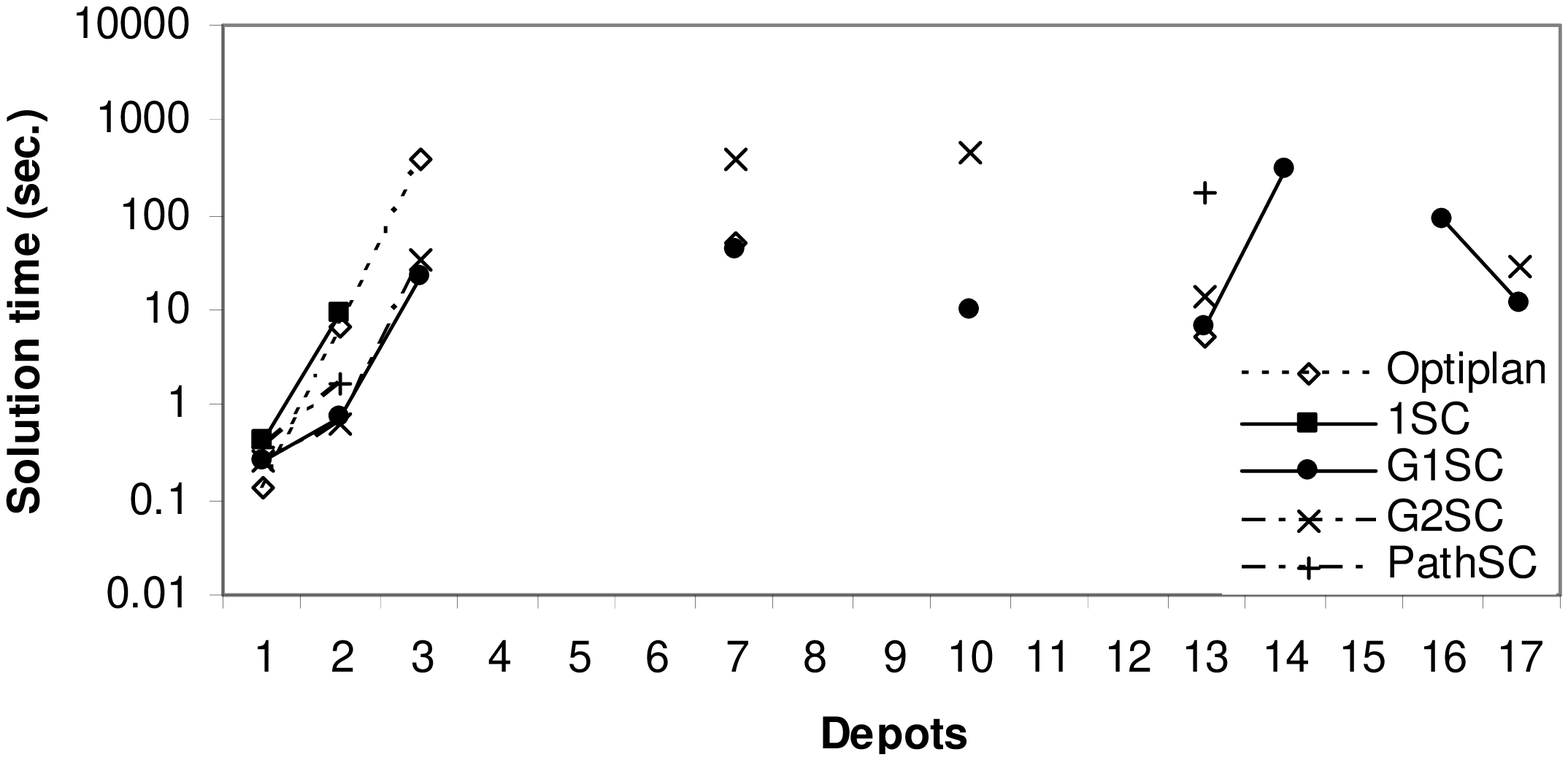} &
  \includegraphics[width=2.9in]{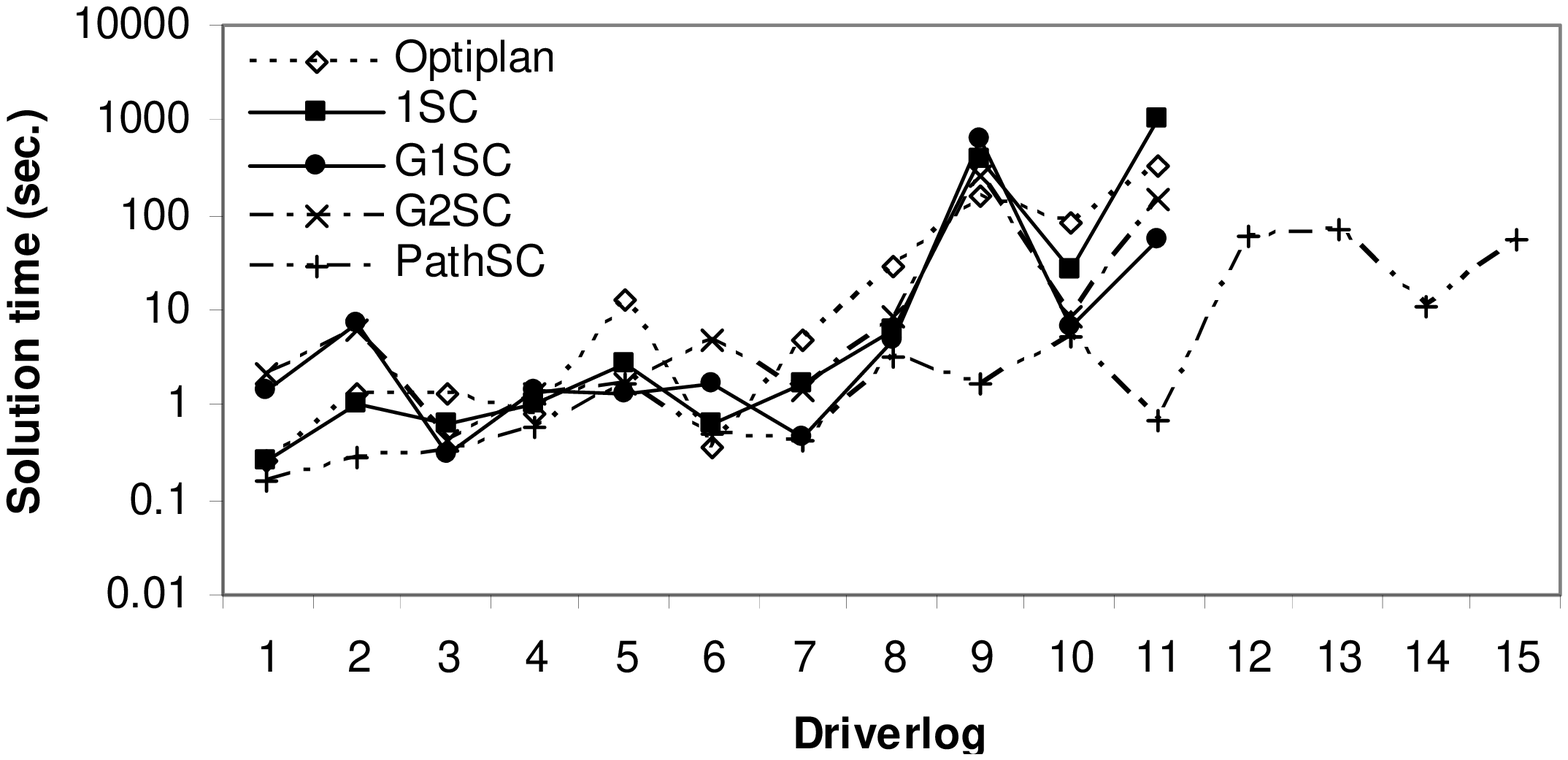} \\
  \includegraphics[width=2.9in]{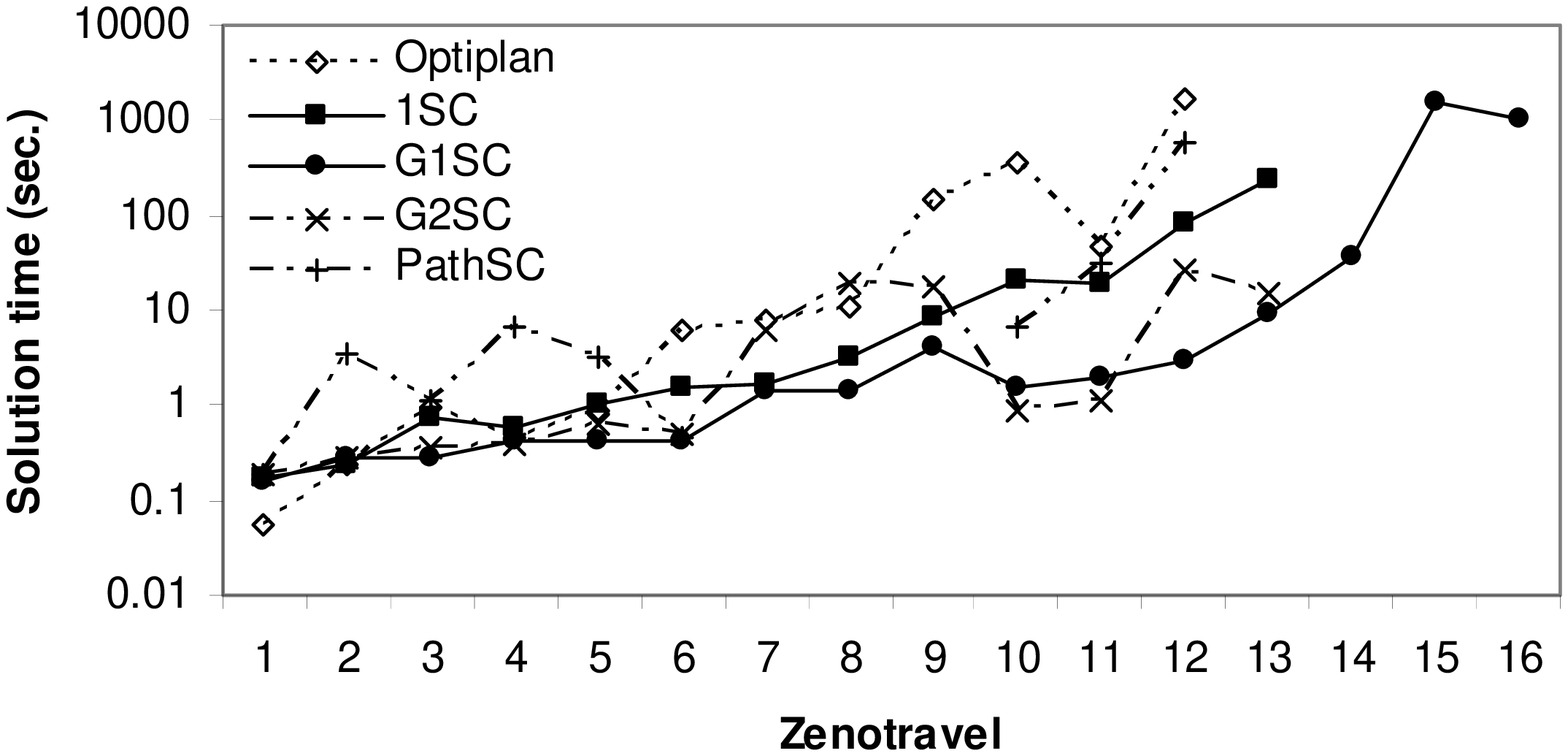} &
  \includegraphics[width=2.9in]{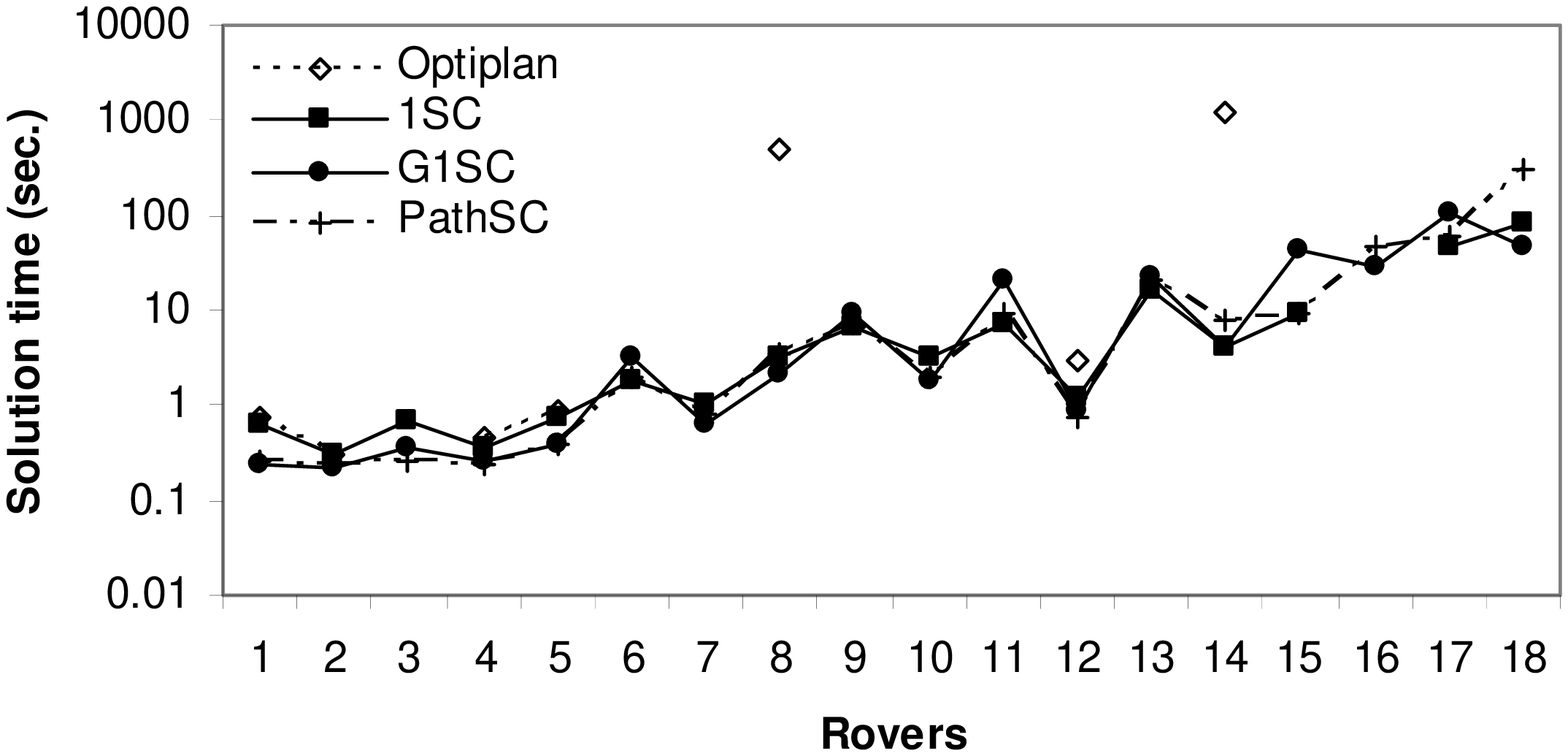} \\
  \includegraphics[width=2.9in]{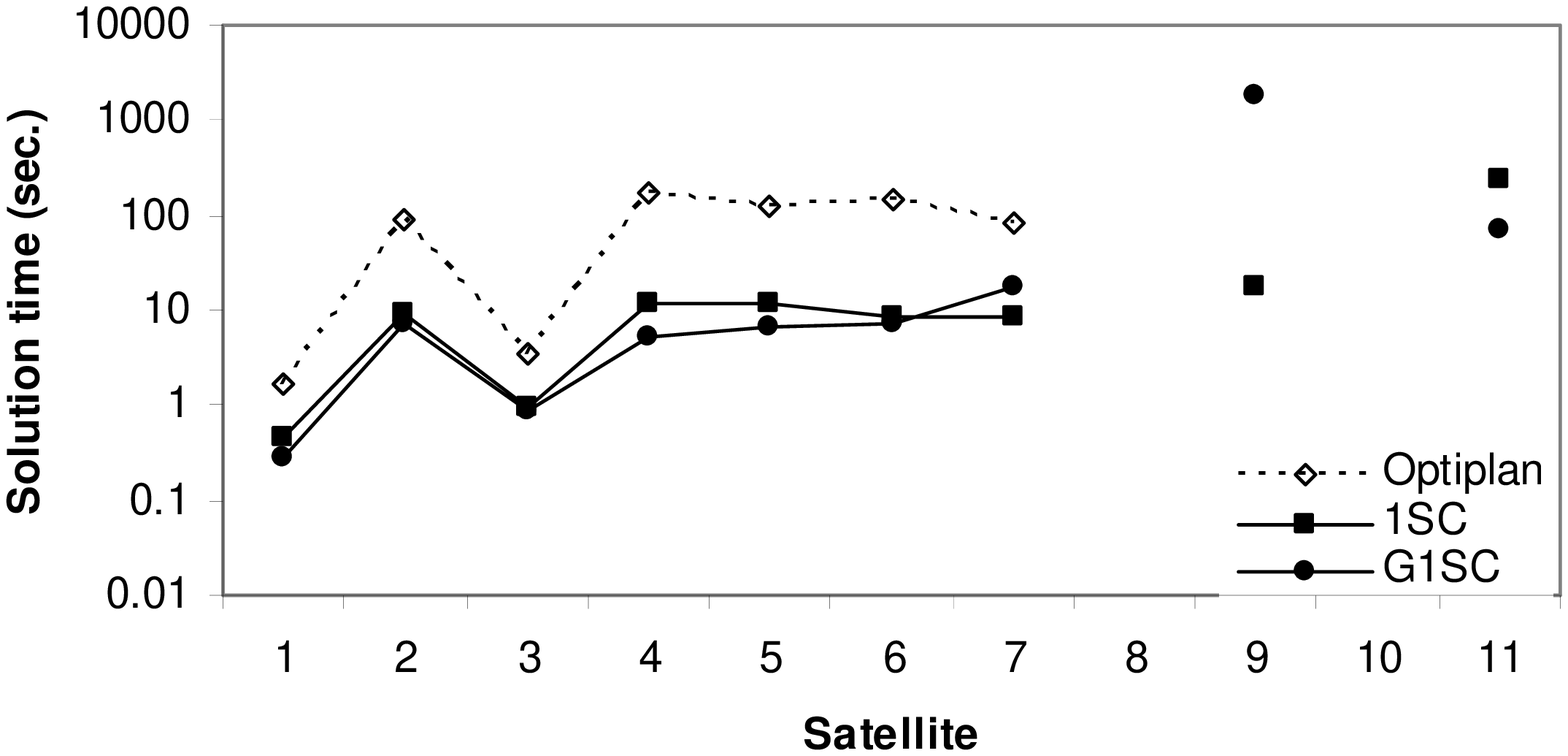} &
  \includegraphics[width=2.9in]{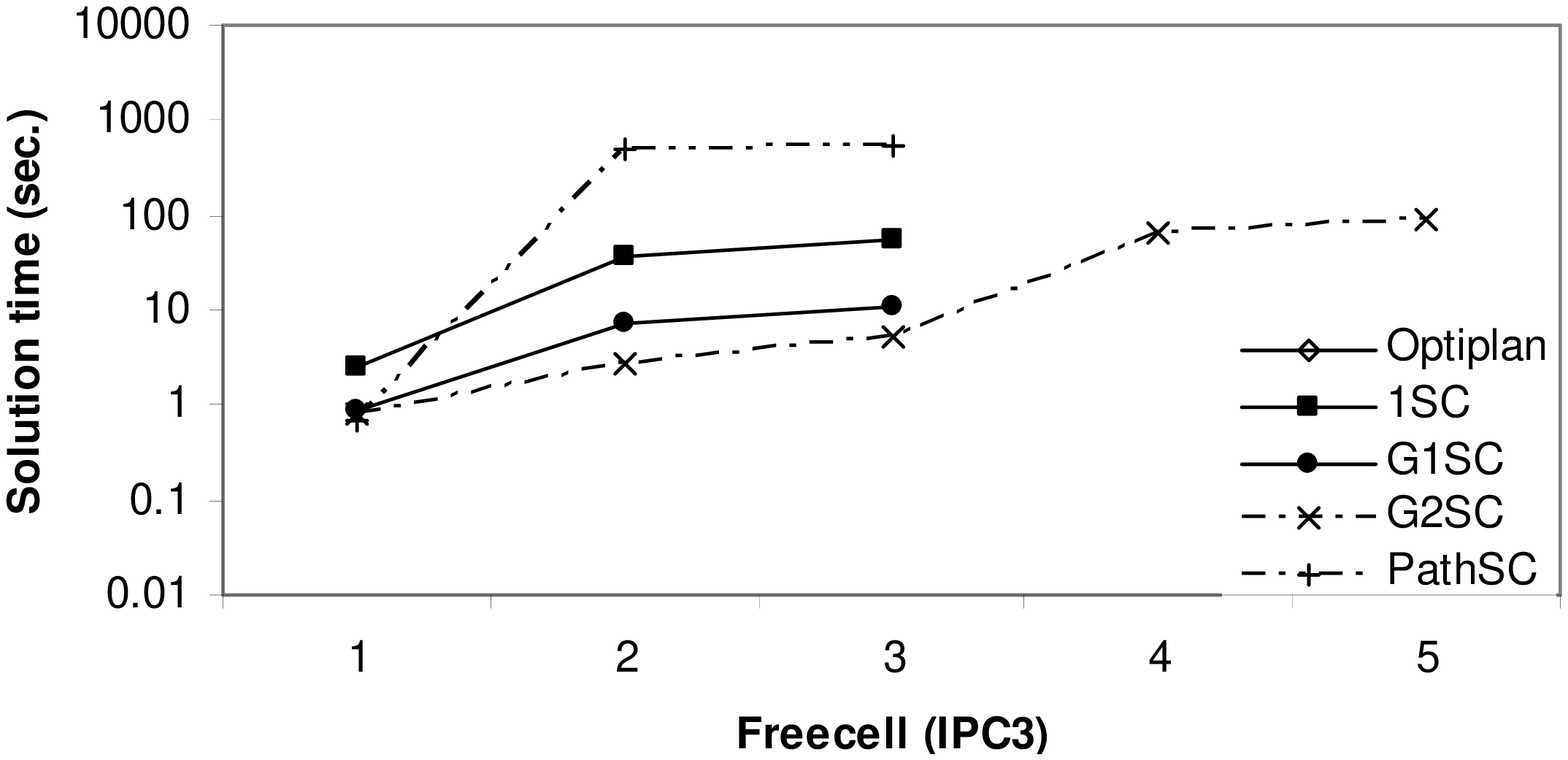}
\end{tabular}
\caption{Solution times in the planning domains of the second and third international planning competition.} \label{fig:ipctime}
\end{figure*}

\textbf{1SC versus G1SC. } When comparing the 1SC formulation with the G1SC formulation we can see that in all domains, except in Blocksworld and Miconic, the G1SC formulation solves at least as many problems as the 1SC formulation. The results in Blocksworld are not too surprising and can be attributed to semantics of this domain. Each operator in Blocksworld requires one state change in the state variable of the arm ($stack$ and $putdown$ change the status of the arm to $arm-empty$, and $unstack$ and $pickup$ change the status of the arm to $holding-x$ where $x$ is the block being lifted). Since, the 1SC and the G1SC formulations both allow at most one state change in each state variable, there is no possibility for the G1SC formulation to allow more than one action to be executed in the same plan period. Given this, one may think that the 1SC and G1SC formulations should solve at least the same number of problems, but in this case the prevail constraints \eqref{eq:1sc prevail} of the 1SC formulation are stronger than the prevail constraints \eqref{eq:g1sc prevail} of the G1SC formulation. That is, the right-hand side of \eqref{eq:g1sc prevail} subsumes (i.e.\ allows for a larger feasible region in the LP relaxation) than the right-hand side of \eqref{eq:1sc prevail}. In Figure \ref{fig:ipctime} we can see this slight advantage of 1SC over G1SC in the Blocksworld domain.

The results in the Miconic domain are, on the other hand, not very intuitive. We would have expected the G1SC formulation to solve at least as many problems as the 1SC formulation, but this did not turn out to be the case. One thing we noticed is that in this domain the G1SC formulation required a lot more time to determine that there is no plan for a given number of plan periods.

\textbf{G1SC versus G2SC and PathSC. } Table \ref{tab:ipccuts} only shows the five largest problems in each domain that were solved by the formulations, yet it is representative for the whole set of problems. The table indicates that when Graphplan-style parallelism is generalized, more ordering constraints are needed to ensure a feasible plan. On average, the G2SC formulation includes more ordering constraints than the G1SC formulation, and the PathSC formulation in its turn includes more ordering constraints than the G2SC formulation. The performance of these formulations as shown by Figure \ref{fig:ipctime} varies per planning domain. The PathSC formulation does well in Miconic and Driverlog, the G2SC formulation does well in Freecell, and the G1SC does well in Zenotravel. Because of these performance differences, we believe that the ideal amount of flexibility in the generalization of Graphplan-style parallelism is different for each planning domain.

\begin{table}[p]
\begin{center}
{\small
\begin{tabular}{ccc}
\begin{tabular}{l|rrr}
  \hline
  Problem & {G1SC} & {G2SC} & {PathSC} \\
  \hline
  Blocksworld &    &    &    \\
  5-1	      &  0 &  0 &  16\\
  5-2	      &  0 &  0 &  62\\
  6-0	      &  0 &  0 &   5\\
  6-1         &  0 &  0 &   5\\
  8-2         &  0 &  0 &  62\\
  \hline
  Logistics   &    &    &    \\
  10-1        &  0 &  7 &  11\\
  11-0        &  0 &  0 &  10\\
  11-1        &  0 &  0 &  49\\
  12-0        &  0 & 16 &   9\\
  14-0        &  0 &  7 & 110\\
  \hline
  Freecell    &    &    &    \\
  2-1         &  0 &  0 &   0\\
  2-2         &  0 &  0 &   0\\
  2-3         &  0 &  0 &  84\\
  2-4         &  0 &  2 &   0\\
  5-5         &  0 &  3 &   *\\
  \hline
  Miconic     &    &    &    \\
  6-4         &  0 &  - &   0\\
  7-0         &  0 &  - &   0\\
  7-2         &  0 &  - &   2\\
  7-3         &  0 &  - &   4\\
  9-4         &  0 &  - &   5\\
  \hline
  \multicolumn{4}{c}{\ } \\
  \multicolumn{4}{c}{\ } \\
  \multicolumn{4}{c}{\ } \\
  \multicolumn{4}{c}{\ } \\
  \multicolumn{4}{c}{\ } \\
  \multicolumn{4}{c}{\ } \\
  \multicolumn{4}{c}{\ } \\
  \multicolumn{4}{c}{\ } \\
  \multicolumn{4}{c}{\ } \\
  \multicolumn{4}{c}{\ } \\
  \multicolumn{4}{c}{\ } \\
  \multicolumn{4}{c}{\ }
\end{tabular}
& \ \ \ \ &
\begin{tabular}{l|rrr}
  \hline
  Problem & {G1SC} & {G2SC} & {PathSC} \\
  \hline
  Depots      &    &    &    \\
  1	          &  0 &  0 &   7\\
  2	          &  0 &  0 &   2\\
  10          &  0 &  2 &   *\\
  13          &  0 &  0 &  30\\
  17          &  0 &  0 &   *\\
  \hline
  Driverlog   &    &    &    \\
  7           &  1 & 16 &   2\\
  8           & 32 & 62 & 108\\
  9           & 58 & 80 &  32\\
  10          &  5 &  5 &  69\\
  11          &  6 & 30 &  11\\
  \hline
  Zenotravel  &    &    &    \\
  5           &  0 &  4 & 485\\
  6           &  0 &  5 &   6\\
  10          &  0 &  0 & 214\\
  11          &  0 &  0 & 586\\
  12          &  0 & 60 &6259\\
  \hline
  Rovers      &    &    &    \\
  14          &  0 &  - &  13\\
  15          & 12 &  - &  11\\
  16          &  1 &  - &  92\\
  17          &  1 &  - &   4\\
  18          &  1 &  - & 192\\
  \hline
  Satellite   &    &    &    \\
  5           &  2 &  - &   -\\
  6           &  6 &  - &   -\\
  7           &  8 &  - &   -\\
  9           & 14 &  - &   -\\
  11          &  0 &  - &   -\\
  \hline
  Freecell    &    &    &    \\
  1           &  0 &  0 &   3\\
  2           &  0 &  0 &2480\\
  3           &  0 &  1 &1989\\
  4           &  * &  1 &   *\\
  5           &  * &  0 &   *\\
  \hline
\end{tabular}
\end{tabular}
}
\end{center}
\caption{Number of ordering constraints, or cuts, that were added dynamically through the solution process to problems in IPC2 (left) and IPC3 (right). A dash `-' indicates that the IP formulation could not be tested on the domain and a star `*' indicates that the formulation could not solve the problem instance within 30 minutes.} \label{tab:ipccuts}
\end{table}

\subsubsection{Results: Number of Plan Periods}\label{sec:results2}
In Figure \ref{fig:ipcsteps}, we see that in all domains the flexible network representation of the G1SC formulation is slightly more general than the 1-linearization semantics that is used by Satplanner. That is, the number of plan periods required by the G1SC formulation is always less than or equal to the number of plan periods used by Satplanner. Moreover, the flexible network representation of the G2SC and PathSC formulations are both more general than the one used by the G1SC formulation. One may think that the network representation of the PathSC formulation should provide the most general interpretation of action parallelism, but since the G2SC network representations allows some values to change back to their original value in the same plan period this is not always the case.

In the domains of Logistics, Freecell, Miconic, and Driverlog, the PathSC never required more than two plan periods to solve the problem instances. For the Miconic domain this is very easy to understand. In Miconic there is an elevator that needs to bring travelers from one floor to another. The state variables representation of this domain has one state variable for the elevator and two for each traveler (one to represent whether the traveler has boarded the elevator and one to represent whether the traveler has been serviced). Clearly, one can devise a plan such that each value of the state variable is visited at most twice. The elevator simply could visit all floors and pickup all the travelers, and then visit all floors again to bring them to their destination floor.

\subsection{Comparing Different State Variable Representations}\label{sec:results3}
An interesting question is to find out whether different state variable representations lead to different performance benefits. In our loosely coupled formulations we have components that represent multi-valued state variables. However, the idea of modeling value transitions as flows in an appropriately defined network can be applied to any binary or multi-valued state variable representation. In this section we concentrate on the efficiency tradeoffs between binary and multi-valued state descriptions. As there are generally fewer multi-valued state variables than binary atoms needed to describe a planning problem, we can expect our formulations to be more compact when they use a multi-valued state description. For this comparison we only concentrate on the G1SC formulation as it showed the overall best performance among our formulations. In our recent work \cite{BRIetal2007B} we analyze different state variable representations in more detail.

Table \ref{tab:binmul} compares the encoding size for the G1SC formulation on a set of problems using either a binary or multi-valued state description. The table clearly shows that the encoding size becomes significantly smaller (both before and after CPLEX presolve) when a multi-valued state description is used. The encoding size before presolve gives an idea of the impact of using a more compact multi-valued state description, whereas the encoding size after presolve shows how much preprocessing can be done by removing redundancies and substituting out variables.

\begin{figure*}
\centering
\begin{tabular}{cc}
  \includegraphics[width=2.9in]{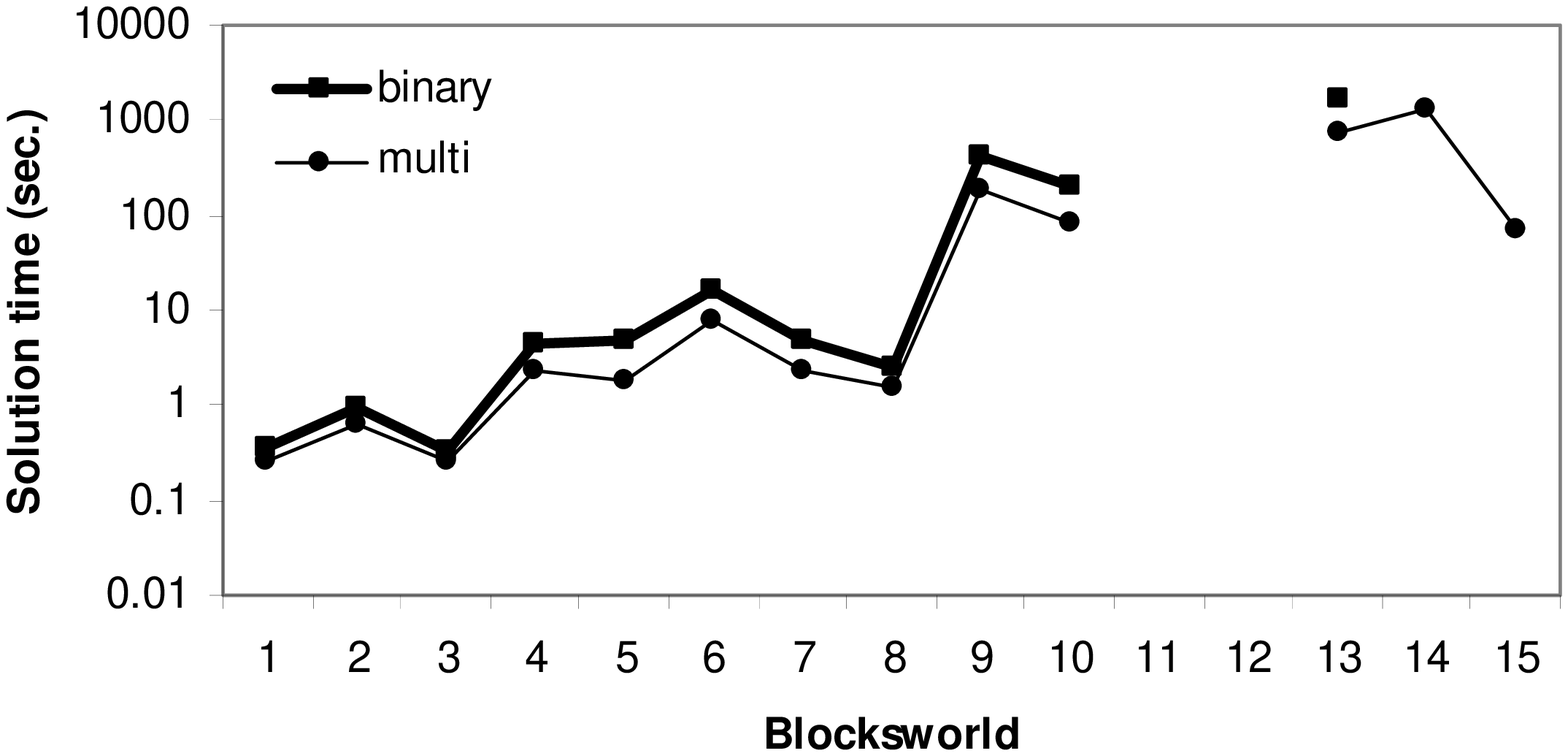} &
  \includegraphics[width=2.9in]{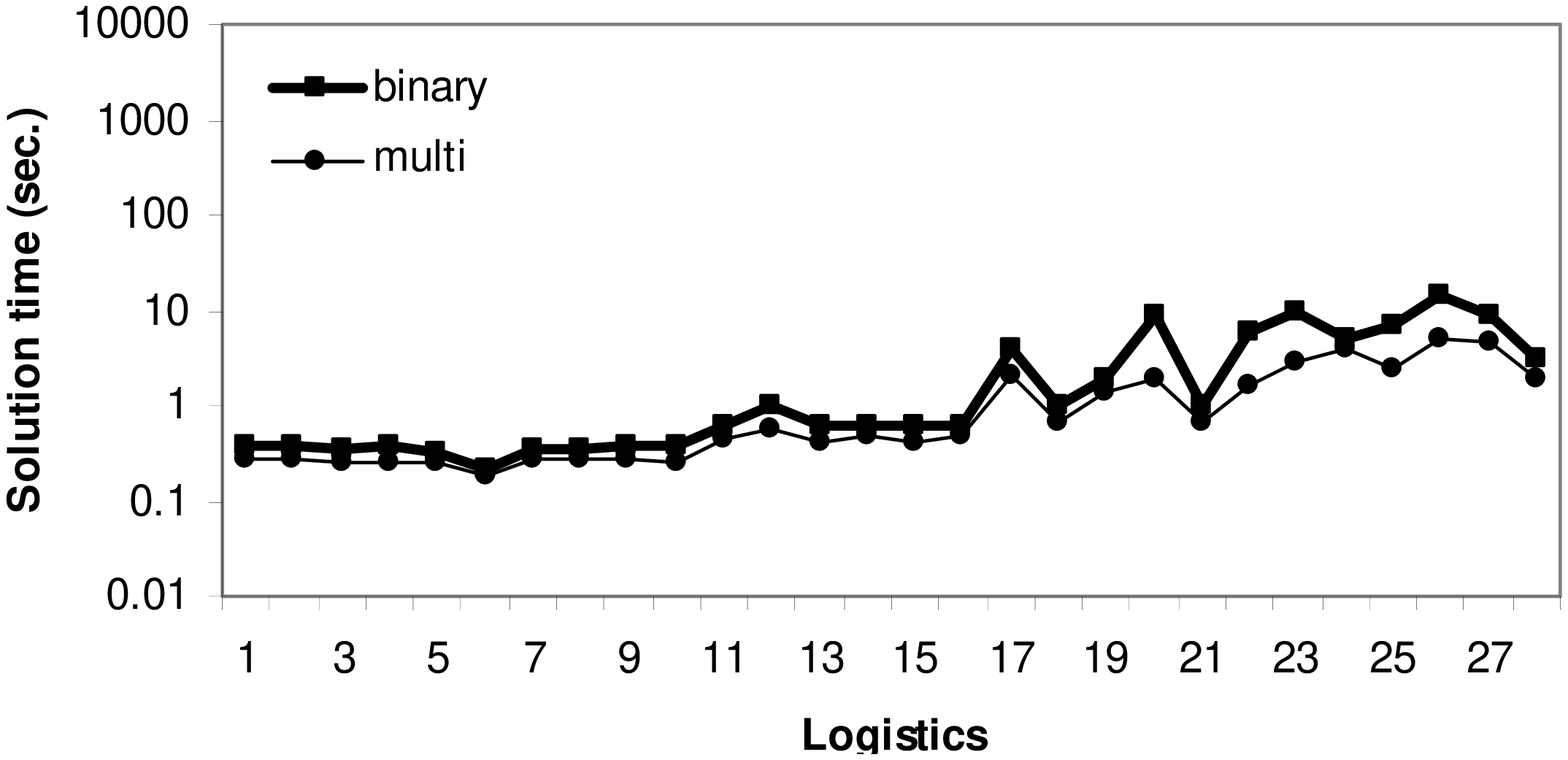} \\
  \includegraphics[width=2.9in]{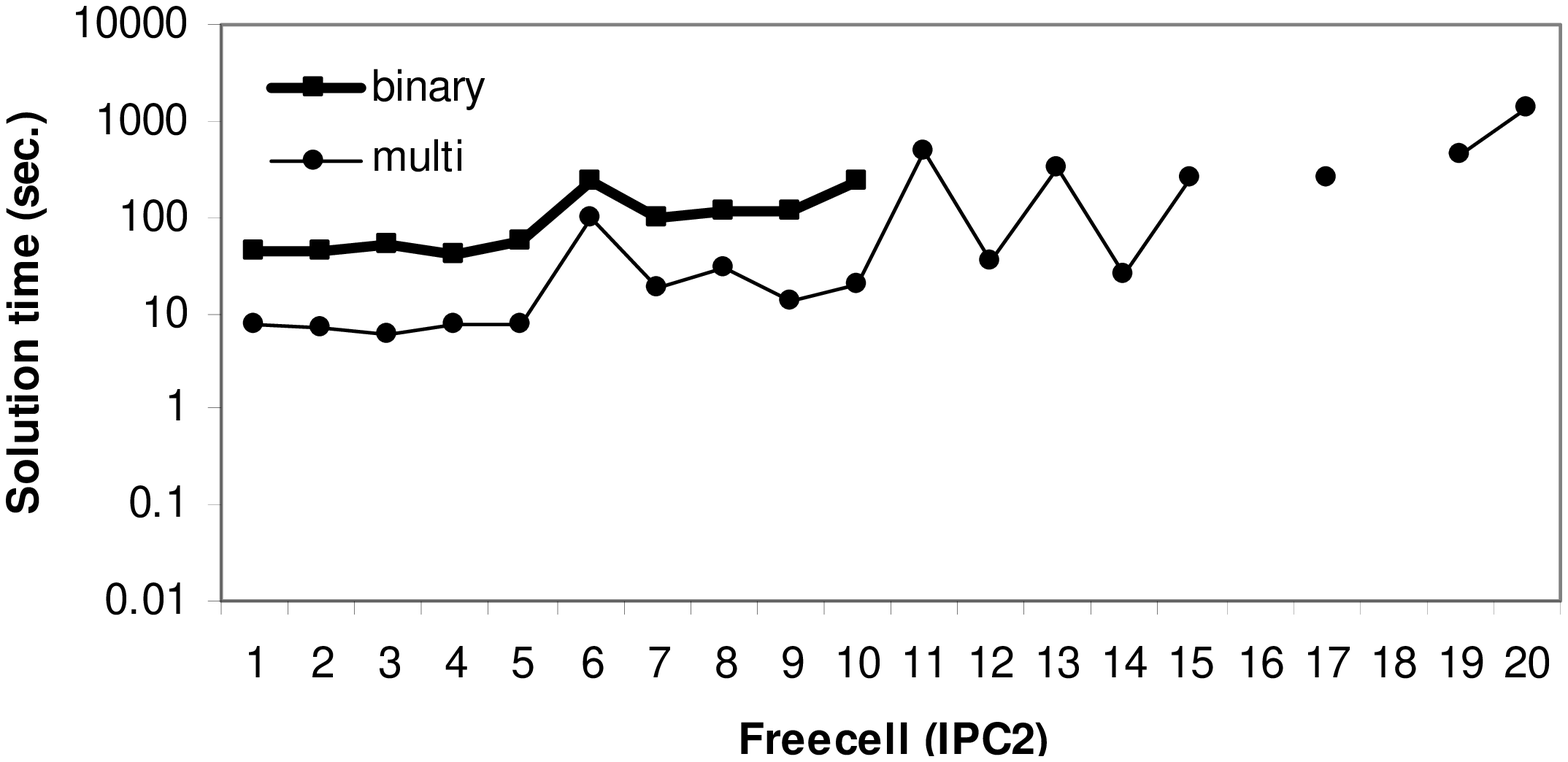} &
  \includegraphics[width=2.9in]{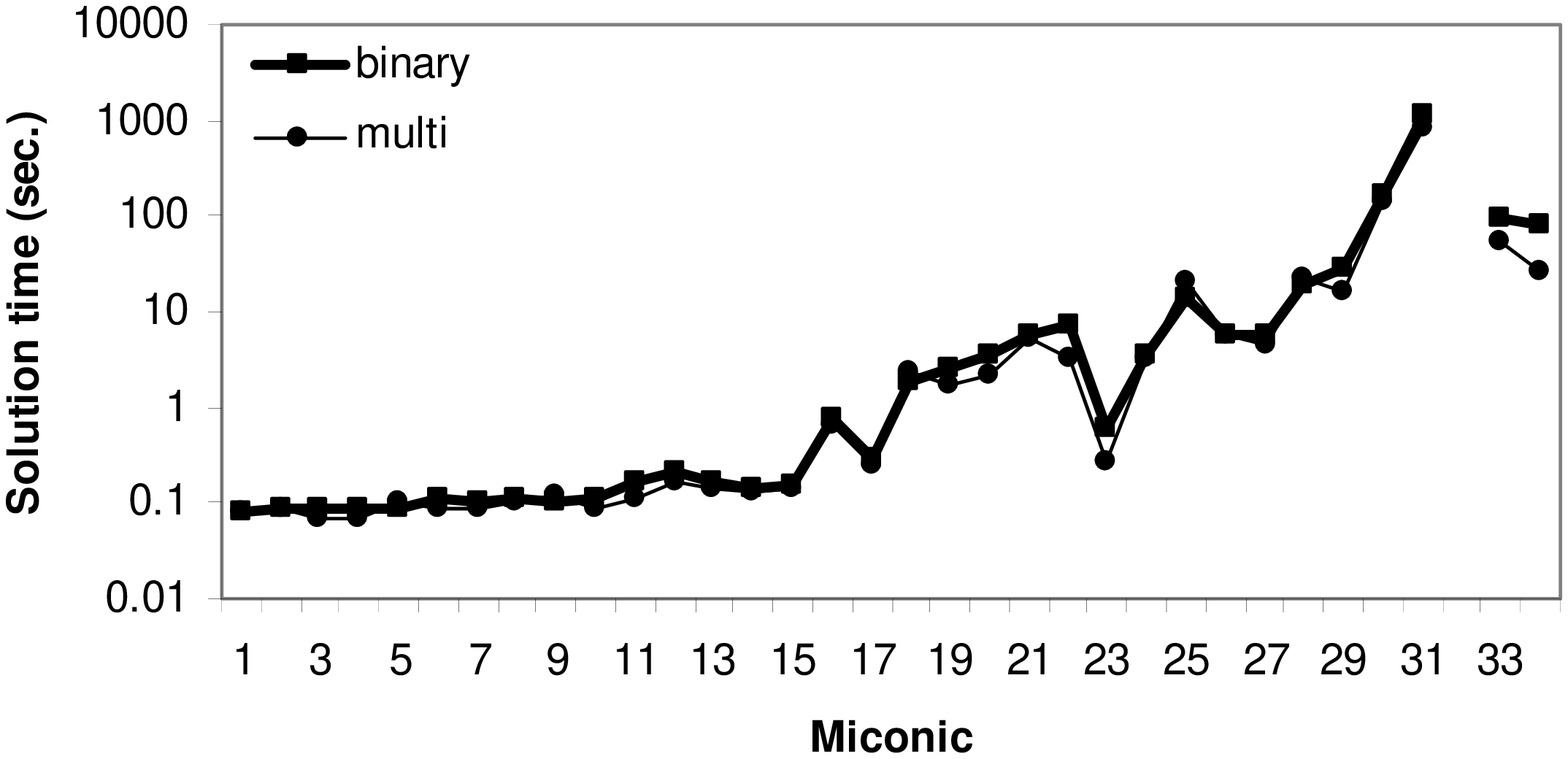} \\
  \includegraphics[width=2.9in]{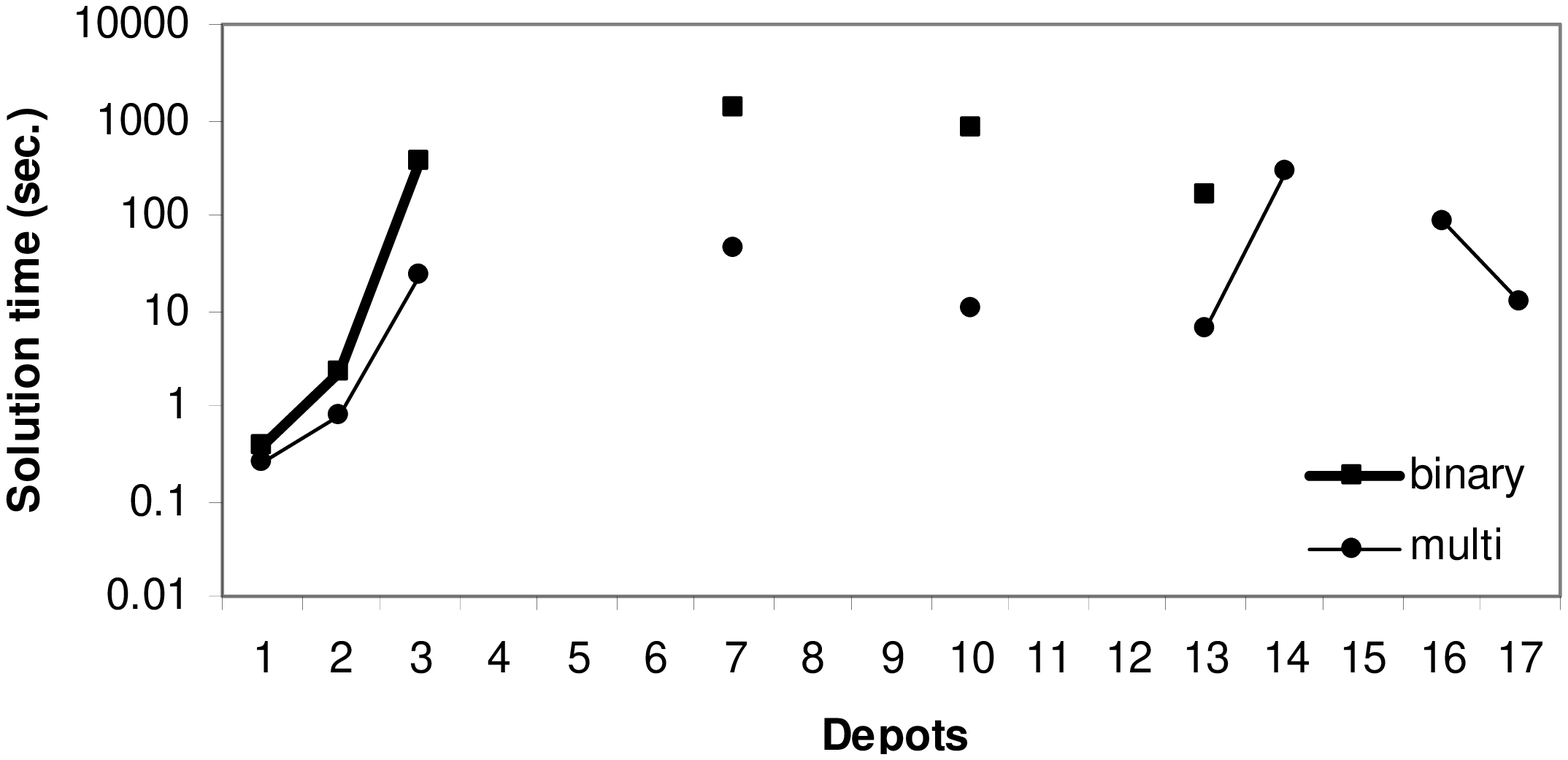} &
  \includegraphics[width=2.9in]{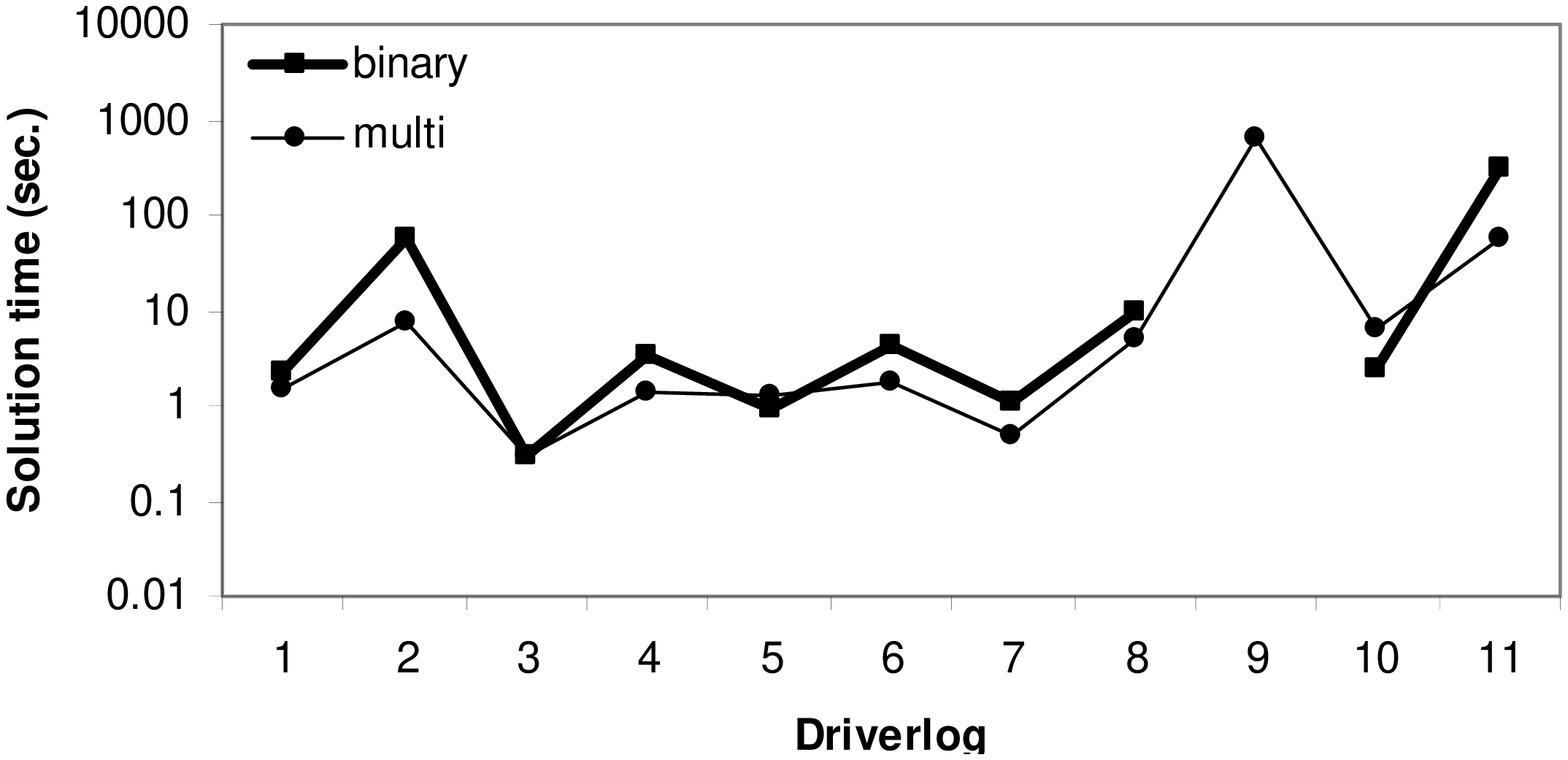} \\
  \includegraphics[width=2.9in]{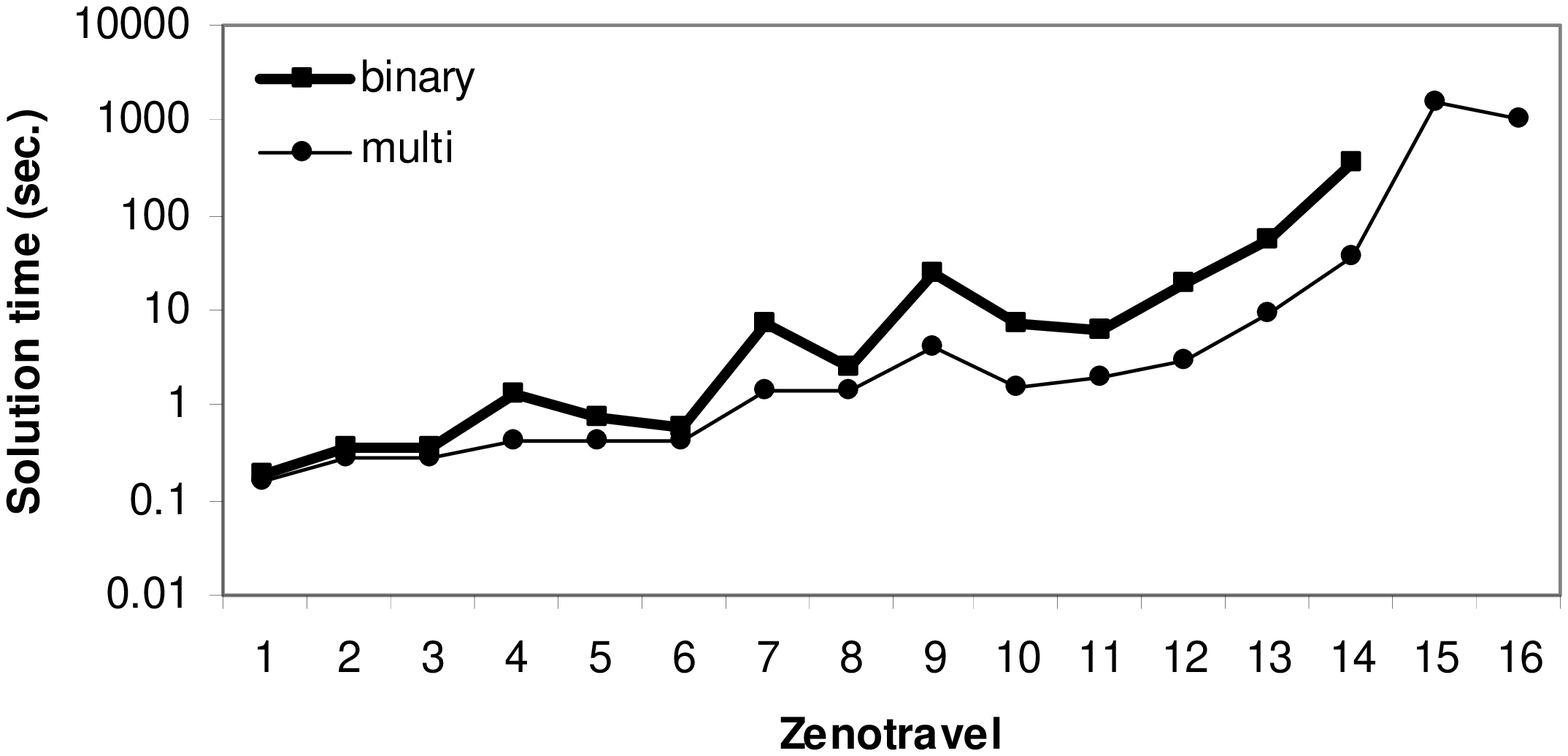} &
  \includegraphics[width=2.9in]{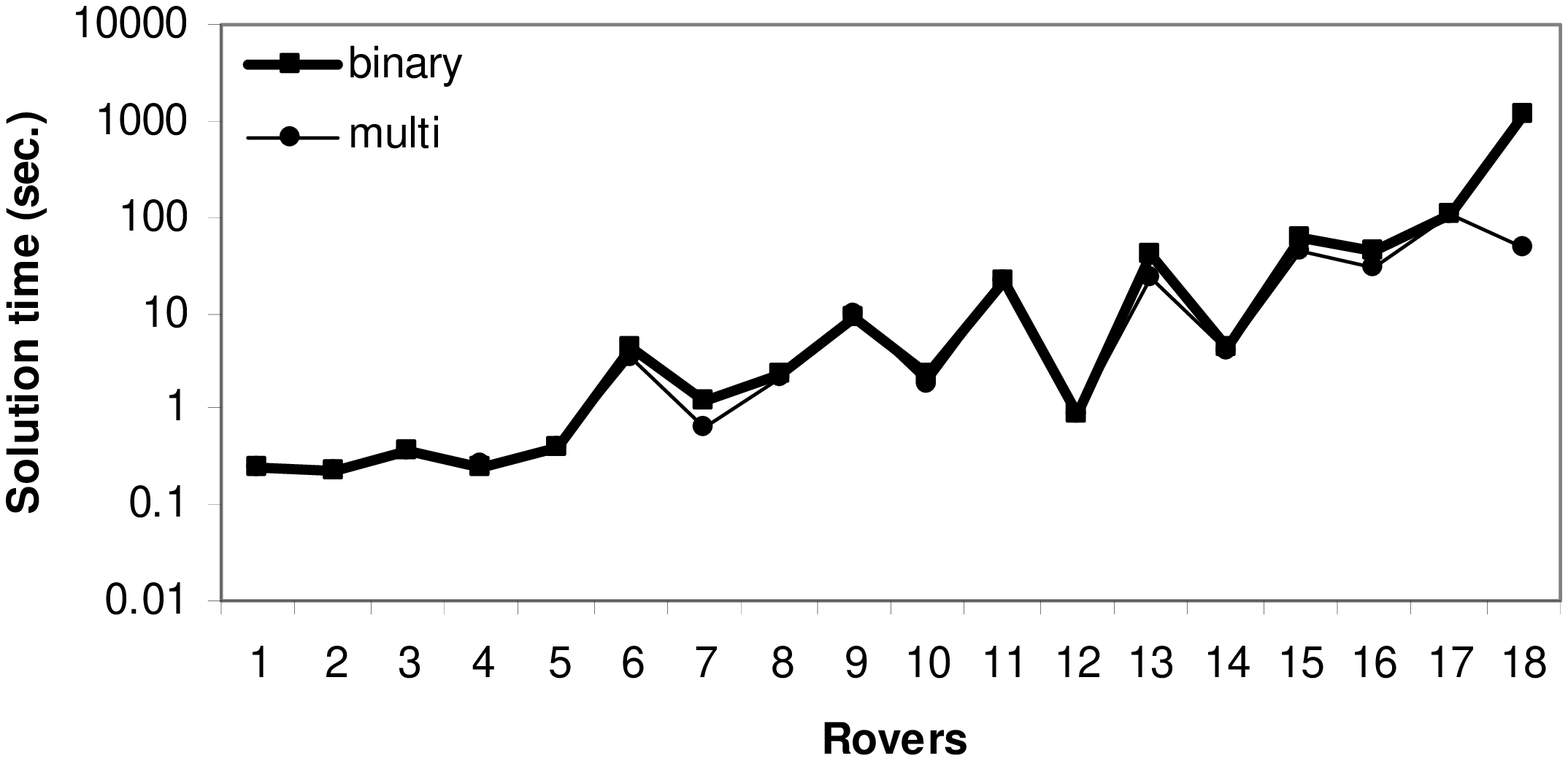} \\
  \includegraphics[width=2.9in]{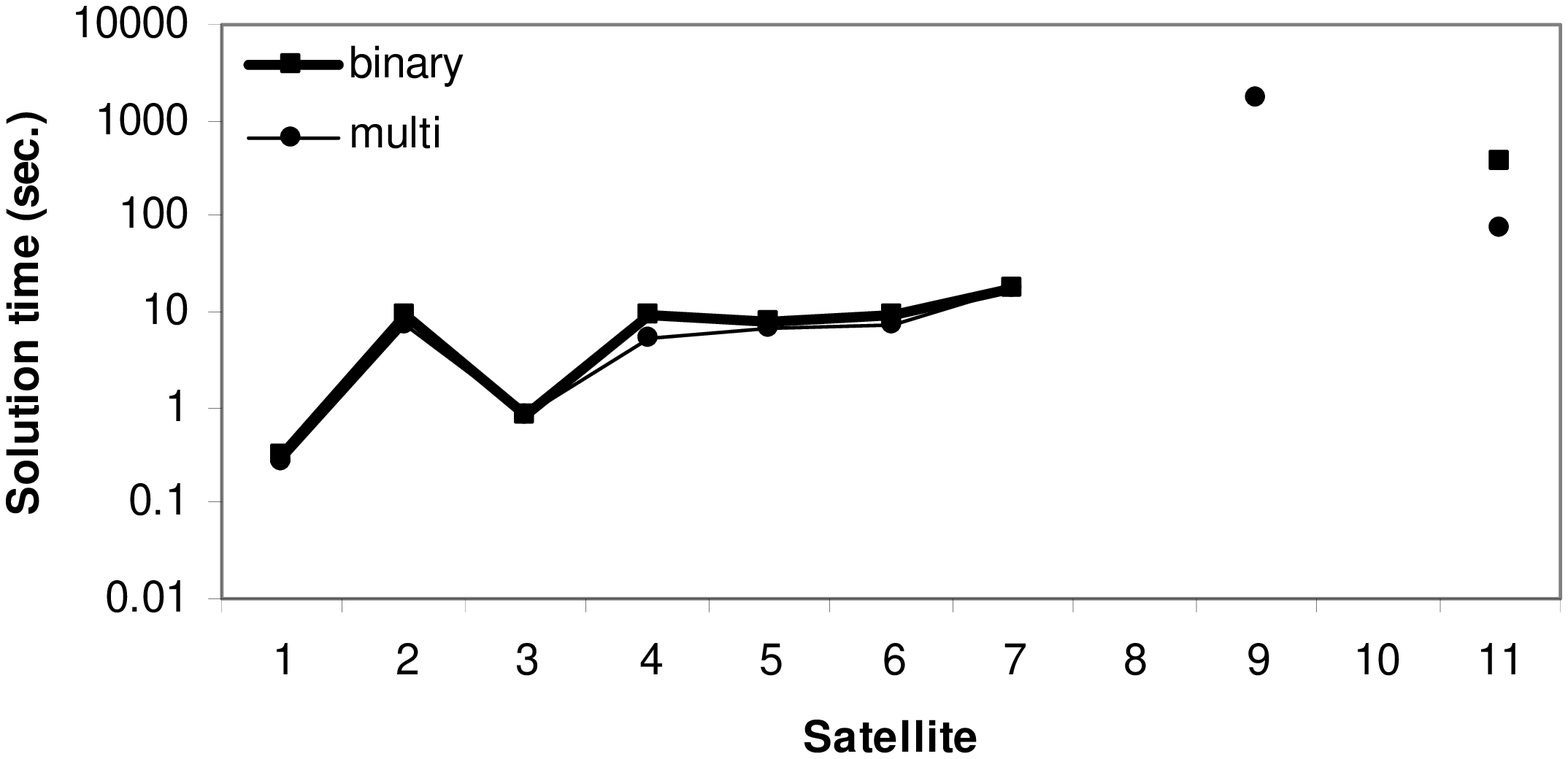} &
  \includegraphics[width=2.9in]{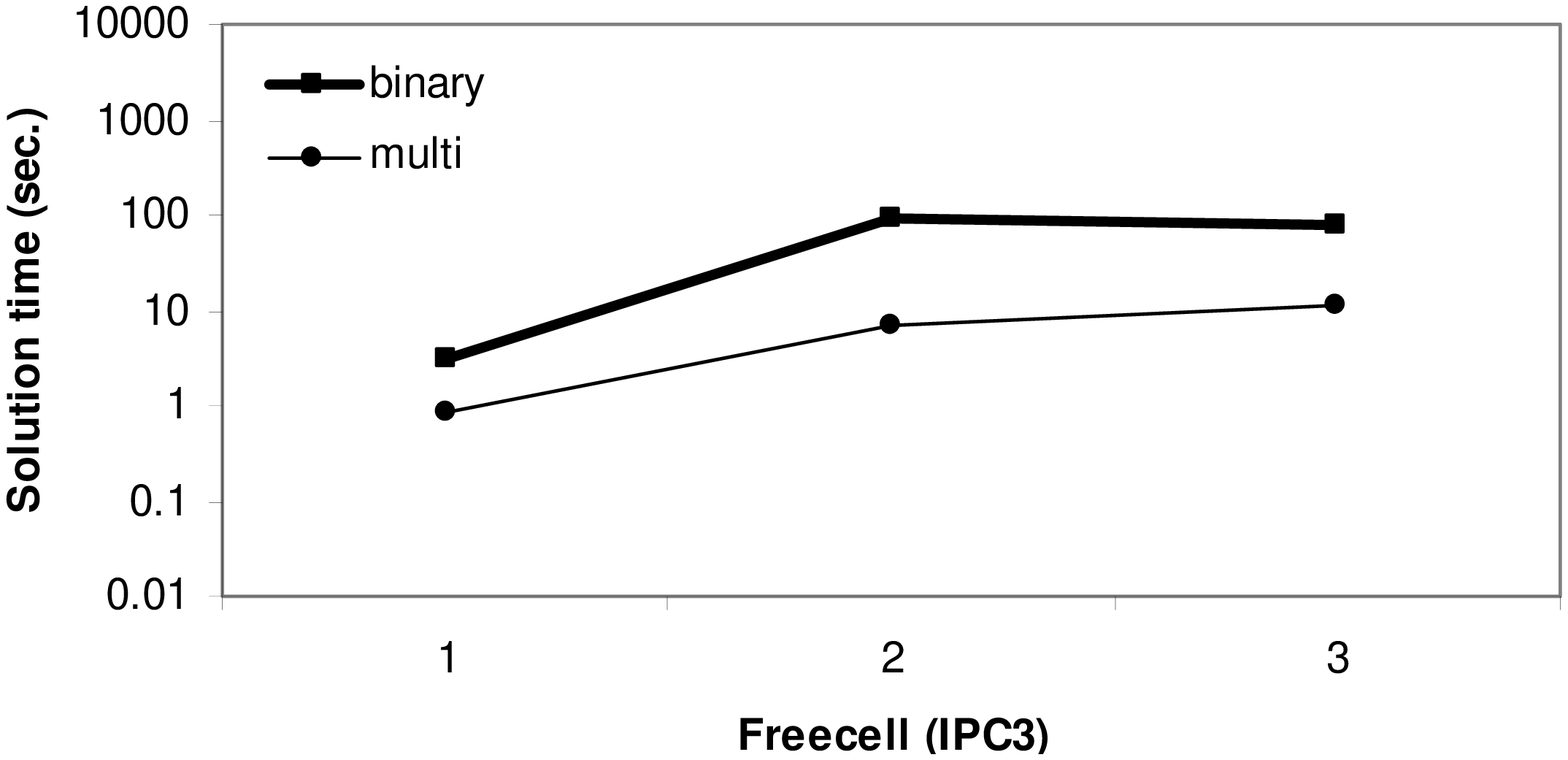}
\end{tabular}
\caption{Comparing binary state descriptions with
multi-valued state descriptions using the G1SC formulation.} \label{fig:binmul}
\end{figure*}

Figure \ref{fig:binmul} shows the total solution time (y-axis) needed to solve the problem instances (x-axis). Since we did not make any changes to the G1SC formulation, the performance differences are the result of using different state descriptions. In several domains the multi-valued state description shows a clear advantage over the binary state description when using the G1SC formulation, but there are also domains in which the multi-valued state description does not provide too much of an advantage. In general, however, the G1SC formulation using a multi-valued state description leads to the same or better performance than using a binary state description. In all our tests, we encountered only one problem instance (Rovers pfile10) in which the binary state description notably outperformed the multi-valued state description.

\begin{table}[p]
\begin{center}
{\small
\begin{tabular}{l|rrrr|rrrr}
  \hline
   & \multicolumn{4}{c|}{Binary} & \multicolumn{4}{c}{Multi}\\
   & \multicolumn{2}{c}{Before presolve} & \multicolumn{2}{c|}{After presolve} & \multicolumn{2}{c}{Before presolve} & \multicolumn{2}{c}{After presolve}\\
  Problem & \#va & \#co & \#va & \#co & \#va & \#co & \#va & \#co \\
  \hline
  Blocksworld &      &      &      &      &      &      &      &      \\
  6-2         & 7645 &12561 & 5784 & 9564 & 5125 & 7281 & 3716 & 5409 \\
  7-0         &10166 &16881 & 7384 &12318 & 6806 & 9741 & 4761 & 6967 \\
  8-0         &11743 &19657 & 9947 &16773 & 7855 &11305 & 6438 & 9479 \\
  \hline
  Logistics   &      &      &      &      &      &      &      &      \\
  14-1        &16464 &16801 & 7052 & 7386 &10843 &11180 & 2693 & 3007 \\
  15-0        &16465 &16801 & 7044 & 7385 &10844 &11180 & 2696 & 3009 \\
  15-1        &14115 &14401 & 4625 & 4935 & 9297 & 9583 & 1771 & 2133 \\
  \hline
  Miconic     &      &      &      &      &      &      &      &      \\
  6-4         & 2220 & 3403 & 1776 & 2843 & 1905 & 3088 &  428 & 1495 \\
  7-0         & 2842 & 4474 & 2295 & 3764 & 2473 & 4105 &  503 & 1972 \\
  7-2         & 2527 & 3977 & 1999 & 3287 & 2199 & 3649 &  431 & 1719 \\
  \hline
  Freecell(IPC2)&    &      &      &      &      &      &      &      \\
  3-3         &128636&399869&27928 &79369 &25267 &62588 & 7123 &15588 \\
  3-4         &129392&401486&28234 &79577 &23734 &61601 & 6346 &15101 \\
  3-5         &128636&399869&27947 &79444 &23342 &61083 & 6237 &14931 \\
  \hline
  Depots      &      &      &      &      &      &      &      &      \\
  7           &21845 &36181 &11572 &23233 &17250 &15381 & 4122 & 5592 \\
  10          &30436 &50785 &13727 &27570 &24120 &21713 & 4643 & 6731 \\
  13          &36006 &59425 &14729 &29712 &27900 &25297 & 4372 & 6806 \\
  \hline
  Driverlog   &      &      &      &      &      &      &      &      \\
  8           & 3431 & 3673 & 2245 & 2506 & 2595 & 2513 & 1146 & 1102 \\
  10          & 4328 & 4645 & 2159 & 2333 & 3551 & 3292 & 1409 & 1171 \\
  11          & 8457 & 9101 & 5907 & 6404 & 6997 & 6471 & 3558 & 3073 \\
  \hline
  Zenotravel  &      &      &      &      &      &      &      &      \\
  12          & 9656 &15589 & 4294 & 7046 & 2858 & 5821 & 1051 & 2398 \\
  13          &13738 &21649 & 7779 &12449 & 4466 & 8417 & 1882 & 4174 \\
  14          &40332 &70021 &17815 &32959 & 9282 &24121 & 3367 &10619 \\
  \hline
  Rovers      &      &      &      &      &      &      &      &      \\
  16          & 8631 & 8093 & 5424 & 5297 & 7367 & 6637 & 4394 & 4155 \\
  17          &25794 &23906 &19549 &18384 &22889 &20700 &16652 &15257 \\
  18          &20895 &20241 &12056 &12144 &18351 &17377 &10081 & 9988 \\
  \hline
  Satellite   &      &      &      &      &      &      &      &      \\
  6           & 4471 & 4945 & 3584 & 3774 & 4087 & 4561 & 2288 & 2478 \\
  7           & 5433 & 5833 & 4294 & 4267 & 5013 & 5413 & 2974 & 2925 \\
  11          &16758 &21537 &13643 &16713 &15578 &20357 & 7108 &10118 \\
  \hline
  Freecell(IPC3)&    &      &      &      &      &      &      &      \\
  1           & 7332 &19185 & 2965 & 7339 & 1624 & 3265 &  624 & 1339 \\
  2           &28214 &76343 &16218 &43427 & 4873 &11383 & 2604 & 6416 \\
  3           &39638 &105995&19603 &50819 & 7029 &16003 & 3394 & 8055 \\
  \hline
 \end{tabular}}
\caption{Formulation size for binary and multi-valued state description of problem instances from the IPC2 and IPC3 in number of variables (\#va), number of constraints (\#co), and number of ordering constraints, or cuts, (\#cu) that were added dynamically through the solution process.}\label{tab:binmul}
\end{center}
\end{table}

\section{Related Work}\label{sec:related work}
There are only few integer programming-based planning systems. Bylander \citeyear{BYL1997} considers an IP formulation based on converting the propositional representation given by Satplan \cite{KAUSEL1992} to an IP formulation with variables that take the value 1 if a certain proposition is true, and 0 otherwise. The LP relaxation of this formulation is used as a heuristic for partial order planning, but tends to be rather time-consuming. A different IP formulation is given by Vossen et al.\ \citeyear{VOSetal1999}. They consider an IP formulation in which the original propositional variables are replaced by state change variables. State change variables take the value 1 if a certain proposition is added, deleted, or persisted, and 0 otherwise. Vossen et al.\ show that the formulation based on state change variables outperforms a formulation based on converting the propositional representation. Van den Briel and Kambhampati \citeyear{BRIKAM2005} extend the work by Vossen et al.\ by incorporating some of the improvements described by Dimopoulos \citeyear{DIM2001}. Other integer programming approaches for planning rely on domain-specific knowledge \cite{BOCDIM1998,BOCDIM1999} or explore non-classical planning problems \cite{DIMGER2002,KAUWAL1999}.

In our formulations we model the transitions of each state variable as a separate flow problem, with the individual problems being connected through action constraints. The Graphplan planner introduced the idea of viewing planning as a network flow problem, but it did not decompose the domain into several loosely coupled components. The encodings that we described are related to the loosely-coupled modular planning architecture by Srivastava, Kambhampati, and Do \citeyear{SRIetal2001}, as well as factored planning approaches by Amir and Engelhardt \citeyear{AMIENG2003}, and Brafman and Domshlak \citeyear{BRADOM2006}. The work by Brafman and Domshlak, for example, proposes setting up a separate CSP problem for handling each factor. These individual factor plans are then combined through a global CSP. In this way, it has some similarities to our work (with our individual state variable flows corresponding to encodings for the factor plans). Although Brafman and Domshlak do not provide empirical evaluation of their factored planning framework, they do provide some analysis on when factored planning is expected to do best. It would be interesting to adapt their minimal tree-width based analysis to our scenario.

The branch-and-cut concept was introduced by Gr\"{o}tschel, R\"{u}nger, and Reinelt \citeyear{GROetal1984} and Padberg and Rinaldi \citeyear{PADRIN1991}, and has been successfully applied in the solution of many hard large-scale optimization problems \cite{capfis1997}. In planning, approaches that use dynamic constraint generation during search include RealPlan \cite{SRIetal2001} and LPSAT \cite{WOLWEL1999}.

Relaxed definitions for Graphplan-style parallelism have been investigated by several other researchers. Dimopoulos et al.\ \citeyear{DIMetal1997} were the first to point out that it is not necessary to require two actions to be independent in order to execute them in the same plan period. In their work they introduce the property of post-serializability of a set of actions. A set of actions $A'\subseteq A$ is said to be post-serializable if (1) the union of their preconditions is consistent, (2) the union of their effects is consistent, and (3) the preconditions-effects graph is acyclic. Where the preconditions-effects graph is a directed graph that contains a node for each action in $A'$, and an arc $(a,b)$ for $a,b\in A'$ if the preconditions of $a$ are inconsistent with the effects of $a$. For certain planning problems Dimopoulos et al.\ \citeyear{DIMetal1997} show that their modifications reduce the number of plan periods and improve performance. Rintanen \citeyear{rin1998} provides a constraint-based implementation of their idea and shows that the improvements hold over a broad range of planning domains.

Cayrol et al.\ \citeyear{cayetal2001} introduce the notion of authorized linearizations, which implies an order for the execution of two actions. In particular, an action $a\in A$ authorizes an action $b\in A$ implies that if $a$ is executed before $b$, then the preconditions of $b$ will not be deleted by $a$ and the effects of $a$ will not be deleted by $b$. The notion of authorized linearizations is very similar to the property of post-serializability. If we were to adopt these ideas in our network-based representations it would compare to the G1SC network in which the generalized state change arcs (see Figure \ref{fig:g1scnetwork}) only allows values to prevail after, but not before, each of the transition arcs.

A more recent discussion on the definitions of parallel plans is given by Rintanen, Heljanko and Niemel\"{a} \citeyear{RINetal2006}. Their work introduces the idea of $\exists$-step semantics, which says that it is not necessary that all parallel actions are non-interfering as long as they can be executed in at least one order. $\exists$-step semantics provide a more general interpretation of parallel plans than the notion of authorized linearizations used by LCGP \cite{cayetal2001}. The current implementation of $\exists$-step semantics in Satplanner is, however, somewhat restricted. While the semantics allow actions to have conflicting effects, the current implementation of Satplanner does not.

\section{Conclusions}\label{sec:conclusions}
This work makes two contributions: (1) it improves state of the art in modeling planning as integer programming, and (2) it develops novel decomposition methods for solving bounded length (in terms of number of plan periods) planning problems.

We presented a series of IP formulations that represent the planning problem as a set of loosely coupled network flow problems, where each network flow problem corresponds to one of the state variables in the planning domain. We incorporated different notions of action parallelism in order to reduce the number of plan periods needed to find a plan and to improve planning efficiency.
The IP formulations described in this paper have led to their successful use in solving partial satisfaction planning problems \cite{DOetal2007}. Moreover, they have initiated a new line of work in which integer and linear programming are used in heuristic state-space search for automated planning \cite{BENetal2007,BRIetal2007A}.
It would be interesting to see how our approach in the context of IP formulations applies to formulations based on satisfiability and constraint satisfaction.

\newpage
\acks
This research is supported in part by the NSF grant IIS–308139, the ONR grant N000140610058, and by the Lockheed Martin subcontract TT0687680 to Arizona State University as part of the DARPA integrated learning program. We thank Derek Long for his valuable input, and we especially thank the anonymous reviewers whose attentive comments and helpful suggestions have greatly improved this paper.

\bibliographystyle{theapa}
\bibliography{bibliography}
\end{document}